\documentclass[10pt,twocolumn,letterpaper]{article}

\usepackage[pagenumbers]{cvpr} % To force page numbers, e.g. for an arXiv version

\usepackage[accsupp]{axessibility} % Improves PDF readability for those with visual impairments.

\usepackage[dvipsnames]{xcolor}

\usepackage{graphicx}
\usepackage{amsmath}
\usepackage{amssymb}
\usepackage{booktabs}
\usepackage{microtype}
\usepackage{enumitem}
\usepackage{array,multirow,textcomp}
\usepackage{balance}
\usepackage[normalem]{ulem}

\usepackage{algorithm}  
\usepackage{algpseudocode}
\usepackage{pifont}% http://ctan.org/pkg/pifont
\newcommand{\cmark}{\ding{51}}%
\newcommand{\xmark}{\ding{55}}%
\usepackage{arydshln}
\usepackage{bm}

\usepackage{xcolor}

\usepackage{supertabular,booktabs}

\newcommand{\method}{Marigold}%
\definecolor{darkgreen}{rgb}{0.0, 0.7, 0.0}

\definecolor{cvprblue}{rgb}{0.21,0.49,0.74}
\usepackage[pagebackref,breaklinks,colorlinks,citecolor=cvprblue]{hyperref}
\usepackage[capitalize]{cleveref}
\usepackage{fmtcount}

\usepackage{makecell}  % for wrapping multi-row 

\title{Repurposing Diffusion-Based Image Generators for Monocular Depth Estimation}

\author{
Bingxin Ke\qquad 
Anton Obukhov\qquad 
Shengyu Huang\\
Nando Metzger \quad
Rodrigo Caye Daudt\quad 
Konrad Schindler \\
{\normalsize Photogrammetry and Remote Sensing, ETH Zürich}
}

\begin{document}
% \maketitle

\twocolumn[{%
		\renewcommand\twocolumn[1][]{#1}%
		\maketitle
            \vspace{-2em}
		\begin{center}
			\includegraphics[width=\textwidth]{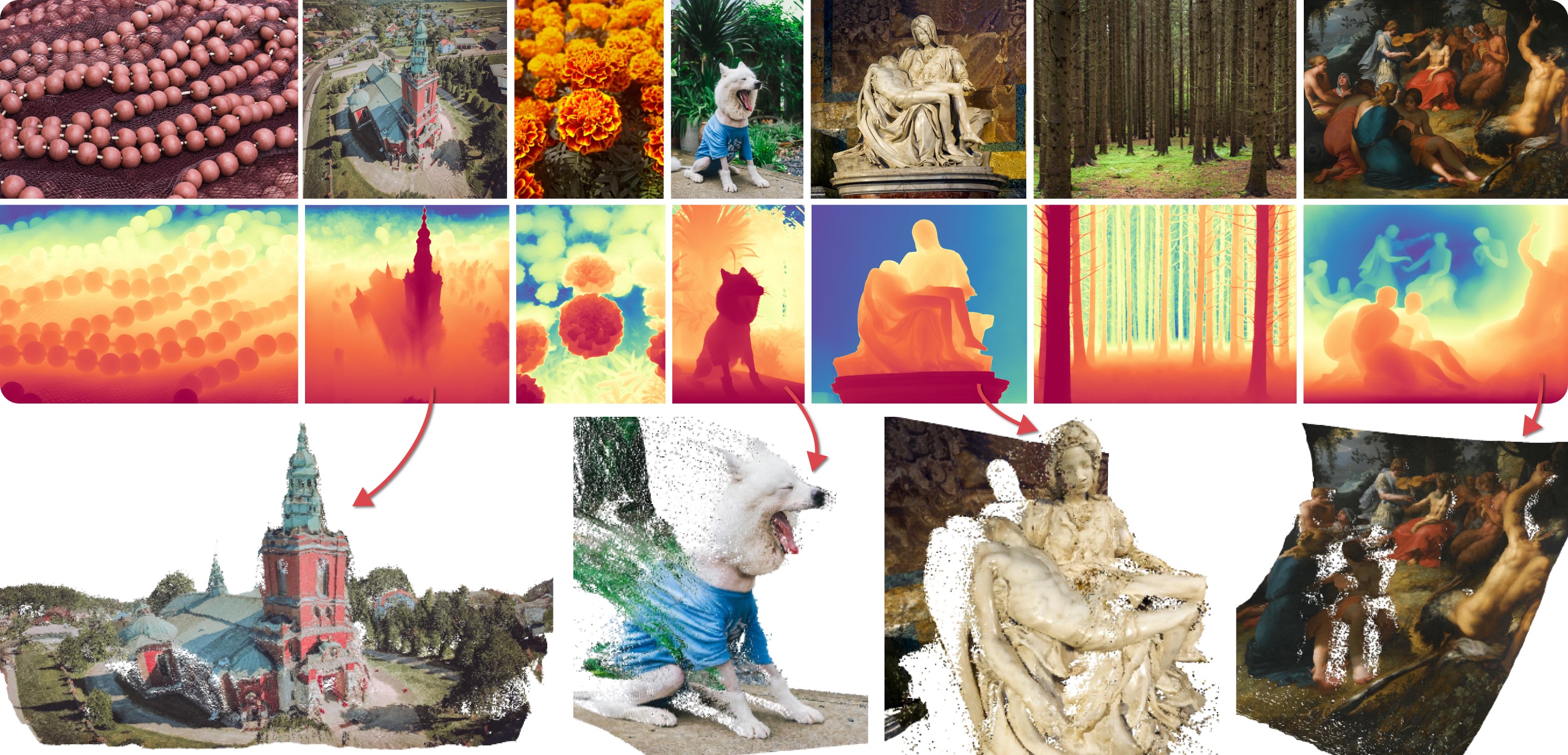}
\vspace{-1em}			
   \captionsetup{type=figure}
			\captionof{figure}{
				\textbf{
					We present \method{}, a diffusion model and associated fine-tuning protocol for monocular depth estimation.
				}
				Its core principle is to leverage the rich visual knowledge stored in modern generative image models. Our model, derived from Stable Diffusion 
				and fine-tuned with 
				synthetic data,
				can zero-shot transfer to unseen datasets,  
				offering state-of-the-art monocular depth estimation results.
			}
			\label{fig:teaser}
		\end{center}
            \vspace{1em}
	}]

\begin{abstract}
Monocular depth estimation is a fundamental computer vision task. Recovering 3D depth from a single image is geometrically ill-posed and requires scene understanding, so it is not surprising that the rise of deep learning has led to a breakthrough. The impressive progress of monocular depth estimators has mirrored the growth in model capacity, from relatively modest CNNs to large Transformer architectures. Still, monocular depth estimators tend to struggle when presented with images with unfamiliar content and layout, since their knowledge of the visual world is restricted by the data seen during training, and challenged by zero-shot generalization to new domains. This motivates us to explore whether the extensive priors captured in recent generative diffusion models can enable better, more generalizable depth estimation.
We introduce \emph{\method{}}, a method for affine-invariant monocular depth estimation that is derived from Stable Diffusion and retains its rich prior knowledge. The estimator can be fine-tuned in a couple of days on a single GPU using only synthetic training data.
It delivers state-of-the-art performance across a wide range of datasets, including over 20\% performance gains in specific cases.
Project page: \href{https://marigoldmonodepth.github.io}{https://marigoldmonodepth.github.io}.
\end{abstract}    
\section{Introduction}
\label{sec:intro}

Monocular depth estimation aims to transform a photographic image into a depth map, \ie, regress a range value for every pixel.
The task arises whenever the 3D scene structure is needed, and no direct range or stereo measurements are available. 
Clearly, undoing the projection from the 3D world to a 2D image is a geometrically ill-posed problem and can only be solved with the help of prior knowledge, such as typical object shapes and sizes, likely scene layouts, occlusion patterns, \etc.
In other words, monocular depth implicitly requires scene understanding, and it is no coincidence that the advent of deep learning brought about a leap in performance.
Depth estimation is nowadays cast as neural image-to-image translation and learned in a supervised (or semi-supervised) fashion using collections of paired, aligned RGB images and depth maps.
Early methods of this type were limited to a narrow domain defined by their training data, often indoor~\cite{silberman2012indoor} or driving~\cite{Geiger2013IJRR} scenes. More recently, there has been a quest to train generic depth estimators that can be either used off-the-shelf across a broad range of scenes or fine-tuned to a specific application scenario with a small amount of data.
These models generally follow the strategy first employed by MiDAS~\cite{Ranftl2020_midas} to achieve generality, namely to train a high-capacity model with data sampled from many different RGB-D datasets (respectively, domains).
The latest developments include moving from convolutional encoder-decoder networks~\cite{Ranftl2020_midas} to increasingly large and powerful vision transformers~\cite{ranftl2021_dpt}, and training on more and more data and with additional surrogate tasks~\cite{eftekhar2021omnidata} to amass even more knowledge about the visual world, and consequently to produce better depth maps.
Importantly, visual cues for depth depend not only on the scene content but also on the (generally unknown) camera intrinsics~\cite{yin2023metric3d}. For general in-the-wild depth estimation, it is often preferred to estimate affine-invariant depth (\ie, depth values up to a global offset and scale), which can also be determined without objects of known sizes that could serve as ``scale bars''.

The intuition behind our work is the following: Modern image diffusion models have been trained on internet-scale image collections specifically to generate high-quality images across a wide array of domains~\cite{rombach2022high, betker_dalle3improving_nodate, saharia2022photorealistic}. If the cornerstone of monocular depth estimation is indeed a comprehensive, encyclopedic representation of the visual world, then it should be possible to derive a broadly applicable depth estimator from a pretrained image diffusion model.
In this paper, we set out to explore this option and develop \textbf{\method{}}, a latent diffusion model (LDM) based on Stable Diffusion~\cite{rombach2022high}, along with a fine-tuning protocol to adapt the model for depth estimation.
The key to unlocking the potential of a pretrained diffusion model is to keep its latent space intact. We find this can be done efficiently by modifying and fine-tuning only the denoising U-Net. Turning Stable Diffusion into \method{} requires only synthetic RGB-D data (in our case, the Hypersim~\cite{roberts2021hypersim} and Virtual KITTI~\cite{cabon2020virtualkitti2} datasets) and a few GPU days on a single consumer graphics card.
Empowered by the underlying diffusion prior of natural images, \method{}
exhibits excellent zero-shot generalization: Without ever having seen real depth maps, it attains state-of-the-art performance on several real datasets.
To summarize, our contributions are:
\begin{enumerate}
    \item A simple and resource-efficient fine-tuning protocol to convert a pretrained LDM image generator into an image-conditional depth estimator;
    \item \method{}, a state-of-the-art, versatile monocular depth estimation module that offers excellent performance across a wide variety of natural images. 
\end{enumerate}

\section{Related Work}
\label{sec:related_work}

%%%%%%%%%%%%%%%%%%%%%%%%%%%%% Monocular Depth %%%%%%%%%%%%%%%%%%%%%%%%%%%%%
\subsection{Monocular Depth}
At the technical level, monocular depth estimation is a dense, structured regression task.
The pioneering work of \citet{eigen_depth_2014} introduced a multi-scale network and showed that the result can be converted to metric depth for a dataset recorded with a single sensor. Successive improvements have come from various fronts, including ordinal regression~\cite{fu2018dorn}, planar guidance maps~\cite{lee2019big}, neural conditional random fields~\cite{yuan2022newcrfs}, vision transformers~\cite{yang2021transformers, li2023depthformer, aich2021bidirectional}, a piecewise planarity prior~\cite{patil2022p3depth}, first-order variational constraints~\cite{liu2023vadepth} and variational autoencoders~\cite{ning2023ait}. Some authors treat depth estimation as a combined regression-classification task, using various binning strategies like AdaBins~\cite{Farooq_Bhat_2021} or BinsFormer~\cite{li2022binsformer} to discretize depth range. A notable recent trend involves training generative models, especially diffusion models~\cite{song2020denoising_ddim, ho2020denoising_ddpm} for monocular depth estimation~\cite{ji2023ddp, duan2023diffusiondepth, saxena2023depthgen, saxena2023surprising_ddvm}.
Recently, a few works~\cite{yin2023metric3d,guizilini2023towards_zero_depth} have revisited absolute depth estimation, by explicitly feeding camera intrinsics as additional input.

Estimating depth ``in the wild'' refers to methods that are successful across a wide range of (possibly unfamiliar) settings, a particularly challenging task. It has been addressed by constructing large and diverse depth datasets and designing algorithms to handle that diversity. 
DIW~\cite{chen2016single} was perhaps the earliest work to introduce an uncontrolled dataset and to predict \textit{relative (ordinal) depth} for it. 
OASIS~\cite{chen2020oasis} introduced relative depth and normals to better generalize across scenes.
However, relative depth predictions (depth ordering) are of limited use for many downstream tasks, which has led several authors to consider \textit{affine-invariant depth}. In that setting, depth is estimated up to an unknown (global) offset and scale. 
It offers a viable compromise between the ordinal and metric cases: on the one hand, it can handle general scenes consisting of unfamiliar objects; on the other hand, depth differences between different objects or scene parts are still geometrically meaningful relative to each other.
MegaDepth~\cite{MDLi18_megadepth} and DiverseDepth~\cite{yin2020diversedepth} utilize large internet photo collections to train models that can adapt to unseen data, while MiDaS~\cite{Ranftl2020_midas} achieves generality by training on a mixture of multiple datasets. The step from CNNs to vision transformers has further boosted performance, as evidenced by DPT~\cite{ranftl2021_dpt} and Omnidata~\cite{eftekhar2021omnidata}. 
LeReS~\cite{Wei2021CVPR_leres} proposed a two-stage framework that first predicts affine-invariant depth, then upgrades it to metric depth by estimating the shift and focal length.
HDN~\cite{zhang2022_hdn} introduced multi-scale depth normalization to improve the prediction details and smoothness further. 
While this enables the depth estimator to handle images captured with different known cameras, it does not include the true in-the-wild setting, where the camera intrinsics of the test images are unknown.
Our method addresses affine-invariant depth estimation but does not focus on compiling an extensive, annotated training dataset. Rather, we utilize the broader image priors in image LDMs and apply fine-tuning.

%%%%%%%%%%%%%%%%%%%%%%%%%%%%% Diffusion Models %%%%%%%%%%%%%%%%%%%%%%%%%%%%%
\subsection{Diffusion Models}

Denoising Diffusion Probabilistic Models (DDPMs)~\cite{ho2020denoising_ddpm} have emerged as a powerful class of generative models.
They learn to reverse a diffusion process that progressively degrades images with Gaussian noise so that they can draw samples from the data distribution by applying the reverse process to random noise.
This idea was extended to DDIMs~\cite{song2020denoising_ddim}, which provide a non-Markovian shortcut for the diffusion process.
\textit{Conditional diffusion models} are an extension of DDPMs~\cite{ho2020denoising_ddpm, song2020denoising_ddim} that ingest additional information on which the output is then conditioned, similar to cGAN~\cite{mirza2014conditional} and cVAE~\cite{sohn2015cvae}.
Conditioning can take various forms, including text~\cite{saharia2022photorealistic}, other images~\cite{saharia2022palette}, or semantic maps~\cite{zhang2023adding}.

In the realm of text-based image generation, \citet{rombach2022high} have trained a diffusion model on the large-scale image and text dataset LAION-5B~\cite{schuhmann2022laion5b} and demonstrated image synthesis with previously unattainable quality. 
The cornerstone of their approach is a latent diffusion model (LDM), where the denoising process is run in an efficient latent space, drastically reducing the complexity of the learned mapping.
Such models distill internet-scale image sets into model weights, thereby developing a rich scene understanding prior, which we harness for monocular depth estimation.

%%%%%%%%%%%%%%%%%%%%%%%%%%%%% Diffusion for Monocular Depth Estimation %%%%%%%%%%%%%%%%%%%%%%%%%%%%%
\subsection{Diffusion for Monocular Depth Estimation}

Several methods have tried to use DDPMs for metric depth estimation.
The DDP approach~\cite{ji2023ddp} proposes an architecture to encode the image but decode a depth map and has obtained state-of-the-art results on the KITTI dataset. 
DiffusionDepth~\cite{duan2023diffusiondepth} performs diffusion in the latent space, conditioned on image features extracted with a SwinTransformer.
DepthGen~\cite{saxena2023depthgen} extends a multi-task diffusion model to metric depth prediction, including handling noisy ground truth. 
Its successor DDVM~\cite{saxena2023surprising_ddvm} emphasizes pretraining on synthetic and real data for enhanced depth estimation.
Finally, VPD~\cite{zhao2023vpd} employs a pretrained Stable Diffusion model as an image feature extractor with additional text input.

Our approach advances beyond these methods, which perform well but only in their specific training domains. We explore the potential of pretrained LDMs for single-image depth estimation across diverse, real-world settings.

%%%%%%%%%%%%%%%%%%%%%%%%%%%%% Foundation Models %%%%%%%%%%%%%%%%%%%%%%%%%%%%%
\subsection{Foundation Models}
Vision foundation models (VFMs) are large neural networks trained on internet-scale data.
The extreme scaling leads to the emergence of high-level visual understanding, such that the model can then be used as is~\cite{wang2023breathing} or fine-tuned to a wide range of downstream tasks with minimal effort~\cite{bommasani2022opportunities}. 
\emph{Prompt tuning} methods~\cite{yao2023visual, zhang2023prompt, bahng2022exploring} can efficiently adapt VFMs towards dedicated scenarios by designing suitable prompts.
\emph{Feature adaptation} methods~\cite{gao2023clip, zhang2021tip, svladapterbmvc2022, zhou2022extract, zhao2023vpd, NEURIPS2023_e7407ab5} can further pivot VFMs towards different tasks.
\Eg, VPD~\cite{zhao2023vpd} showed the potential to extract features from a pre-trained text-to-image model for (domain-specific) depth estimation.
Concurrently, I-LoRA~\cite{du2023generative} demonstrated the multi-modal capabilities of pre-trained image generators.
\emph{Direct tuning} enables more flexible adaptation, not only for few-shot customization scenarios like DreamBooth~\cite{ruiz2023dreambooth} but also for object detection, as in 3DiffTection~\cite{xu20233difftection}.

The \method{} depth estimator proposed here can be interpreted as an instance of such direct tuning, where StableDiffusion plays the role of the foundation model. With as few as 74k synthetic depth samples, we obtain state-of-the-art depth estimates on real image datasets, and convincing performance in the wild (\cf \cref{fig:teaser}).

\section{Method}
\label{sec:method}

\newcommand{\img}{\mathbf{x}}
\newcommand{\depth}{\mathbf{d}}
\newcommand{\latent}{\mathbf{z}}
\newcommand{\latentdepth}{\latent^{(\depth)}}
\newcommand{\latentimage}{\latent^{(\img)}}
\newcommand{\noise}{\bm{\epsilon}}
\newcommand{\denoiser}{\bm{\epsilon}_{\theta}}
\newcommand{\denoiserlong}{\denoiser(\latentdepth_t, \latentimage, t)}
\newcommand{\catinput}{\mathbf{z}}
\newcommand{\encoder}{\mathcal{E}}
\newcommand{\decoder}{\mathcal{D}}

%%%%%%%%%%%%%%%%%%%%%%%%%%%%% Generative Formulation %%%%%%%%%%%%%%%%%%%%%%%%%%%%%
\subsection{Generative Formulation}
\label{sec:preliminary}

We pose monocular depth estimation as a conditional denoising diffusion generation task and train \method{} to model the conditional distribution $D(\depth~|~\img)$ over depth $\depth \in \mathbb{R}^{W \times H}$, where the condition $\img \ {\in}\  \mathbb{R}^{W \times H \times 3}$ is an RGB image.

In the \textit{forward} process, which starts at $\depth_0 := \depth$ from the conditional distribution, Gaussian noise is gradually added at levels $t \in \{1,..., T\}$ 
to obtain noisy samples $\depth_t$ as
\begin{equation}
    \depth_t = \sqrt{\bar{\alpha}_t} \depth_0 + \sqrt{1 - \bar{\alpha}_t} \noise
\end{equation}
where $\noise \sim \mathcal{N}(0, I),
\bar{\alpha}_t := 
\prod_{s=1}^{t}{1 \!-\! \beta_s}$, 
and $\{\beta_1, \ldots, \beta_T\}$ is the variance schedule of a process with $T$ steps. 
In the \textit{reverse} process, the conditional denoising model $\denoiser(\cdot)$ parameterized with learned parameters $\theta$ gradually removes noise from $\depth_t$ to obtain  $\depth_{t-1}$. 

At training time, parameters $\theta$ are updated by taking a data pair $(\img, \depth)$ from the training set, noising $\depth$ with sampled noise $\noise$ at a random timestep $t$, computing the noise estimate $\hat{\noise} = \denoiser(\depth_t, \img, t)$ and minimizing one of the denoising diffusion objective functions.
The canonical standard noise objective $\mathcal{L}$ is given as follows~\cite{ho2020denoising_ddpm}:
\begin{equation}
\mathcal{L} = \mathbb{E}_{\depth_0, \noise \sim \mathcal{N}(0,I),t \sim \mathcal{U}(T)} \left\| \noise - \hat{\noise} \right\|^2_2.
\label{eq:diffusion_objective}
\end{equation}
At inference time, $\depth := \depth_0$ 
is reconstructed starting from a normally-distributed variable $\depth_T$, by iteratively applying the learned denoiser $\bm{\epsilon}_\theta(\depth_t, \img, t)$. 

Unlike diffusion models that work directly on the data, latent diffusion models perform diffusion steps in a low-dimensional latent space, offering computational efficiency and suitability for high-resolution image generation~\cite{rombach2022high}. 
The latent space is constructed in the bottleneck of a variational autoencoder (VAE) trained independently of the denoiser to enable latent space compression and perceptual alignment with the data space.
To translate our formulation into the latent space, for a given depth map $\depth$, the corresponding latent code is given by the encoder $\encoder$: $\latentdepth = \encoder(\depth)$.
Given a depth latent code, a depth map can be recovered with the decoder $\decoder$: $\hat{\depth} = \decoder(\latentdepth)$.
The conditioning image $\img$ is also naturally translated into the latent space as $\latentimage = \encoder(\img)$.
The denoiser is henceforth trained in the latent space: $\denoiserlong$.
The adapted inference procedure involves one extra step -- the decoder $\decoder$ reconstructing the data $\hat{\depth}$ from the estimated clean latent $\latentdepth_0$: $\hat{\depth} = \decoder(\latentdepth_0)$.

%%%%%%%%%%%%%%%%%%%%%%%%%%%%% Network Architecture %%%%%%%%%%%%%%%%%%%%%%%%%%%%%
\subsection{Network Architecture}
\label{sec:architecture}

\begin{figure}[t!]
    \centering
    \includegraphics[width=\linewidth]{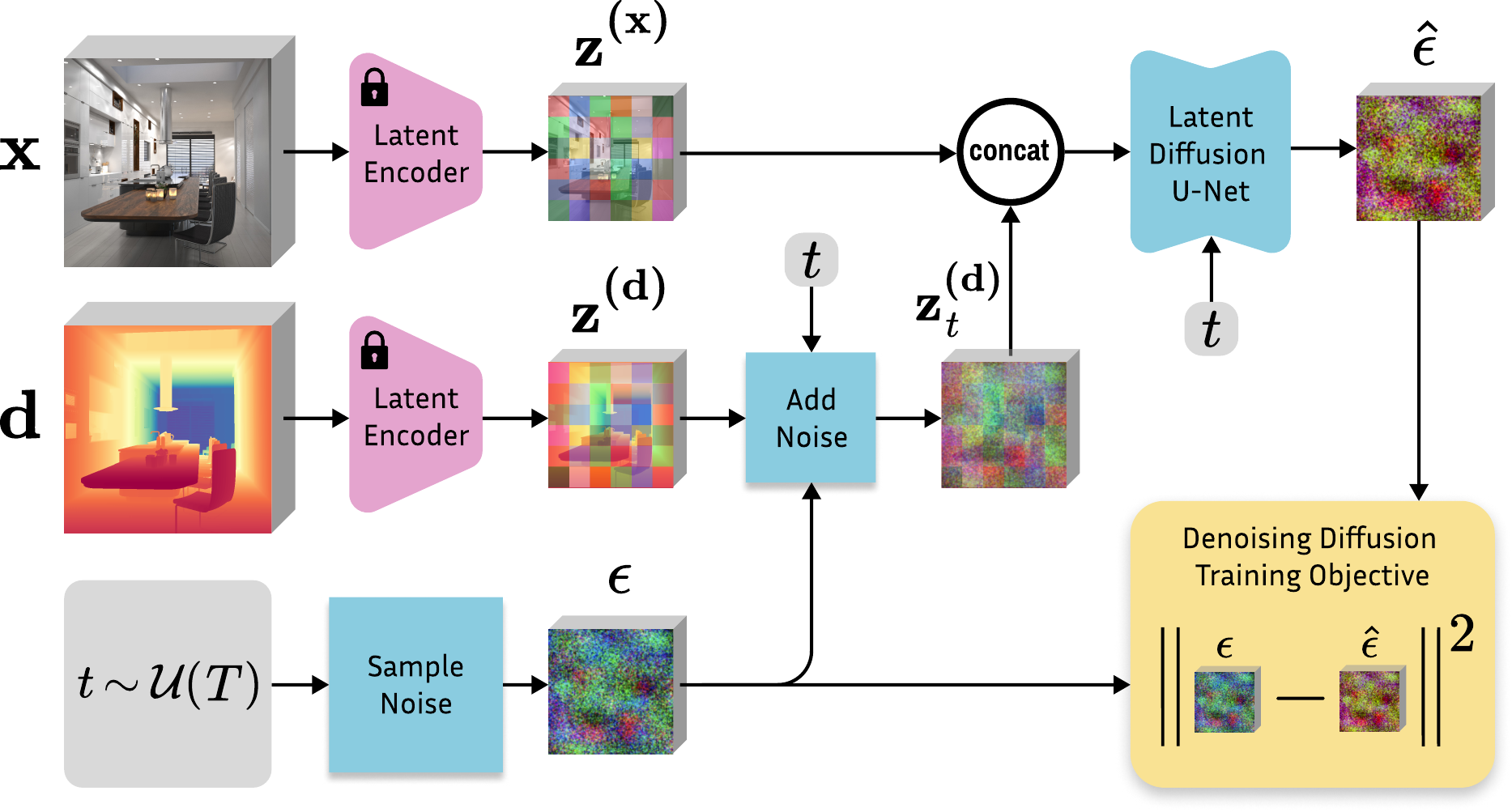}
    \caption{
    \textbf{Overview of the Marigold fine-tuning protocol.}
    Starting from pretrained Stable Diffusion, we encode the image $\img$ and depth $\depth$ into the latent space using the original Stable Diffusion VAE.
    We fine-tune just the U-Net by optimizing the standard diffusion objective relative to the depth latent code.
    Image conditioning is achieved by concatenating the two latent codes before feeding them into the U-Net. 
    The first layer of the U-Net is modified to accept concatenated latent codes.
    See details in \cref{sec:architecture} and \cref{sec:finetuning}.
    }
    \label{fig:method-train}
\end{figure}

One of our main objectives is training efficiency since diffusion models are often extremely resource-intensive to train. Therefore, we base our model on a pretrained text-to-image LDM~(Stable Diffusion v2~\cite{rombach2022high}), which has learned very good image priors from LAION-5B~\cite{schuhmann2022laion5b}. With minimal changes to the model components, we turn it into an image-conditioned depth estimator.
\cref{fig:method-train} contains an overview of the proposed fine-tuning procedure.

% ----------------------------------
\vspace{0.35em}
\noindent\textbf{%
Depth encoder and decoder.
}
We take the frozen VAE to encode both the image and its corresponding depth map into a latent space for training our conditional denoiser. 
Given that the encoder, which is designed for 3-channel (RGB) inputs, receives a single-channel depth map, we replicate the depth map into three channels to simulate an RGB image.
At this point, the data range of the depth data plays a significant role in enabling affine-invariance. We discuss our normalization approach in Sec.~\ref{sec:finetuning}.
We verified that
without any modification of the VAE or the latent space structure, 
the depth map can be reconstructed from the encoded latent code with a negligible error, \ie, $\depth \approx \decoder(\encoder(\depth))$.
At inference time, the depth latent code is decoded once at the end of diffusion, and the average of three channels is taken as the predicted depth map.

% ----------------------------------
\vspace{0.35em}
\noindent\textbf{%
Adapted denoising U-Net.
}
To implement the conditioning of the latent denoiser $\denoiserlong$ on input image $\img$, we concatenate the image and depth latent codes into a single input $\catinput_t = \text{cat}(\latent^{(\depth)}_t, \latent^{(\img)})$ along the feature dimension. 
The input channels of the latent denoiser are then doubled to accommodate the expanded input $\catinput_t$. 
To prevent inflation of activations magnitude of the first layer and keep the pre-trained structure as faithfully as possible, we duplicate the weight tensor of the input layer and divide its values by two.

%%%%%%%%%%%%%%%%%%%%%%%%%%%%% Fine-Tuning Protocol %%%%%%%%%%%%%%%%%%%%%%%%%%%%%
\subsection{Fine-Tuning Protocol}
\label{sec:finetuning}

\noindent\textbf{%
Affine-invariant depth normalization.
}
For the ground truth depth maps $\depth$, we implement a linear normalization such that the depth primarily falls in the value range $[-1, 1]$,
to match the designed input value range of the VAE.
Such normalization serves two purposes.
First, it is the convention for working with the original Stable Diffusion VAE.
Second, it enforces a canonical affine-invariant depth representation independent of the data statistics -- any scene must be bounded by near and far planes with extreme depth values.
The normalization is achieved through an affine transformation computed as
\begin{equation}
\Tilde{\depth} = \left(\frac{\depth - \depth_{2}}{\depth_{98} - \depth_{2}} - 0.5\right) \times 2,
\end{equation}
where $\depth_{2}$ and $\depth_{98}$ correspond to the $2\%$ and $98\%$ percentiles of individual depth maps.
This normalization allows \method{} to focus on pure affine-invariant depth estimation.

% ----------------------------------
\vspace{0.35em}
\noindent\textbf{%
Training on synthetic data.
}
Real depth datasets suffer from missing depth values caused by the physical constraints of the capture rig and the physical properties of the sensors. 
Specifically, the disparity between cameras and reflective surfaces diverting LiDAR laser beams are inevitable sources of ground truth noise and missing pixels~\cite{wagner2006gaussian,Huang2023nfl}.
In contrast to prior work that utilized diverse real datasets to achieve generalization~\cite{Ranftl2020_midas, eftekhar2021omnidata}, we train exclusively with synthetic depth datasets. 
As with the depth normalization rationale, this decision has two objective reasons.
First, synthetic depth is inherently dense and complete, meaning that every pixel has a valid ground truth depth value, allowing us to feed such data into the VAE, which can not handle data with invalid pixels.
Second, synthetic depth is the cleanest possible form of depth, which is guaranteed by the rendering pipeline. 
If our assumption about the possibility of fine-tuning a generalizable depth estimation from a text-to-image LDM is correct, then synthetic depth gives the cleanest set of examples and reduces noise in gradient updates during the short fine-tuning protocol.
Thus, the remaining concern is the sufficient diversity or domain gaps between synthetic and real data, which sometimes limits generalization ability.
As demonstrated in our experiments, our choice of synthetic datasets leads to impressive zero-shot transfer. 

% ----------------------------------
\vspace{0.35em}
\noindent\textbf{%
Annealed multi-resolution noise.
}
\begin{figure}[t!]
    \centering
    \includegraphics[width=\linewidth]{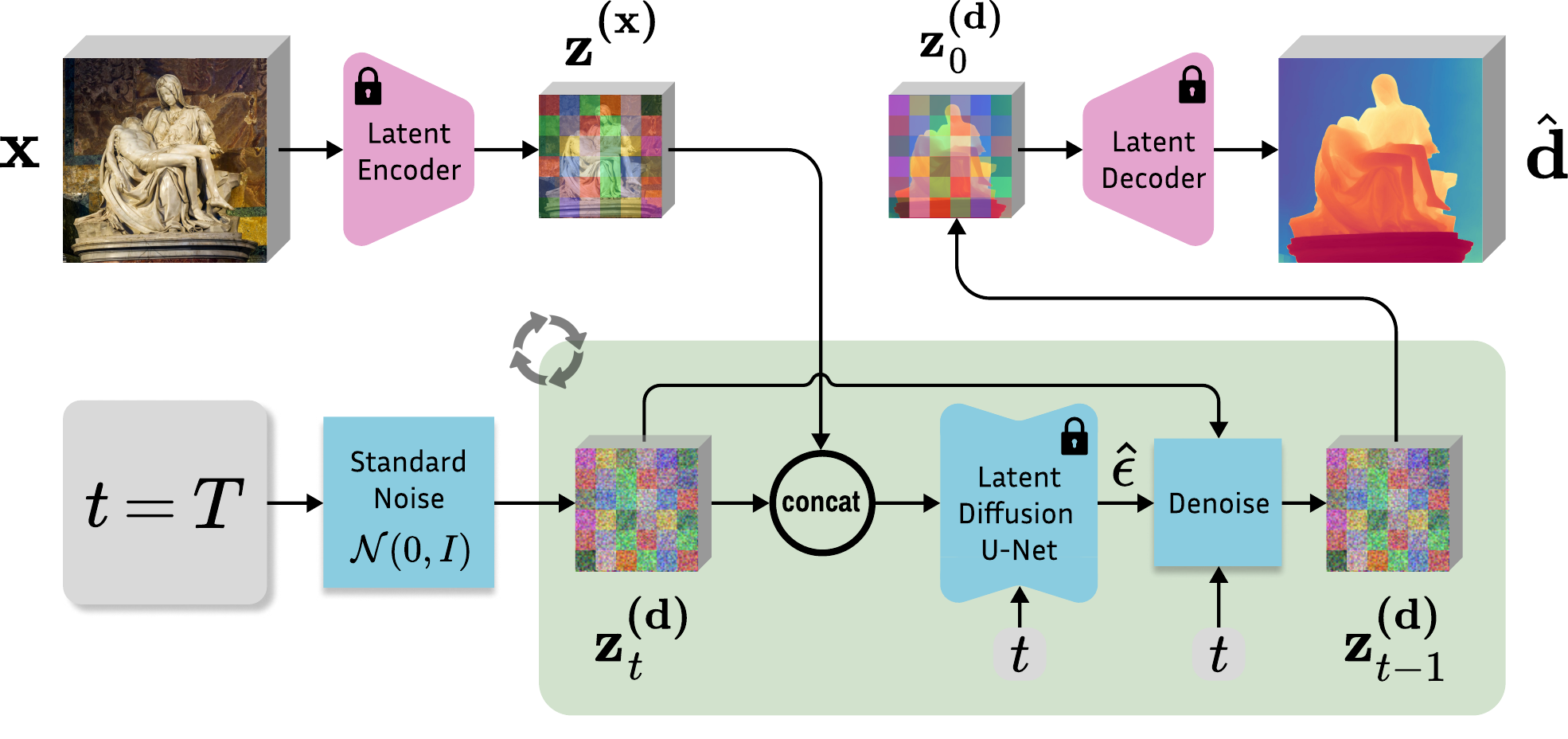}
    \caption{
    \textbf{Overview of the Marigold inference scheme.}
    Given an input image $\img$, we encode it with the original Stable Diffusion VAE into the latent code $\latentimage$, and concatenate with the depth latent $\latentdepth_t$ before giving it to the modified fine-tuned U-Net on every denoising iteration.
    After executing the schedule of $T$ steps, the resulting depth latent $\latentdepth_0$ is decoded into 
    an image, whose 3 channels are averaged to get the final estimation $\hat\depth$. 
    See \cref{sec:inference} for details.
    }
    \label{fig:method-inference}
\end{figure}
Previous works have explored deviations from the original DDPM formulations, such as non-Gaussian noise~\cite{nachmani2021non} or non-Markovian schedule shortcuts~\cite{song2020denoising_ddim}.
Our proposed setting and the fine-tuning protocol outlined above are permissive to changes to the noise schedule at the fine-tuning stage. 
We identified a combination of multi-resolution noise~\cite{whitaker2023multi-resolution}
and an annealed schedule to converge faster and substantially improve performance over the standard DDPM formulation.
The multi-resolution noise is composed by superimposing several random Gaussian noise images of different scales, all upsampled to the U-Net input resolution. 
The proposed annealed schedule interpolates between the multi-resolution noise at $t=T$ and standard Gaussian noise at $t=0$.

%%%%%%%%%%%%%%%%%%%%%%%%%%%%% Inference %%%%%%%%%%%%%%%%%%%%%%%%%%%%%
\subsection{Inference}
\label{sec:inference}

\noindent\textbf{%
Latent diffusion denoising.
}
The overall inference pipeline is presented in Fig.~\ref{fig:method-inference}.
We encode the input image into the latent space, initialize depth latent as standard Gaussian noise, and progressively denoise it with the same schedule as during fine-tuning.
We empirically find that initializing with standard Gaussian noise gives better results than with multi-resolution noise, although the model is trained on the latter.
We follow DDIM's~\cite{song2020denoising_ddim} approach to perform non-Markovian sampling with re-spaced steps for accelerated inference.
The final depth map is decoded from the latent code using the VAE decoder and postprocessed by averaging channels.

% ----------------------------------
\newcommand{\pred}{\mathbf{\hat{d}}}
\newcommand{\translated}{\mathbf{\hat{d^{\prime}}}}
\newcommand{\merged}{\mathbf{m}}

\vspace{0.35em}
\noindent\textbf{%
Test-time ensembling.
}
The stochastic nature of the inference pipeline leads to varying predictions depending on the initialization noise in $\latentdepth_T$.
Capitalizing on that, we propose the following test-time ensembling scheme, capable of combining multiple inference passes over the same input.
For each input sample, we can run inference $N$ times.
To aggregate these affine-invariant depth predictions $\{ \pred_1, \ldots, \pred_N\}$, we jointly estimate the corresponding scale $\hat{s}_i$ and shift $\hat{t}_i$, relative to some canonical scale and range, in an iterative manner.
The proposed objective minimizes the distances between each pair of scaled and shifted predictions $(\translated_i, \translated_j)$, where $\translated = \pred \times \hat{s}+ \hat{t}$. 
In each optimization step, we calculate the merged depth map $\merged$ by the taking pixel-wise median $\merged(x, y) = \text{median}(\translated_1(x, y), \ldots, \translated_N(x, y))$.
An extra regularization term $\mathcal{R} = |\! \min (\merged)| + |1 - \max (\merged)|$, is added to prevent collapse to the trivial solution and enforce the unit scale of $\merged$.
Thus, the objective function can be written as follows:
\begin{equation}
      \min_{\substack{s_1,\ldots,s_N \\ t_1, \ldots, t_N}} \Bigg( \sqrt{\frac{1}{b} \sum_{i=1}^{N-1} \sum_{j=i+1}^{N} \| \translated_i - \translated_j \|_2^2} + \lambda \mathcal{R} \Bigg)   
      \label{eq:align}
\end{equation}
where the binominal coefficient $b = \binom{N}{2}$ represents the number of possible combinations of image pairs from $N$ images.
After the iterative optimization for spatial alignment, the merged depth $\merged$ is taken as our ensembled prediction.
Note that this ensembling step requires no ground truth for aligning independent predictions.
This scheme enables a flexible trade-off between computation efficiency and prediction quality by choosing $N$ accordingly.

\section{Experiments}
\label{sec:exp}

%%%%%%%%%%%%%%%%%%%%%%%%%%%%% Implementation %%%%%%%%%%%%%%%%%%%%%%%%%%%%%
\subsection{Implementation} \label{sec:implementation}

We implement \method~using PyTorch and utilize Stable Diffusion v2~\cite{rombach2022high} as our backbone, following the original pre-training setup with a \mbox{$v$-objective}~\cite{salimans2022progressive}. 
We disable text conditioning and perform the steps outlined in Sec.~\ref{sec:architecture}.
During training, we apply the DDPM noise scheduler~\cite{ho2020denoising_ddpm} with 1000 diffusion steps. 
At inference time, we apply the DDIM scheduler~\cite{song2020denoising_ddim} and only sample 50 steps. 
For the final prediction, we aggregate results from 10 inference runs with varying starting noise. 
Training our method takes 18K iterations using a batch size of 32. 
To fit one GPU, we accumulate gradients for 16 steps. 
We use the Adam optimizer with a learning rate of $3\cdot 10^{-5}$. 
Additionally, we apply random horizontal flipping augmentation to the training data. 
Training our method to convergence takes approximately 2.5 days on a single Nvidia RTX 4090 GPU card.

% ----------------- Quantitative comparison -----------------
\begin{table*}[t!]
    \centering
    \caption{
        \textbf{Quantitative comparison}
        of \method{} with SOTA affine-invariant depth estimators on several zero-shot benchmarks.
        All metrics$^\dagger$ are presented in percentage terms; \textbf{bold} numbers are the best, \underline{underscored} second best.
        Our method outperforms other methods on both indoor and outdoor scenes in most cases, without having seen a real depth sample.
        % (but a huge number of image-text pairs during pretraining).
    }
    \resizebox{\linewidth}{!}{
	\begin{tabular}{
@{}l 
r@{}r @{}p{2.0em}@{} 
c@{\hspace{0.5em}}c @{}p{2.0em}@{} 
c@{\hspace{0.5em}}c @{}p{2.0em}@{} 
c@{\hspace{0.5em}}c @{}p{2.0em}@{} 
c@{\hspace{0.5em}}c @{}p{2.0em}@{} 
c@{\hspace{0.5em}}c @{}p{1.5em}@{} 
c@{}
}

\toprule

\multirow{2}{*}{Method} &
\multicolumn{2}{c}{\# Training samples} & &
\multicolumn{2}{c}{NYUv2} & &
\multicolumn{2}{c}{KITTI} & &
\multicolumn{2}{c}{ETH3D} & &
\multicolumn{2}{c}{ScanNet} & &
\multicolumn{2}{c}{DIODE} & &
\multirow{2}{*}{Avg. Rank} \\

& 
Real & 
Synthetic & &
AbsRel↓ & 
$\delta$1↑ & &
AbsRel↓ & 
$\delta$1↑ & &
AbsRel↓ & 
$\delta$1↑ & &
AbsRel↓ & 
$\delta$1↑ & &
AbsRel↓ & 
$\delta$1↑ & &
\\

\midrule

DiverseDepth~\cite{yin2020diversedepth} & 
320K & 
--- & & 
11.7 & 
87.5 & & 
19.0 & 
70.4 & &
22.8 & 
69.4 & &
10.9 & 
88.2 & &
37.6 & 
63.1 & &
7.6 \\

MiDaS~\cite{Ranftl2020_midas} & 
2M & 
--- & &
11.1 & 
88.5 & &
23.6 & 
63.0 & &
18.4 & 
75.2 & &
12.1 & 
84.6 & &
33.2 & 
71.5 & &
7.3 \\

LeReS~\cite{Wei2021CVPR_leres} & 
300K & 
54K & &
9.0 & 
91.6 & &
14.9 & 
78.4 & &
17.1 & 
77.7 & &
9.1 & 
91.7 & &
27.1 & 
76.6 & &
5.2 \\

Omnidata~\cite{eftekhar2021omnidata} & 
11.9M & 
310K & &
7.4 & 
94.5 & &
14.9 & 
83.5 & &
16.6 & 
77.8 & &
7.5 & 
93.6 & &
33.9 & 
74.2 & &
4.8 \\

HDN~\cite{zhang2022_hdn} & 
300K & 
--- & &
6.9 & 
94.8 & &
11.5 & 
86.7 & &
12.1 & 
83.3 & &
8.0 & 
93.9 & &
\underline{24.6} & 
\textbf{78.0} & &
3.2 \\

DPT~\cite{ranftl2021_dpt} & 
1.2M & 
188K & &
9.8 & 
90.3 & &
\underline{10.0} & 
90.1 & &
7.8 & 
94.6 & &
8.2 & 
93.4 & &
\textbf{18.2} & 
75.8 & &
3.9 \\

\midrule
Ours (w/o ensemble) & 
\multirow{2}{*}{---}{\tiny$*$} & 
\multirow{2}{*}{74K} &  &
\underline{6.0} &
\underline{95.9} & &
10.5 &
\underline{90.4} & &
\underline{7.1} & 
\underline{95.1} & &
\underline{6.9} &
\underline{94.5} & &
31.0 &
77.2 & &
\underline{2.5} \\

Ours (w/ ensemble) &
& 
& &
\textbf{5.5} & 
\textbf{96.4} & &
\textbf{9.9} & 
\textbf{91.6} & &
\textbf{6.5} & 
\textbf{96.0} & &
\textbf{6.4} & 
\textbf{95.1} & &
30.8 & 
\underline{77.3} & &
\textbf{1.4} \\

\bottomrule

\end{tabular}

	\label{table:zeroshot_test}
    }
    \\
    \begin{minipage}{0.96\linewidth}
        \scriptsize
        \vspace{0.4em}
        \begin{itemize}
        \item[$^\dagger$]
        Most baselines are sourced from Metric3D~\cite{yin2023metric3d}, except for the ScanNet benchmark. 
        For ScanNet, Metric3D used a different random split that is not publicly accessible, therefore we re-ran all baselines on our split. 
        For HDN~\cite{zhang2022_hdn} we show the ScanNet results from Metric3D, as no source code is available. 
        \item[$^*$] Image-text data is used in the pretrained model.  
        \end{itemize}
    \end{minipage}
\end{table*}

%%%%%%%%%%%%%%%%%%%%%%%%%%%%% Evaluation %%%%%%%%%%%%%%%%%%%%%%%%%%%%%
\subsection{Evaluation} 
\label{sec:evaluation}

\noindent\textbf{%
Training datasets.
} 
We train \method~on two synthetic datasets covering both indoor and outdoor scenes.
\textbf{Hypersim}~\cite{roberts2021hypersim} is a photorealistic dataset with 461 indoor scenes. 
We use the official split with around 54K samples from 365 scenes for training. 
Incomplete samples are filtered out.
RGB images and depth maps are resized to $480 \times 640$ size.
Additionally, we transform the original distances relative to the focal point into conventional depth values relative to the focal plane. 
The second dataset, \textbf{Virtual KITTI}~\cite{cabon2020virtualkitti2} is a synthetic street-scene dataset featuring 5 scenes under varying conditions like weather or camera perspectives. 
Four scenes containing a total of around 20K samples are used for training. 
We crop the images to the KITTI benchmark resolution~\cite{Geiger2012CVPR} and set the far plane to 80 meters.

% ----------------- -----------------
\vspace{0.35em}
\noindent\textbf{%
Evaluation datasets.
}
We evaluate \method{} on 5 real datasets that are not seen during training. 
\textbf{NYUv2}~\cite{SilbermanECCV12nyu} and \textbf{ScanNet}~\cite{dai2017scannet} are both indoor scene datasets captured with an RGB-D Kinect sensor. 
For NYUv2, we utilize the designated test split, comprising a total of 654 images. 
In the case of the ScanNet dataset, we randomly sampled 800 images from the 312 official validation scenes for testing.  
\textbf{KITTI}~\cite{Geiger2012CVPR} is a street-scene dataset with sparse metric depth captured by a LiDAR sensor. 
We employ the Eigen test split~\cite{eigen_depth_2014} made of 652 images.
\textbf{ETH3D}~\cite{schops2017multiEth3d} and \textbf{DIODE}~\cite{diode_dataset} are two high-resolution datasets, both featuring depth maps derived from LiDAR sensor measurements. 
For ETH3D, we incorporate all 454 samples with available ground truth depth maps. 
For DIODE, we use the entire validation split, which encompasses 325 indoor samples and 446 outdoor samples.

% ----------------- -----------------
\vspace{0.35em}
\noindent\textbf{%
Evaluation protocol.
} 
Following the protocol of affine-invariant depth evaluation~\cite{Ranftl2020_midas}, we first align the estimated merged prediction $\merged$ to the ground truth $\depth$ with the least squares fitting. 
This step gives us the absolute aligned depth map $\mathbf{a} = \merged \times s + t$ in the same units as the ground truth.
Next, we apply two widely recognized metrics~\cite{yin2023metric3d, Ranftl2020_midas, ranftl2021_dpt, Wei2021CVPR_leres} for assessing quality of depth estimation.
The first is Absolute Mean Relative Error (\textit{AbsRel}), calculated as: $\frac{1}{M} \sum_{i=1}^M {|\mathbf{a}_i - \depth_i|} / {\depth_i}$, where $M$ is the total number of pixels. 
The second metric, $\delta1$ accuracy, measures the proportion of pixels satisfying $\max({\mathbf{a}_i}/{\depth_i}, {\depth_i}/{\mathbf{a}_i}) < 1.25$.

% ----------------- -----------------
\vspace{0.35em}
\noindent\textbf{%
Comparison with other methods.
}
We compare \method{} to six baselines, each claiming zero-shot generalization.
DiverseDepth~\cite{yin2020diversedepth}, LeReS~\cite{Wei2021CVPR_leres} and HDN~\cite{zhang2022_hdn} estimate affine-invariant depth maps, while MiDaS~\cite{Ranftl2020_midas}, DPT~\cite{ranftl2021_dpt}, and Omnidata~\cite{eftekhar2021omnidata} produce affine-invariant disparities. 
As shown in \cref{table:zeroshot_test}, \method{} outperforms prior art in most cases and secures the highest overall ranking. 
Despite being trained solely on synthetic depth datasets, the model can well generalize to a wide range of real scenes.
This successful adaptation of diffusion-based image generation models toward depth estimation confirms our initial hypothesis that a comprehensive representation of the visual world is the cornerstone of monocular depth estimation. It also shows that our fine-tuning protocol was successful in adapting Stable Diffusion for this task without unlearning such visual priors.

\begin{figure*}[t]
    \centering
    \setlength\tabcolsep{0.5pt}

\newcommand{\sampleNameSingle}[1]{
    \multirow{1}{*}[1.3cm]{\rotatebox{90}{#1}}
}

\newcommand{\viscolumn}{0.166\textwidth}

\resizebox{\linewidth}{!}{
    \begin{tabular}[ht]{r p{\viscolumn} p{\viscolumn} p{\viscolumn} p{\viscolumn} p{\viscolumn} p{\viscolumn}}
        & \parbox[c]{\viscolumn}{\centering Input RGB Image} 
        & \parbox[c]{\viscolumn}{\centering MiDaS} 
        & \parbox[c]{\viscolumn}{\centering Omnidata} 
        & \parbox[c]{\viscolumn}{\centering DPT} 
        & \parbox[c]{\viscolumn}{\centering Marigold (ours)} 
        & \parbox[c]{\viscolumn}{\centering Ground Truth} \\ 

        \sampleNameSingle{NYUv2}
        & \multicolumn{6}{c}{\includegraphics[width=\textwidth]{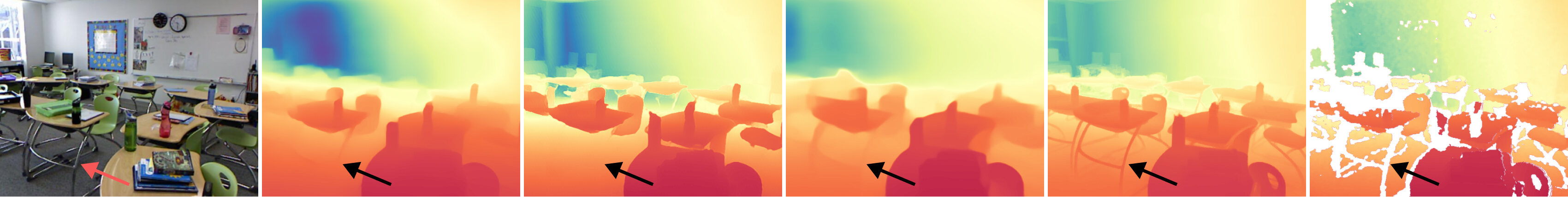}} 
        \\
        \sampleNameSingle{KITTI}
        & \multicolumn{6}{c}{\includegraphics[width=\textwidth]{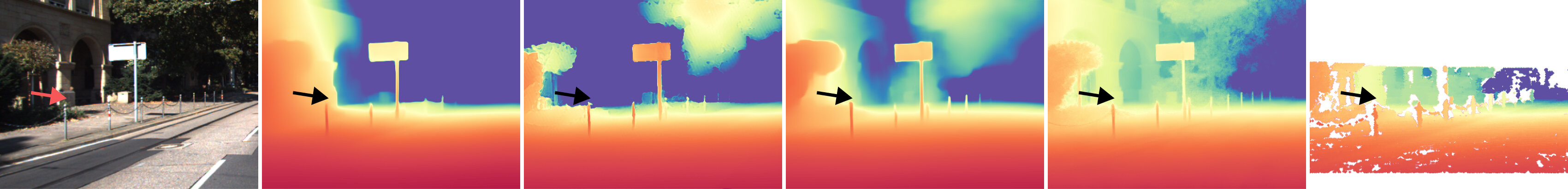}} 
        \\
        \sampleNameSingle{ETH3D}
        & \multicolumn{6}{c}{\includegraphics[width=\textwidth]{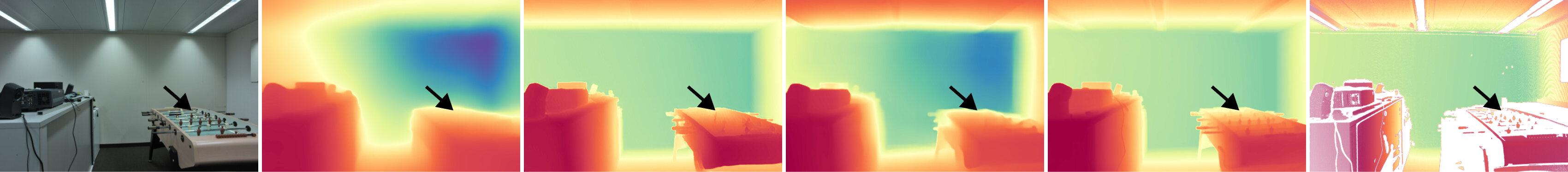}} 
        \\
        \sampleNameSingle{Scannet}
        & \multicolumn{6}{c}{\includegraphics[width=\textwidth]{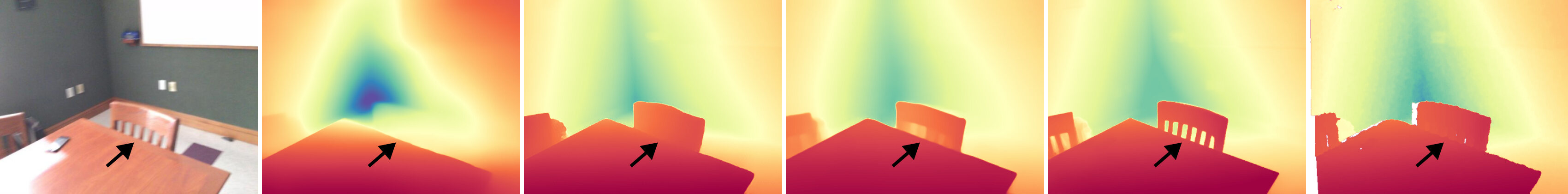}} 
        \\
        \sampleNameSingle{DIODE}
        & \multicolumn{6}{c}{\includegraphics[width=\textwidth]{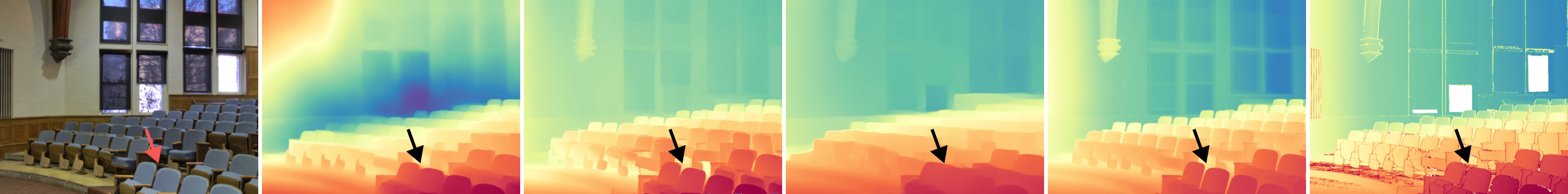}} 
    \end{tabular}
}

    \caption{
        \textbf{Qualitative comparison (depth)}
        of monocular depth estimation methods across different datasets. 
        \method{} excels at capturing thin structures~(\eg, chair legs) and preserving overall layout of the scene~(\eg, walls in ETH3D example and chairs in DIODE example). 
    }
    \label{fig:depth_comparison}
\end{figure*}

\begin{figure*}[t]
    \centering
    \setlength\tabcolsep{0.5pt}

\newcommand{\sampleNameSingle}[1]{
    \multirow{1}{*}[1.3cm]{\rotatebox{90}{#1}}
}

\newcommand{\viscolumn}{0.166\textwidth}

\resizebox{\linewidth}{!}{
    \begin{tabular}[ht]{r p{\viscolumn} p{\viscolumn} p{\viscolumn} p{\viscolumn} p{\viscolumn} p{\viscolumn}}
        & \parbox[c]{\viscolumn}{\centering Input RGB Image} 
        & \parbox[c]{\viscolumn}{\centering MiDaS} 
        & \parbox[c]{\viscolumn}{\centering Omnidata} 
        & \parbox[c]{\viscolumn}{\centering DPT} 
        & \parbox[c]{\viscolumn}{\centering Marigold (ours)} 
        & \parbox[c]{\viscolumn}{\centering Ground Truth} \\ 
        \sampleNameSingle{NYUv2}
        & \multicolumn{6}{c}{\includegraphics[width=\textwidth]{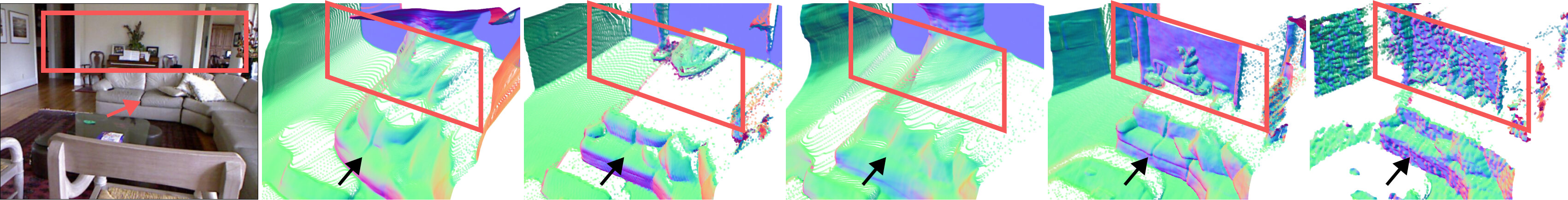}} 
        \\
        \sampleNameSingle{ScanNet}
        & \multicolumn{6}{c}{\includegraphics[width=\textwidth]{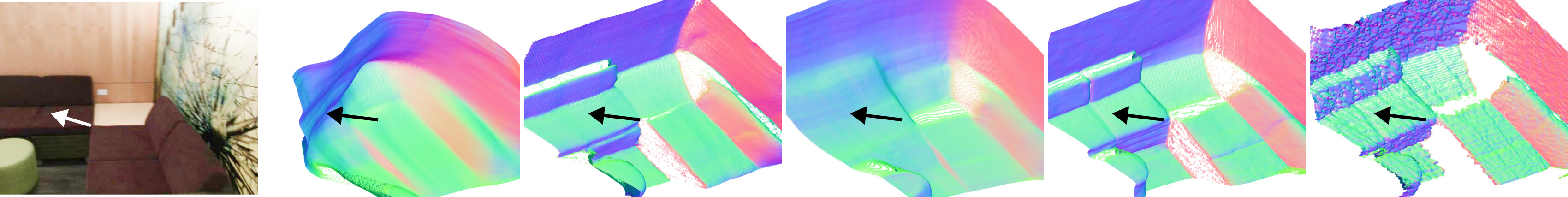}} 
        \\
        \sampleNameSingle{DIODE}
        & \multicolumn{6}{c}{\includegraphics[width=\textwidth]{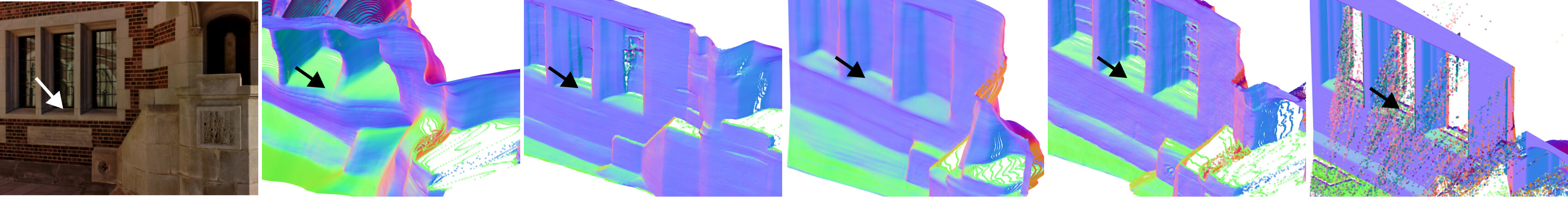}} 
    \end{tabular}
}

    \caption{
        \textbf{Qualitative comparison (unprojected, colored as normals)}
        of monocular depth estimation methods across different datasets. 
        \method{} stands out for its superior reconstruction of flat surfaces and detailed structures. 
    }
    \label{fig:normal_comparison3d}
\end{figure*}

For a visual assessment, we present qualitative comparison in \cref{fig:depth_comparison}. Additionally, in~\cref{fig:normal_comparison3d}, we provide 3D visualizations of reconstructed surface normals. 
\method{} not only correctly captures the scene layout, such as the spatial relationships between walls and furniture in the first example in \cref{fig:normal_comparison3d}, but also captures fine-grained details, as indicated by the arrows in \cref{fig:depth_comparison}.
Moreover, the reconstruction of flat surfaces, especially walls, is significantly better (see \cref{fig:depth_comparison}).
Furthermore, our method effectively models common shapes and their layouts, once again aligning with our expectations regarding the generative prior.

% ----------------- Qualitative comparison (depth) -----------------

% ----------------- Qualitative comparison (normal) -----------------

%%%%%%%%%%%%%%%%%%%%%%%%%%%%% Ablation Studies %%%%%%%%%%%%%%%%%%%%%%%%%%%%%
\subsection{Ablation Studies}
\label{sec:ablation}

Two zero-shot validation sets are selected for the ablation studies -- the official training split of NYUv2~\cite{SilbermanECCV12nyu}, consisting of 785 samples, and a randomly selected subset of 800 images from the KITTI Eigen~\cite{eigen_depth_2014} training split.
Refer to supplementary sections for extra ablations and discussion.

\begin{figure}[t!]
    \centering
    \includegraphics[width=0.99\linewidth]{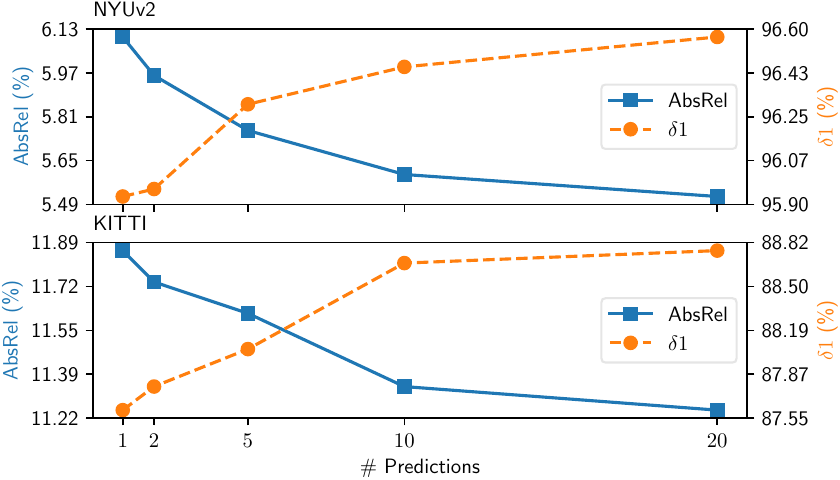}
    \vspace{-0.4em}
    \caption{
        \textbf{Ablation of ensemble size.} 
        We observe a monotonic improvement with the growth of ensemble size. This improvement starts to diminish after 10 predictions per sample. 
    }
   \label{fig:merge_pred}
\end{figure}

\begin{table}[t]
	\centering
	\caption{
		\textbf{Ablation of training noise.}
            Multi-resolution noise improves over Gaussian noise; annealing yields further improvement.
	}
    \vspace{-0.1em}
    \resizebox{!}{0.12\linewidth}{
        \begin{tabular}{cc c cccc}

\toprule

\multicolumn{1}{c}{\multirow{2}{*}{\makecell{Multi-res. \\ noise}}} & 
\multicolumn{1}{c}{\multirow{2}{*}{Annealed}} & 
& 
\multicolumn{2}{c}{NYUv2} & 
\multicolumn{2}{c}{KITTI} \\

\multicolumn{1}{c}{} & 
\multicolumn{1}{c}{} & 
& 
AbsRel↓ & 
$\delta$1↑ & 
AbsRel↓ & 
$\delta$1↑ \\

\midrule      

\xmark & -      & & 7.7          & 93.4          & 14.2           & 82.1            \\
\cmark & \xmark & & 5.8          & 96.1          & 12.1           & 87.1            \\
\cmark & \cmark & & \textbf{5.6} & \textbf{96.5} & \textbf{11.3 } & \textbf{88.7  } \\

\bottomrule

\end{tabular}
	}
    \vspace{-0.4em}
	\label{table:anneal}
\end{table}

\vspace{0.35em}
\noindent\textbf{%
Training noise.
}
We investigate the impact of three types of noise during the training phase. As shown in~\cref{table:anneal}, training with multi-resolution noise significantly improves the depth prediction accuracy over using standard Gaussian noise.
Furthermore, the gradual annealing of multi-resolution noise yields an additional improvement. 
We also noticed that training with multi-resolution noise leads to more consistent predictions given different initial noise at inference time and annealing further enhances this consistency. 

\vspace{0.35em}
\noindent\textbf{%
Training data domain.
}
To better understand the impact of the synthetic datasets used for our fine-tuning protocol, we ablate on a photorealistic street-scene Virtual KITTI~\cite{cabon2020virtualkitti2}, and a more diverse and higher-quality indoor dataset Hypersim~\cite{roberts2021hypersim}. 
The results are shown in~\cref{tbl:train_domain}. 
When fine-tuned on a single synthetic dataset, the pretrained LDM can already be adapted for monocular depth estimation to a certain degree, while the more diverse and photorealistic data leads to better performance on both indoor and outdoor scenes.
Interestingly, adding additional training data from a different domain not only improves the performance on the new domain but also brings improvements in the original domain.

\vspace{0.35em}
\noindent\textbf{%
Test-time ensembling.
}
We test the effectiveness of the proposed test-time ensembling scheme by aggregating various numbers of predictions.
As shown in~\cref{fig:merge_pred}, a single prediction of \method{} already yields reasonably good results. Ensembling 10 predictions reduces the absolute relative error on NYUv2 by ${\approx}8\%$ and ensembling 20 predictions brings an improvement of ${\approx}9.5\%$.
It has been observed as a systematic effect that the performance is constantly improved as the number of predictions increases, while the marginal improvement diminishes with more than 10 predictions.

\begin{figure}[t!]
    \centering
    \includegraphics[width=0.99\linewidth]{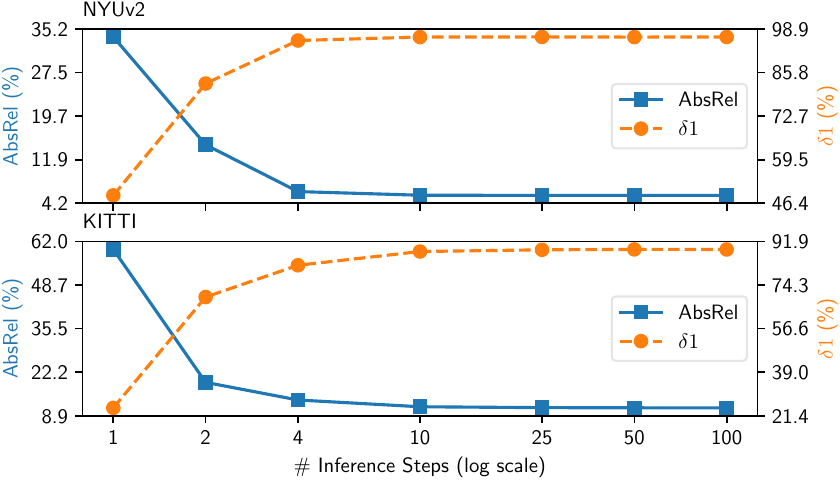}
    \vspace{-0.4em}
    \caption{
        \textbf{Ablation of denoising steps.}
        The performance improves as the number of denoising steps increases, while we observe saturation after 10 steps.
    }
   \label{fig:inference_denoise_step}
\end{figure}

\begin{table}[t]
    \centering
    \caption{\textbf{Ablation of training datasets.} 
        Hypersim~\cite{roberts2021hypersim} delivers strong results;
        Virtual KITTI~\cite{cabon2020virtualkitti2} improves outdoor performance.
    }
    \vspace{-0.1em}
    \resizebox{!}{0.12\linewidth}{
        \begin{tabular}{@{}cc c cccc@{}}

\toprule

\multicolumn{1}{c}{\multirow{2}{*}{\makecell{Hypersim}}} & 
\multicolumn{1}{c}{\multirow{2}{*}{\makecell{Virtual \\KITTI}}} & 
& 
\multicolumn{2}{c}{NYUv2} & 
\multicolumn{2}{c}{KITTI} \\

\multicolumn{1}{c}{} & 
\multicolumn{1}{c}{} & 
& 
AbsRel↓ & 
$\delta$1↑ & 
AbsRel↓ & 
$\delta$1↑ \\

\midrule      

\xmark & \cmark &  & 13.9         & 83.4          & 15.4          & 79.3          \\
\cmark & \xmark &  & 5.7          & 96.3          & 13.7          & 82.5          \\
\cmark & \cmark &  & \textbf{5.6} & \textbf{96.5} & \textbf{11.3} & \textbf{88.7} \\

\bottomrule

\end{tabular}%
    }
    \vspace{-0.4em}
    \label{tbl:train_domain}
\end{table}

\vspace{0.35em}
\noindent\textbf{%
Number of denoising steps.
}
We evaluate the effect of the re-spaced inference denoising steps driven by the DDIM scheduler~\cite{song2020denoising_ddim}. The results are shown in~\cref{fig:inference_denoise_step}. 
Although trained with 1000 DDPM steps, the choice of 50 steps is sufficient to produce accurate results during inference.
As expected, we obtain better results when using more denoising steps. 
We observe that the elbow point of marginal returns given more denoising steps depends on the dataset but is always under 10 steps.
This implies that one can further reduce the denoising steps to 10 or even less to gain efficiency while keeping comparable performance.
Interestingly, this threshold is smaller than what is usually required for diffusion-based image generators~\cite{song2020denoising_ddim, rombach2022high}, \ie, 50 steps.

\section{Conclusion}
\label{sec:conclusion}

We have presented \method{}, a fine-tuning protocol for Stable Diffusion and a model for state-of-the-art affine-invariant depth estimation.
Our results confirm the importance of a detailed visual scene understanding prior for depth estimation, which we source from the pretrained text-to-image diffusion model.
Future research directions to overcome current limitations include improving inference efficiency, ensuring that similar inputs yield consistent outputs despite the model's generative nature, and better handling of distant scene parts.

{
    \small
    \clearpage
    \bibliographystyle{ieeenat_fullname}
    \bibliography{main}
}

% Supplementary material
\vspace{5mm}
\appendix
\section*{\Large Appendix}

\renewcommand*{\thesection}{\Alph{section}}
\newcommand{\multiref}[2]{\cref{#1}--\ref{#2}}
\renewcommand{\thetable}{S\arabic{table}}
\renewcommand{\thefigure}{S\arabic{figure}}

\setcounter{table}{0}
\setcounter{figure}{0}

\noindent
In this supplementary material, we provide additional implementation details in~\cref{sec:supp_implementation} and present additional quantitative and qualitative results in~\cref{sec:supp_quan} and~\cref{sec:supp_qualitative}, respectively. 

%%%%%%%%%%%%%%%%%%%%%%%%%%%%% Implementation Details %%%%%%%%%%%%%%%%%%%%%%%%%%%%%
\section{Implementation Details}\label{sec:supp_implementation}

\subsection{Mixed Dataset Training}
\label{sec:mixing}
We train on two synthetic datasets, Hypersim~\cite{roberts2021hypersim} and Virtual KITTI~\cite{cabon2020virtualkitti2}, whose images have different resolutions and aspect ratios.
For each batch, we probabilistically choose the dataset and then draw samples from it. 
We ablate the Bernoulli parameter of dataset sampling in \cref{sec:vkitti_ratio}.

% ----------------------------------
\subsection{Annealed Multi-Resolution Noise}
In the standard multi-resolution noise, multiple Gaussian noise images are sampled to form a pyramid of resolutions and then subsequently combined by upsampling, weighted averaging, and renormalization.
The weight for the $i$-th pyramid level is computed as $s^i$, where $0<s<1$ is a strength of influence of lower-resolution noise.
To bring such noise closer to the Gaussian used in the original DDPM formulation, we propose to anneal the weight of levels $i > 0$ based on the diffusion schedule.
Specifically, we assign the $i$-th level at timestep $t$ the weight $(st / T)^i$, where $T$ is the total number of diffusion steps.
Thus, a smaller weight is given to lower-resolution levels at timesteps closer to the noise-free end of the schedule. 
In addition to the ablation study in the main paper, we further demonstrate the effectiveness of annealing and other noise settings in \cref{sec:abl:noise}.

% ----------------------------------
\subsection{Alignment with Ground Truth Depth}
Following the established evaluation protocol~\cite{Ranftl2020_midas}, we use least squares fitting over pixels with valid ground truth values to compute the scale and shift factors of the affine-invariant predictions. 
Note that, while some methods predict affine-invariant disparities~\cite{Ranftl2020_midas, ranftl2021_dpt, eftekhar2021omnidata}, others (including ours) predict affine-invariant depth values~\cite{yin2020diversedepth, Wei2021CVPR_leres, zhang2022_hdn}. 
We apply least squares fitting accordingly, \ie the disparities are aligned to the inverse ground truth depth.

% ----------------------------------
\subsection{Visualization in 3D}

We compute the scale and shift scalars between the prediction and ground truth. 
Subsequently, we unproject pixels into the metric 3D space using the camera intrinsics. 
We manually estimate the scale, shift, and intrinsics of ``in-the-wild'' samples, where ground truth and camera intrinsics are unavailable. 
For some samples, camera intrinsics can also be extracted from the EXIF metadata.
To visualize normals, we perform least squares plane fitting at each position, considering a neighborhood area of $3 \times 3$ pixels around it.

%%%%%%%%%%%%%%%%%%%%%%%%%%%%% Experimental Results %%%%%%%%%%%%%%%%%%%%%%%%%%%%%
\section{Experimental Results}\label{sec:supp_quan}

\subsection{Stable Diffusion VAE with Depth} 
\label{sec:encoder-decoder}
To assess how well the pre-trained image variational autoencoder of Stable Diffusion~\cite{rombach2022high} works with depth maps, we tested it with 800 samples from the Hypersim~\cite{roberts2021hypersim} training set. 
To this end, each sample is normalized to the operational range of VAE as explained in the main paper, and replicated three times to accommodate the RGB interface.
Upon decoding the latents, the reconstructed depth map is derived by averaging the three RGB channels. 
Over the chosen set of depth maps, the Mean Absolute Error (MAE) of reconstructions is $0.0095 \pm 0.0091$,  
which is safely below the current state-of-the-art depth estimation errors.

% ----------------------------------
\subsection{Consistency of Channels After VAE Decoder}
To further understand the suitability of the Stable Diffusion latent space for depth representation,
we evaluate the agreement of depth channels obtained from the VAE decoder during inference. 
We validate with the training split of NYUv2~\cite{SilbermanECCV12nyu} and a subsampled Eigen training split~\cite{eigen_depth_2014} of the KITTI dataset~\cite{Geiger2012CVPR}.
As shown in \cref{tbl:channel_diff}, the channel-wise discrepancy resulting from decoding depth from the latent space is small relative to the value range of the decoder output, \ie, $[-1, 1]$. 
This could be related to the ability of VAE to represent gray-scale RGB images.

\begin{table}[ht]
    \centering
    \caption{
    \textbf{Consistency of channels after VAE decoder.} 
    The reported numbers are averaged over the respective datasets.
    }
    \resizebox{0.5\linewidth}{!}{
        \begin{tabular}{lcc}

\toprule

& std & $\text{max} - \text{min}$ \\

\midrule

NYU   & 0.0027 & 0.0062 \\
KITTI & 0.0022 & 0.0052 \\

\bottomrule

\end{tabular}

    }
    \label{tbl:channel_diff}
\end{table}

% ----------------------------------
\subsection{Prediction Variance and Training Noise}
\label{sec:abl:noise}
\begin{figure}[ht]
    \centering
    \includegraphics[width=\linewidth]{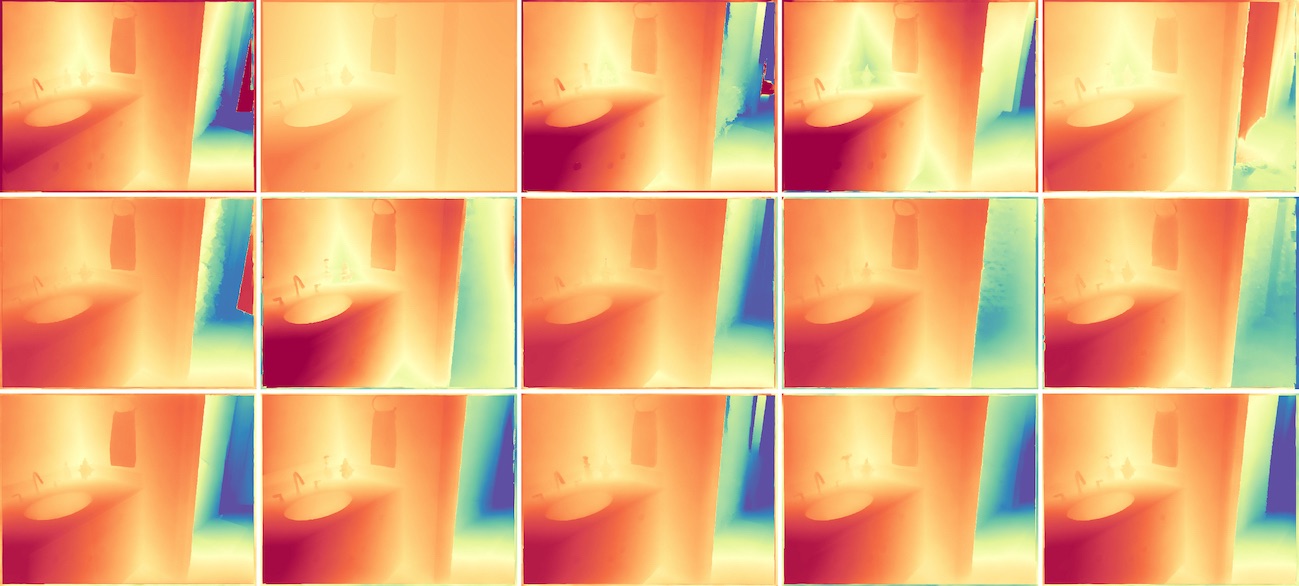}
    \caption{
        \textbf{Example of predictions on the same input} by models trained with (top-down) Gaussian, multi-resolution, and annealed multi-resolution noise.
        The last row exhibits the least variance.
    }
    \label{fig:pred_consistency}
\end{figure}

Since \method{} is a generative model, the predictions vary depending on the initial noise starting the diffusion process. 
We evaluate the consistency of predictions of three models, trained differently, \ie, with Gaussian noise, multi-resolution noise, and annealed multi-resolution noise. 
We train with two synthetic datasets and validate with the training split of NYUv2~\cite{SilbermanECCV12nyu} and a subsampled Eigen training split~\cite{eigen_depth_2014} of the KITTI dataset~\cite{Geiger2012CVPR}.
Specifically, we perform inference 10 times for each sample and compute pixel-wise statistics over the resulting depth predictions.
Subsequently, we aggregate these statistics across entire datasets and report them in \cref{tbl:consistency}.
As seen from the values, training with the multi-resolution noises increases the prediction consistency at inference, and the annealed version brings further improvement.
\cref{fig:pred_consistency} demonstrates predictions for a single sample with three models and varying starting noise.

\begin{table}[ht]
    \centering
    \caption{
    \textbf{Pixel-wise consistency of depth predictions made by models trained with three different noise types.} 
    The reported numbers are averaged over entire datasets, wherein each sample was processed 10 times, starting from a new noise sample.
    }
    \resizebox{1.0\linewidth}{!}{
        \begin{tabular}{@{}cc c cccc@{}}

\toprule

\multicolumn{1}{c}{\multirow{2}{*}{\makecell{Multi-res. \\ noise}}} & 
\multicolumn{1}{c}{\multirow{2}{*}{Annealed}} & 
& 
\multicolumn{2}{c}{NYUv2}    & 
\multicolumn{2}{c}{KITTI} \\

\multicolumn{1}{c}{} & 
\multicolumn{1}{c}{} &
& 
std &  
$\text{max} - \text{min}$ & 
std & 
$\text{max} - \text{min}$ \\

\midrule      

\xmark & \xmark & & 0.086 & 0.260 & 0.050 & 0.152 \\
\cmark & \xmark & & 0.037 & 0.117 & 0.030 & 0.094 \\
\cmark & \cmark & & 0.033 & 0.106 & 0.025 & 0.079 \\

\bottomrule

\end{tabular}%
    }
    \label{tbl:consistency}
\end{table}

\vspace{-2mm}

% ----------------------------------
\subsection{Ratio of Mixed Training Datasets} \label{sec:vkitti_ratio}
To further investigate the impact of the synthetic datasets used in our fine-tuning protocol, we ablate the mixing ratio of the datasets, discussed in \cref{sec:mixing}.
We train with two synthetic datasets, Hypersim~\cite{roberts2021hypersim} and Virtual KITTI~\cite{cabon2020virtualkitti2}, and validate with the training split of NYUv2~\cite{SilbermanECCV12nyu} and a subsampled Eigen training split~\cite{eigen_depth_2014} of the KITTI dataset~\cite{Geiger2012CVPR}.
As shown in \cref{tbl:vkitti_ratio}, training with a mixture of these two synthetic datasets yields better results on both indoor and outdoor real datasets, than training with a single synthetic dataset. 
Interestingly, based on the higher-quality indoor dataset, Hypersim~\cite{roberts2021hypersim}, adding a small portion (5\%) of Virtual KITTI~\cite{cabon2020virtualkitti2}, a street-view dataset, can already increase the performance on the outdoor dataset. 
We find a sweet spot at around 10\% where the performance is improved on both indoor and outdoor scenes. 
When the ratio of Virtual KITTI keeps increasing, the overall performance is impaired. 
This is likely caused by the varying scene diversity and rendering quality of these two datasets.

\begin{table}[ht]
    \centering
    \caption{\textbf{Ablation study of the training dataset mixing strategy.} 
    Our method trained with only Hypersim delivers strong results. 
    Outdoor performance is further enhanced with a small portion of Virtual KITTI. 
    The zero-shot transfer is attained at 10\% ratio.
    }
    \resizebox{0.9\linewidth}{!}{
        \begin{tabular}{@{}cc c cccc@{}}

\toprule

\multicolumn{1}{c}{\multirow{2}{*}{\makecell{Hypersim}}} & 
\multicolumn{1}{c}{\multirow{2}{*}{\makecell{Virtual \\KITTI}}} &
& 
\multicolumn{2}{c}{NYUv2} & 
\multicolumn{2}{c}{KITTI} \\

\multicolumn{1}{c}{} & \multicolumn{1}{c}{} && AbsRel↓ & $\delta$1↑ & AbsRel↓ & $\delta$1↑ \\

\midrule      

100\% & 0\%   & & 5.7  & 96.3 & 13.7 & 82.5 \\
95\%  & 5\%   & & 5.8  & 96.2 & 11.1 & 88.8 \\
90\%  & 10\%  & & 5.6  & 96.5 & 11.3 & 88.7 \\
50\%  & 50\%  & & 6.0  & 96.0 & 12.8 & 85.5 \\
0\%   & 100\% & & 13.9 & 83.4 & 15.4 & 79.3 \\

\bottomrule

\end{tabular}%
    }
    \label{tbl:vkitti_ratio}
\end{table}

\subsection{Inference Speed} \label{sec:inference_speed}
\begin{figure}[h!]
    \centering
    \includegraphics[width=\linewidth]{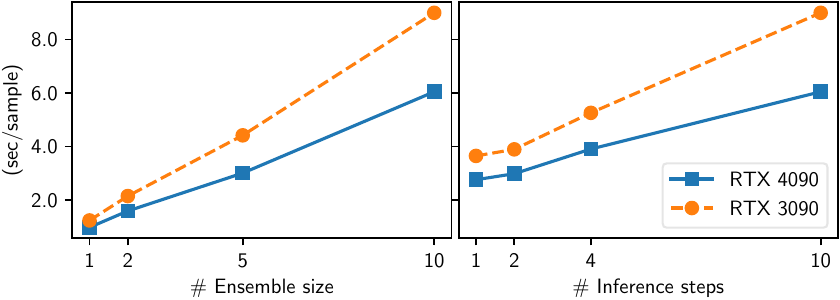}
    \caption{Inference speed on a single GPU with NYUv2 dataset.}
    \label{fig:inference_speed}
\end{figure}

In \cref{fig:inference_speed}, we report inference runtime, aligned with the settings from Figs.~6,~7.
We acknowledge the slower speed \vs higher quality trade-off compared to feed-forward methods. 
Speed can be enhanced in future research, \eg distillation for 2- or 4-step denoising schedules, and reducing prediction variance for smaller ensemble sizes.

%%%%%%%%%%%%%%%%%%%%%%%%%%%%% Qualitative Comparisons %%%%%%%%%%%%%%%%%%%%%%%%%%%%%
\section{Qualitative Comparisons}\label{sec:supp_qualitative}

\subsection{In-the-Wild}
We present the gallery of ``in-the-wild'' images and corresponding predictions in \cref{fig:in-the-wild}. 
The input images are taken in daily life or downloaded from the internet. 
Our method, \method{}, predicts accurate depth maps, exhibiting better overall layout and fine details.
We show the final predictions for each method, that is, depth for \method{} and LeReS, and disparity for MiDaS.

\subsection{Test Datasets}
We show additional qualitative comparisons with our competitors~\cite{yin2020diversedepth, Ranftl2020_midas, Wei2021CVPR_leres, ranftl2021_dpt, eftekhar2021omnidata}, on 5 test datasets~\cite{SilbermanECCV12nyu, Geiger2012CVPR, schops2017multiEth3d, dai2017scannet, diode_dataset}. 
The depth maps are visualized in \cref{fig:supp_qualitative_2d}, and the normal maps can be found in \cref{fig:supp_qualitative_normal}. 
\method{} excels at capturing fine scene details and reflecting the global scene layout. 

% ----------------- in-the-wild -----------------
\newcommand{\sampleNameMulti}[2]{
    \multirow{#2}{*}{\rotatebox[origin=c]{90}{#1}}
}

\newcommand{\viscolumn}{0.232\textwidth}

\newcommand{\upperhead}{
    \parbox[c]{\viscolumn}{\centering Input RGB Image} 
    & \parbox[c]{\viscolumn}{\centering DiverseDepth}
    & \parbox[c]{\viscolumn}{\centering MiDaS}
    & \parbox[c]{\viscolumn}{\centering LeReS}
    }
\newcommand{\lowerhead}{
    \parbox[c]{\viscolumn}{\centering Ground Truth} 
    & \parbox[c]{\viscolumn}{\centering \method{} (ours)}
    & \parbox[c]{\viscolumn}{\centering DPT}
    & \parbox[c]{\viscolumn}{\centering Omnidata}
    }
    
\newcommand{\lowerheadnormal}{
    \parbox[t]{\viscolumn}{\centering Ground Truth} 
    & \parbox[t]{\viscolumn}{\centering \method{} (ours)}
    & \parbox[t]{\viscolumn}{\centering DPT}
    & \parbox[t]{\viscolumn}{\centering Omnidata}
    }

\newcommand{\itwviscolumn}{0.21\textwidth}

\newcommand{\inthewildheader}{
    \parbox[c]{\itwviscolumn}{\centering Input RGB Image}
    & \parbox[c]{\itwviscolumn}{\centering \method{} (ours, depth)}
    & \parbox[c]{\itwviscolumn}{\centering LeReS (depth)}
    & \parbox[c]{\itwviscolumn}{\centering MiDaS (disparity)}
}

\newcommand{\nyu}{NYUv2~\cite{SilbermanECCV12nyu}}
\newcommand{\kitti}{KITTI~\cite{Geiger2012CVPR}}
\newcommand{\eththreed}{ETH3D~\cite{schops2017multiEth3d}}
\newcommand{\scannet}{ScanNet~\cite{dai2017scannet}}
\newcommand{\diode}{DIODE~\cite{diode_dataset}}
% Define a new command for dataset labels if there is only one sample
% \newcommand{\sampleNameSingle}[1]{
%     \multirow{1}{*}[1.3cm]{\rotatebox{90}{#1}}
% }

%\resizebox{\textwidth}{!}{
    \begin{tabular}[p]{ p{\itwviscolumn} p{\itwviscolumn} p{\itwviscolumn} p{\itwviscolumn} }
        \inthewildheader \\
        \multicolumn{4}{c}{\includegraphics[width=0.92\textwidth]{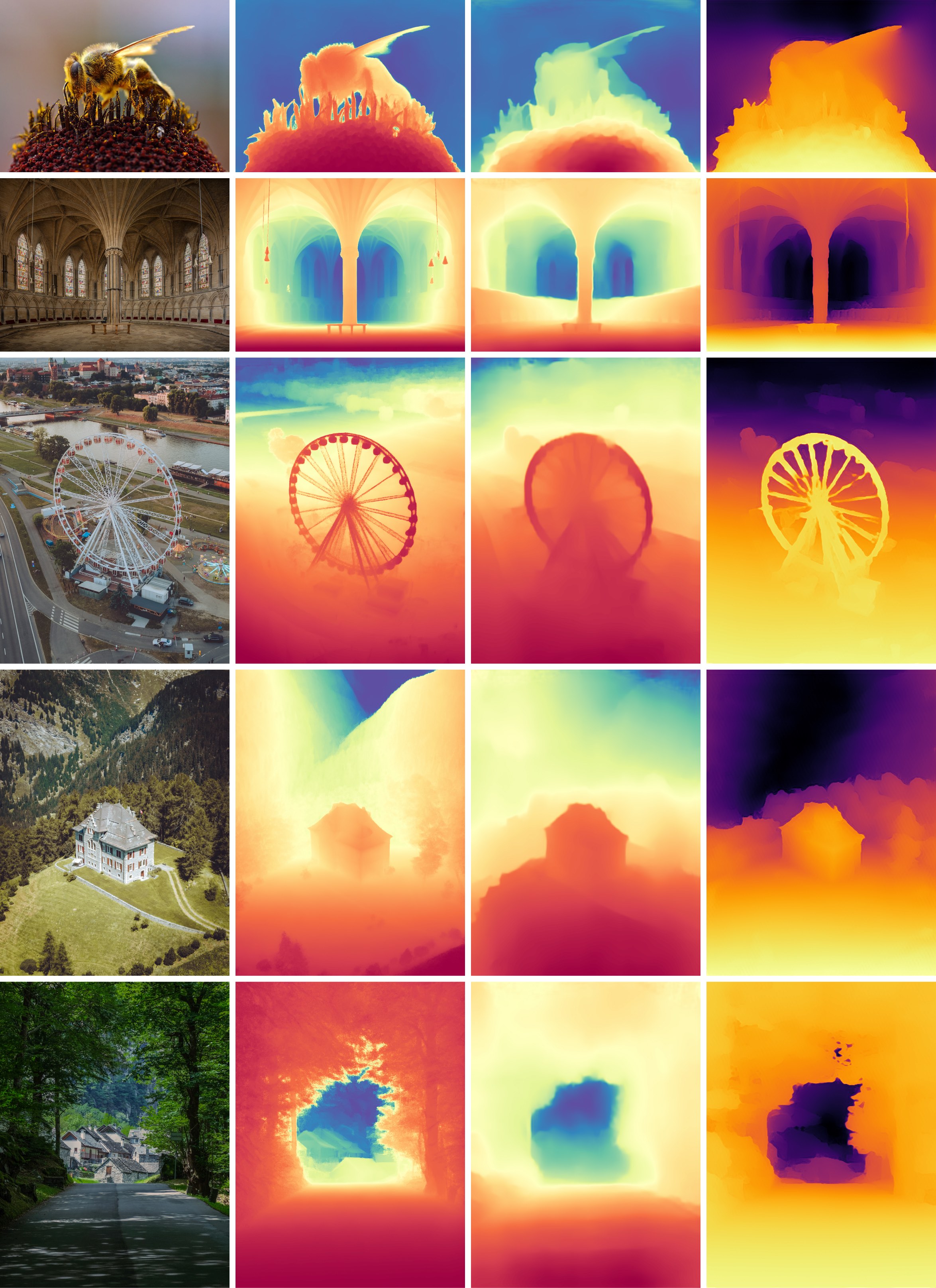}}\\
    \end{tabular}
%}

\clearpage

%\resizebox{\textwidth}{!}{
    \begin{tabular}[p]{ p{\itwviscolumn} p{\itwviscolumn} p{\itwviscolumn} p{\itwviscolumn} }
        \inthewildheader \\
        \multicolumn{4}{c}{\includegraphics[width=0.92\textwidth]{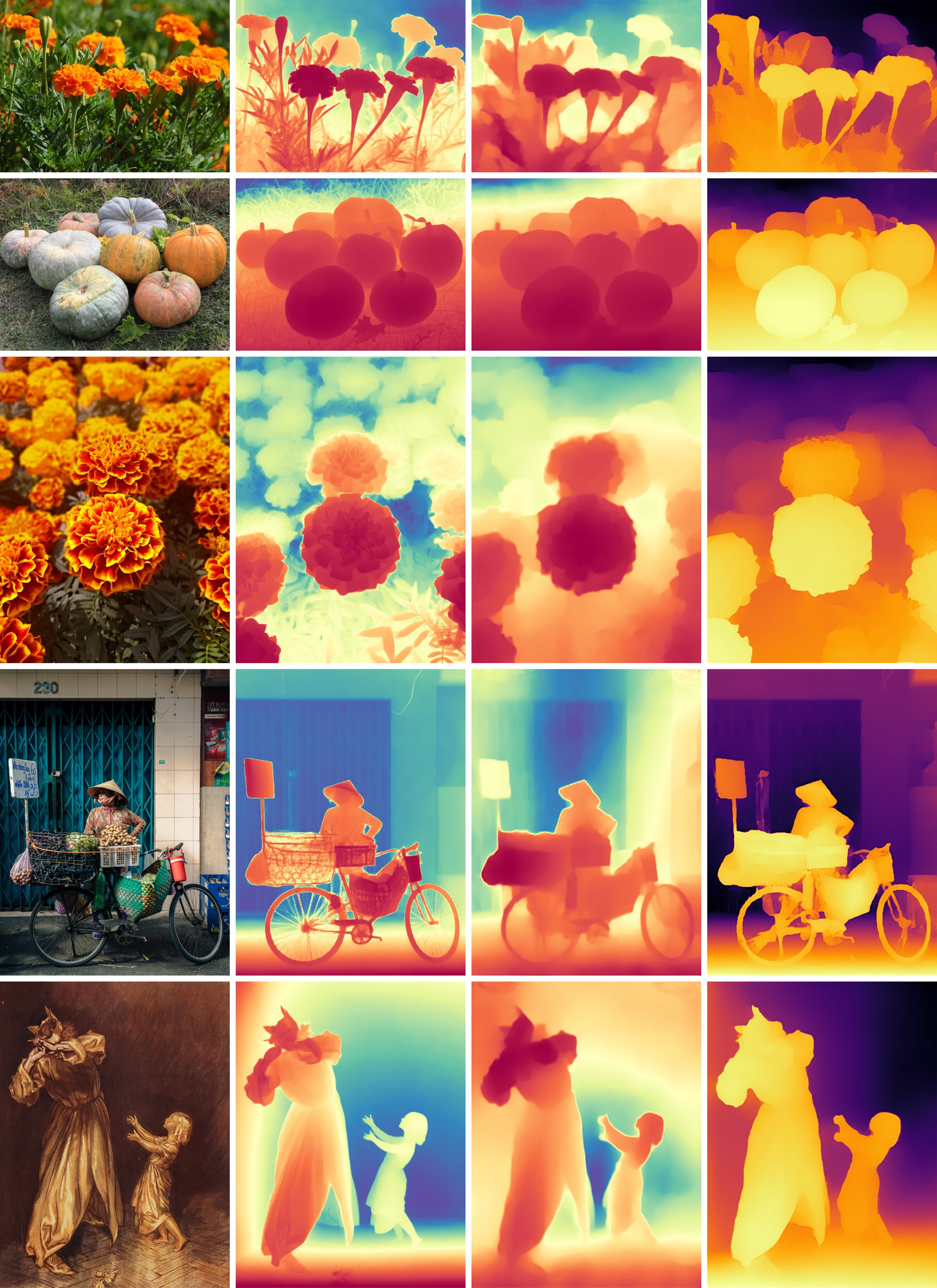}}\\
    \end{tabular}
%}

\clearpage

%\resizebox{\textwidth}{!}{
    \begin{tabular}[p]{ p{\itwviscolumn} p{\itwviscolumn} p{\itwviscolumn} p{\itwviscolumn} }
        \inthewildheader \\
        \multicolumn{4}{c}{\includegraphics[width=0.92\textwidth]{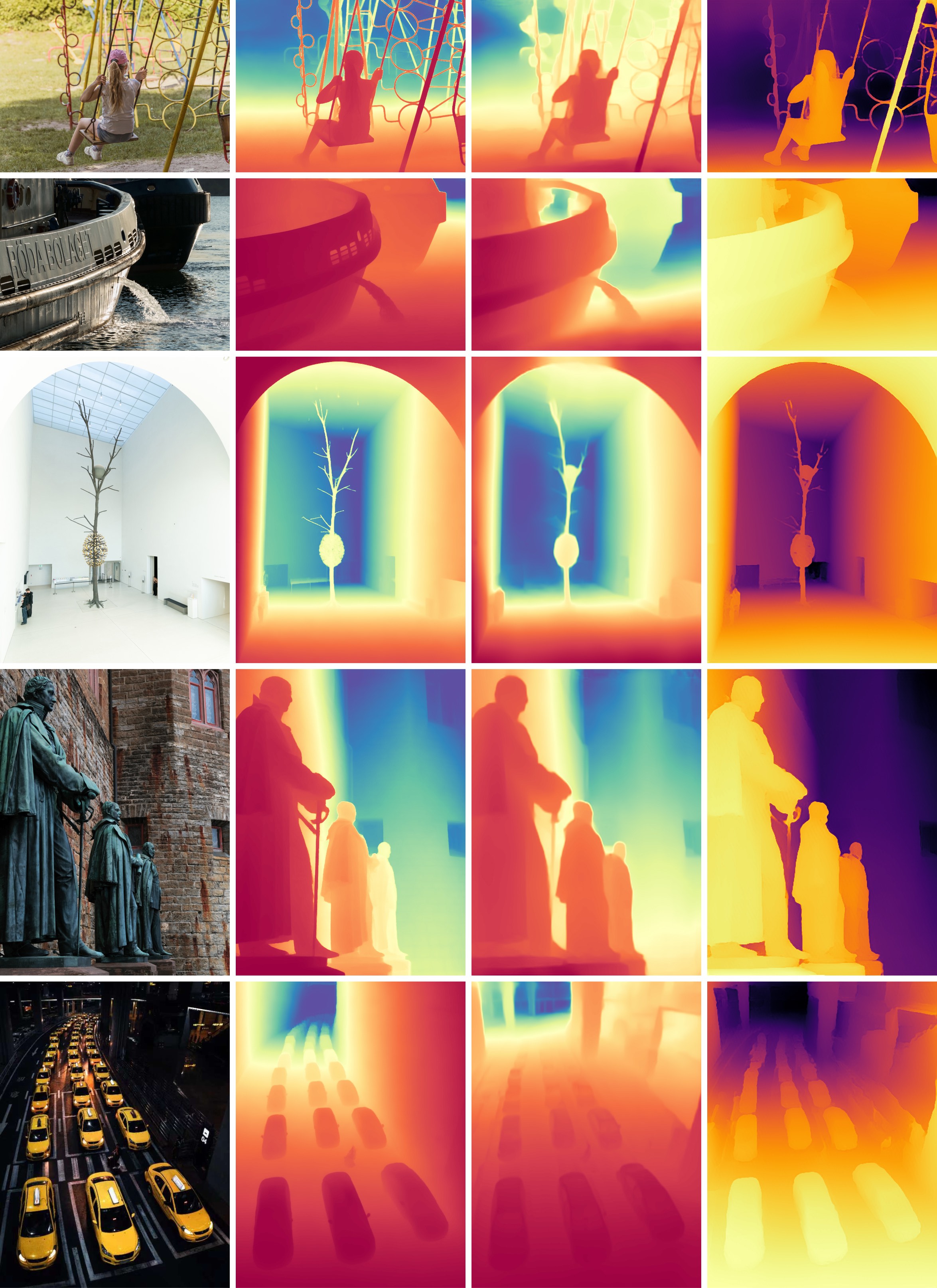}}\\
    \end{tabular}
% }

\clearpage

%\resizebox{\textwidth}{!}{
    \begin{tabular}[p]{ p{\itwviscolumn} p{\itwviscolumn} p{\itwviscolumn} p{\itwviscolumn} }
        \inthewildheader \\
        \multicolumn{4}{c}{\includegraphics[width=0.92\textwidth]{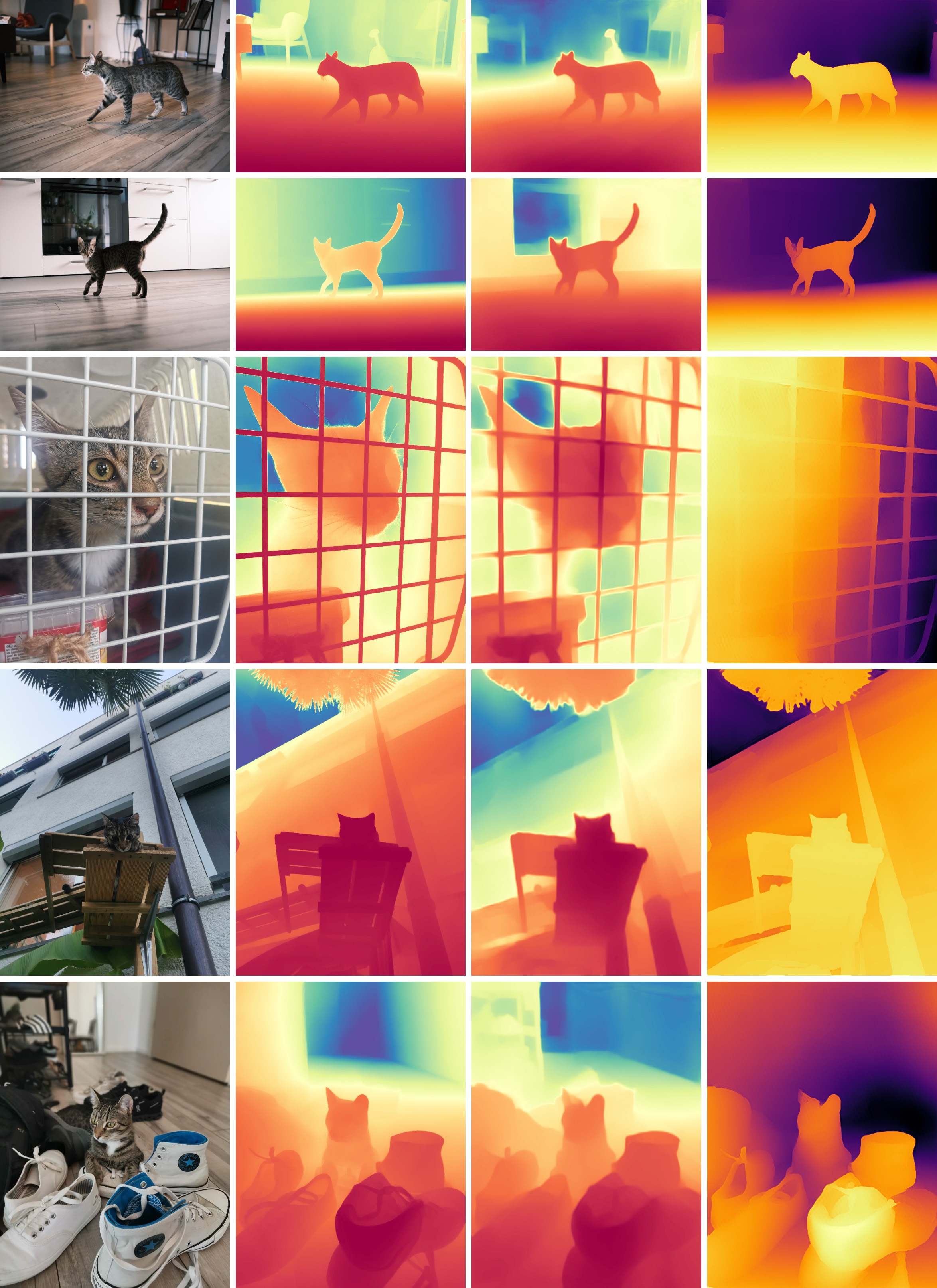}}\\
    \end{tabular}
% }

\clearpage

%\resizebox{\textwidth}{!}{
    \begin{tabular}[p]{ p{\itwviscolumn} p{\itwviscolumn} p{\itwviscolumn} p{\itwviscolumn} }
        \inthewildheader \\
        \multicolumn{4}{c}{\includegraphics[width=0.92\textwidth]{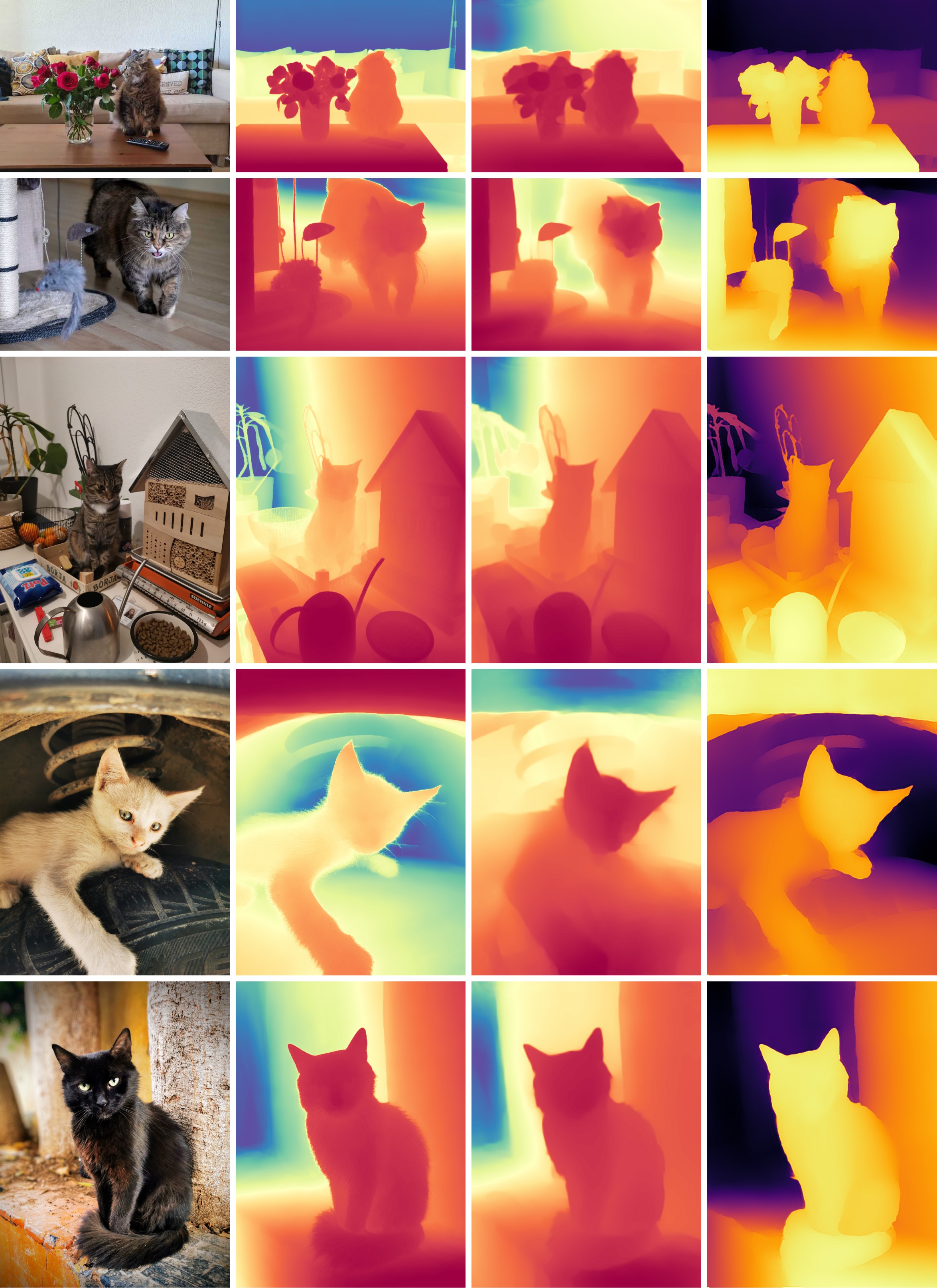}}\\
    \end{tabular}
% }

\clearpage

%\resizebox{\textwidth}{!}{
    \begin{tabular}[p]{ p{\itwviscolumn} p{\itwviscolumn} p{\itwviscolumn} p{\itwviscolumn} }
        \inthewildheader \\
        \multicolumn{4}{c}{\includegraphics[width=0.92\textwidth]{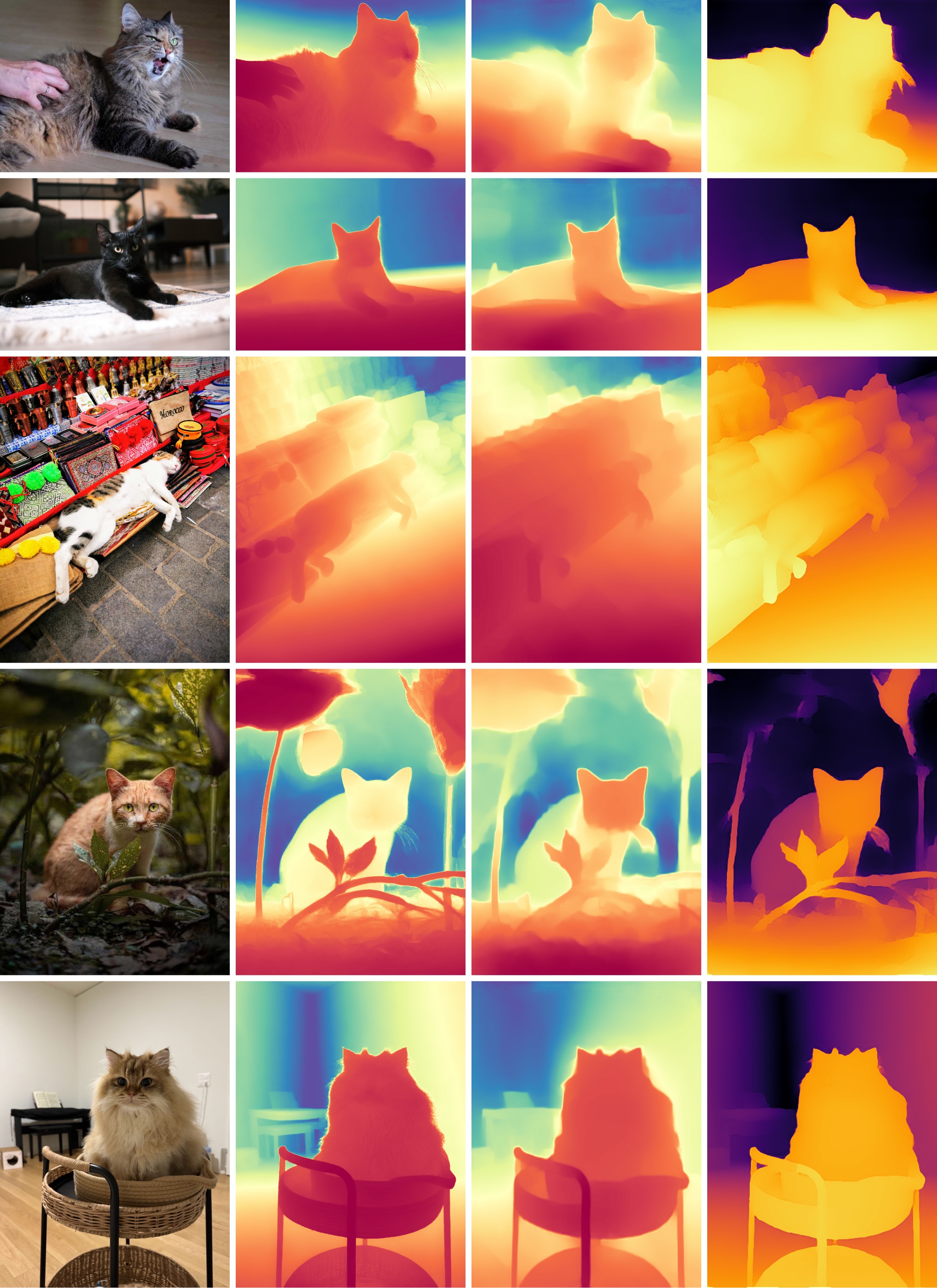}}\\
    \end{tabular}
% }

\clearpage

%\resizebox{\textwidth}{!}{
    \begin{tabular}[p]{ p{\itwviscolumn} p{\itwviscolumn} p{\itwviscolumn} p{\itwviscolumn} }
        \inthewildheader \\
        \multicolumn{4}{c}{\includegraphics[width=0.92\textwidth]{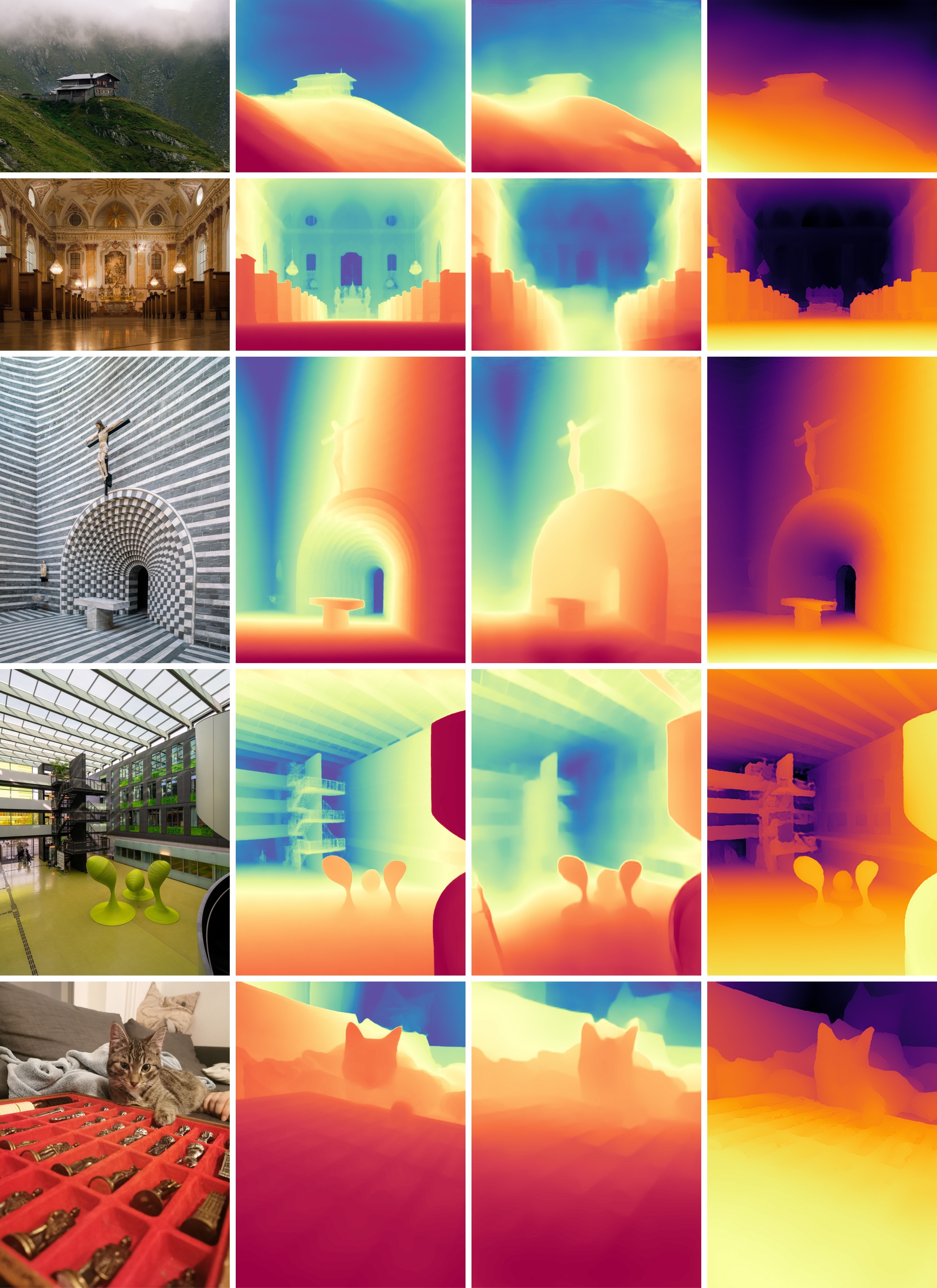}}\\
    \end{tabular}
% }

\clearpage

%\resizebox{\textwidth}{!}{
    \begin{tabular}[p]{ p{\itwviscolumn} p{\itwviscolumn} p{\itwviscolumn} p{\itwviscolumn} }
        \inthewildheader \\
        \multicolumn{4}{c}{\includegraphics[width=0.92\textwidth]{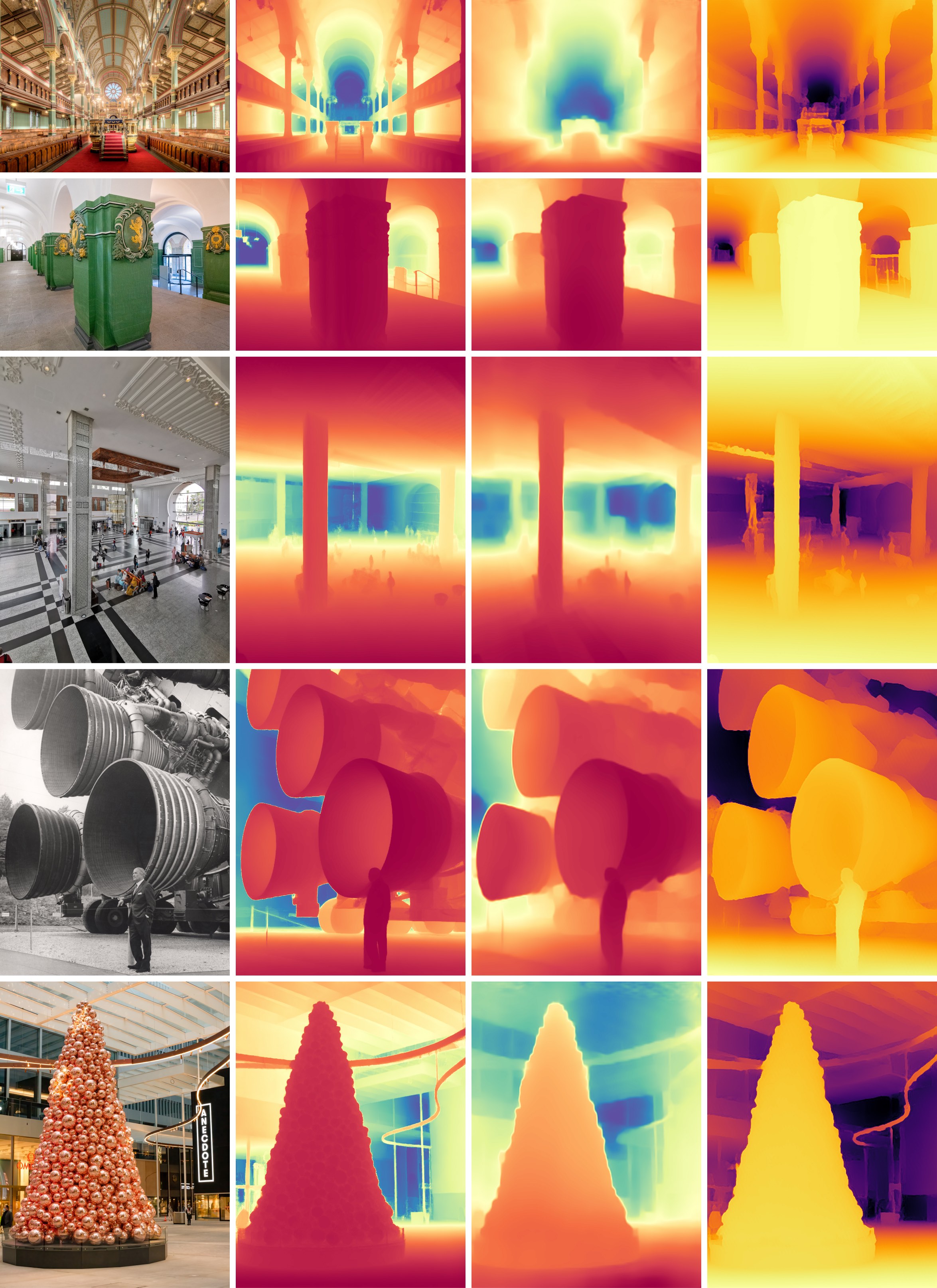}}\\
    \end{tabular}
% }

\clearpage

%\resizebox{\textwidth}{!}{
    \begin{tabular}[p]{ p{\itwviscolumn} p{\itwviscolumn} p{\itwviscolumn} p{\itwviscolumn} }
        \inthewildheader \\
        \multicolumn{4}{c}{\includegraphics[width=0.92\textwidth]{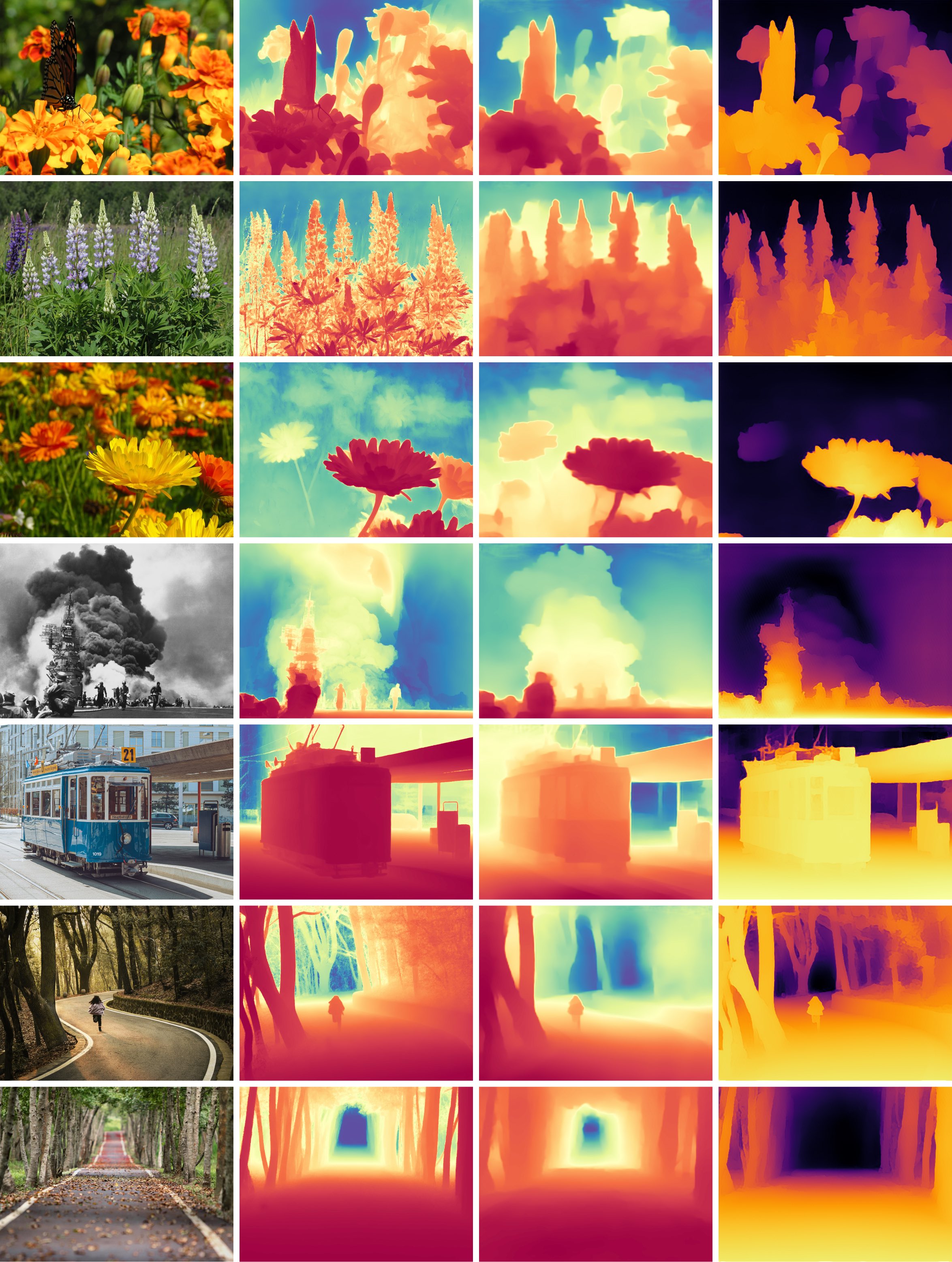}}\\
    \end{tabular}
% }

\clearpage

\begin{figure*}[p]
    \centering
    
% \resizebox{\textwidth}{!}{
    \begin{tabular}[p]{ p{\itwviscolumn} p{\itwviscolumn} p{\itwviscolumn} p{\itwviscolumn} }
        \inthewildheader \\
        \multicolumn{4}{c}{\includegraphics[width=0.92\textwidth]{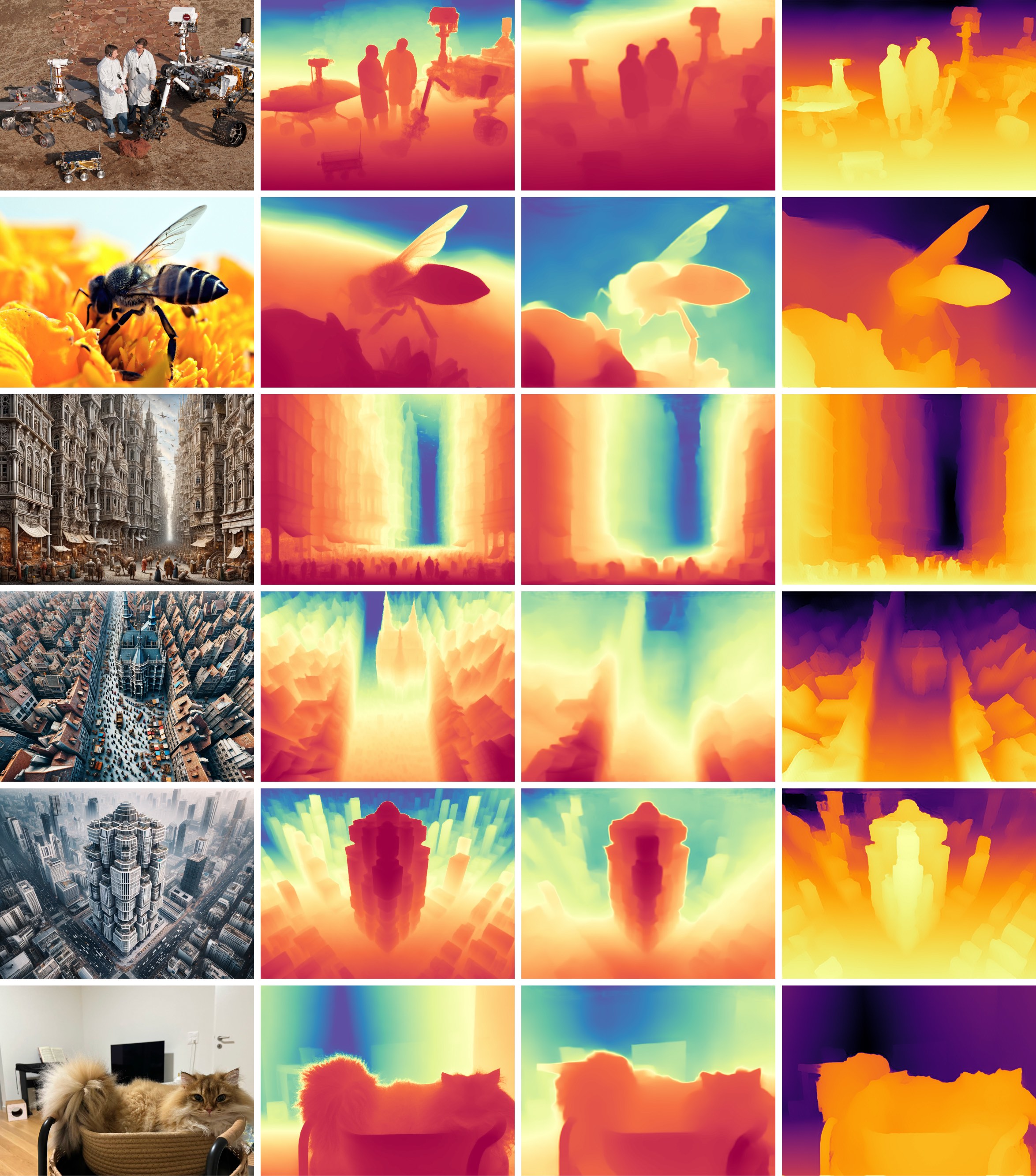}}\\
    \end{tabular}
% }
    \caption{
        \textbf{Qualitative comparison on in-the-wild scenes.} \method{} and LeReS predict depth (with red indicating closer and blue indicating farther distances), while MiDaS predicts disparity (with yellow signifying closer and purple signifying farther distances).
    }
    \label{fig:in-the-wild}
\end{figure*}

% ----------------- depth -----------------
% Define a new command for dataset labels if there is only one sample
% \newcommand{\sampleNameSingle}[1]{
%     \multirow{1}{*}[1.3cm]{\rotatebox{90}{#1}}
% }
\begin{figure*}[p]
    \centering
    \resizebox{\textwidth}{!}{
        \begin{tabular}[ht]{c  p{\viscolumn} p{\viscolumn} p{\viscolumn} p{\viscolumn} }
    
            \multirow{55.5}{*}{\rotatebox[origin=c]{90}{\nyu}}
            & \upperhead \\
            & \multicolumn{4}{c}{\includegraphics[width=\textwidth]{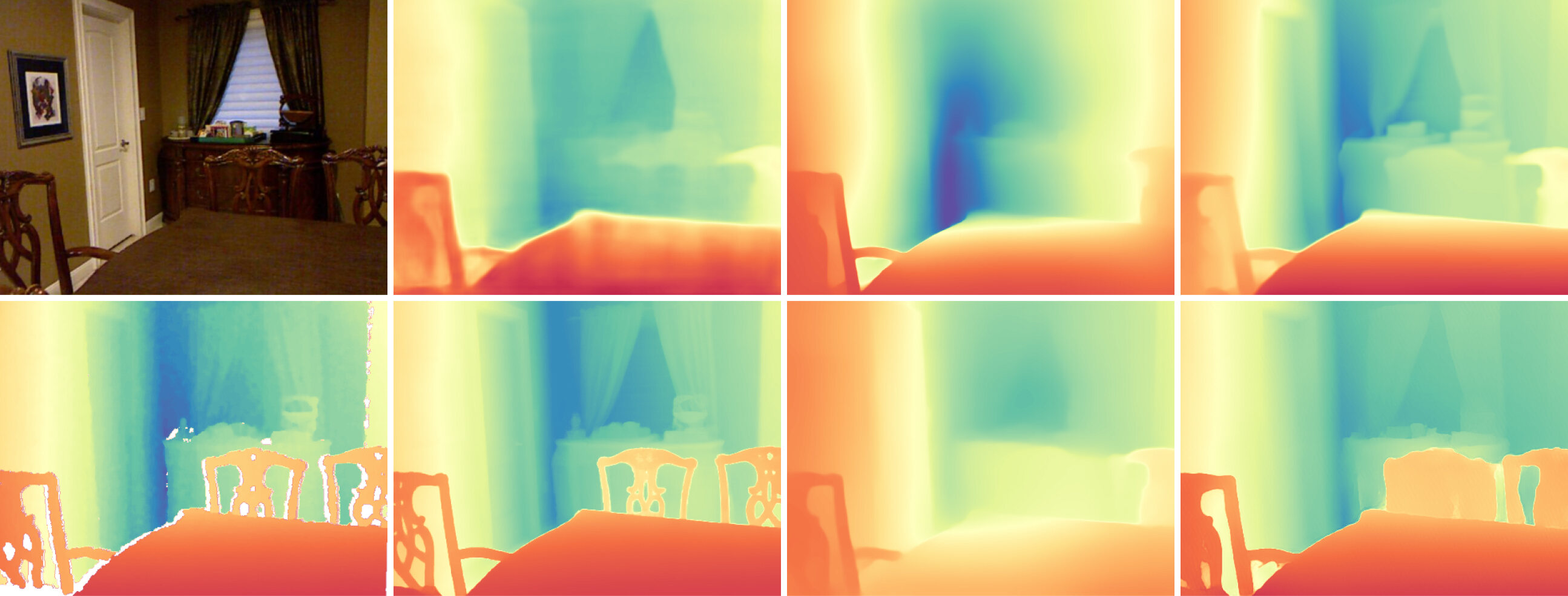}}\\
            & \lowerhead \\
            & { } \\
            & \upperhead \\
            & \multicolumn{4}{c}{\includegraphics[width=\textwidth]{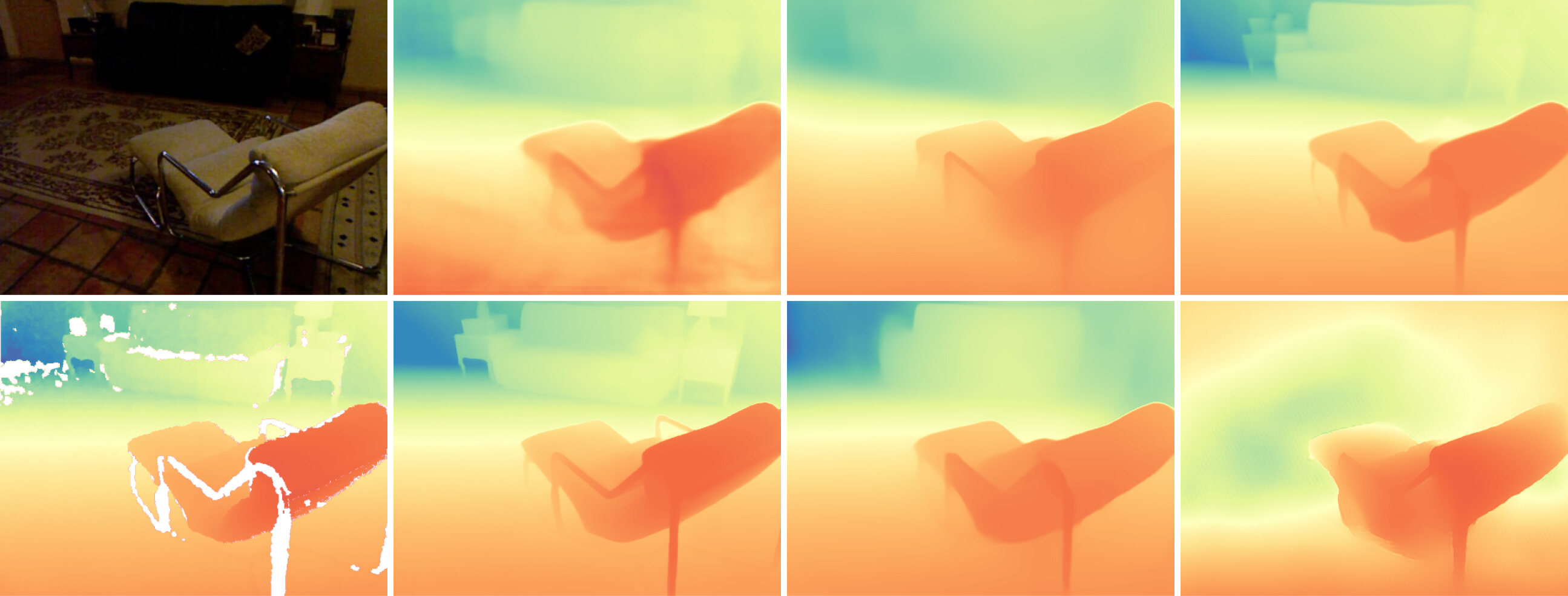}}\\
            & \lowerhead \\
            & { } \\
            & \upperhead \\
            & \multicolumn{4}{c}{\includegraphics[width=\textwidth]{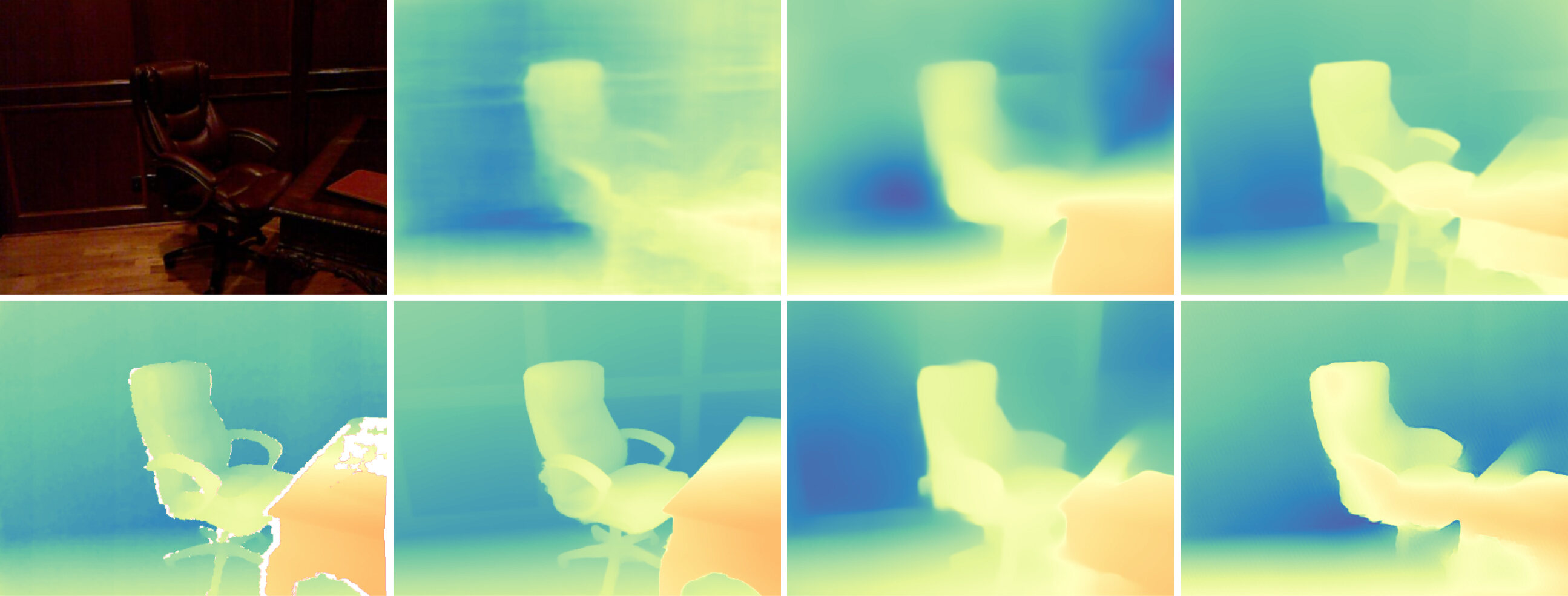}}\\
            & \lowerhead \\
        \end{tabular}
    }

\end{figure*}

\clearpage

\begin{figure*}[p]
    \centering
    \resizebox{\textwidth}{!}{
        \begin{tabular}[ht]{c  p{\viscolumn} p{\viscolumn} p{\viscolumn} p{\viscolumn} }
    
            \multirow{36.8}{*}{\rotatebox[origin=c]{90}{\nyu}}
            & \upperhead \\
            & \multicolumn{4}{c}{\includegraphics[width=\textwidth]{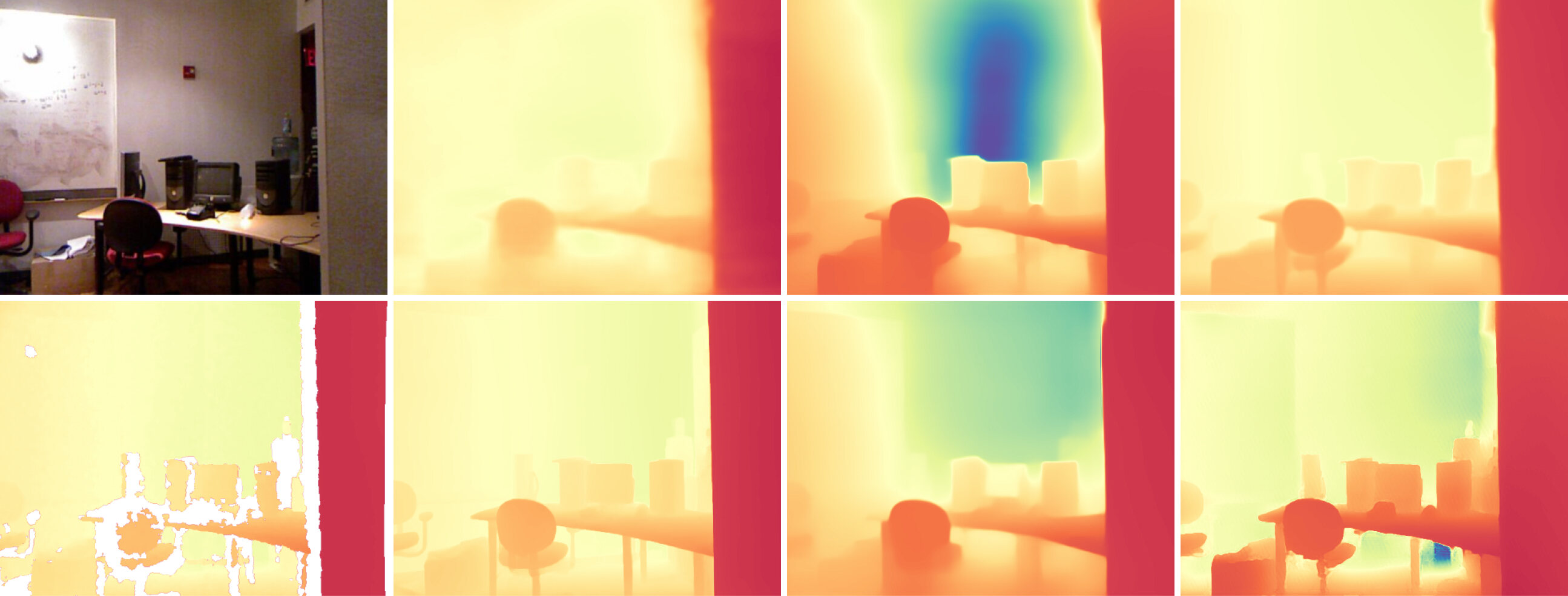}}\\
            & \lowerhead \\
            & { } \\
            & \upperhead \\
            & \multicolumn{4}{c}{\includegraphics[width=\textwidth]{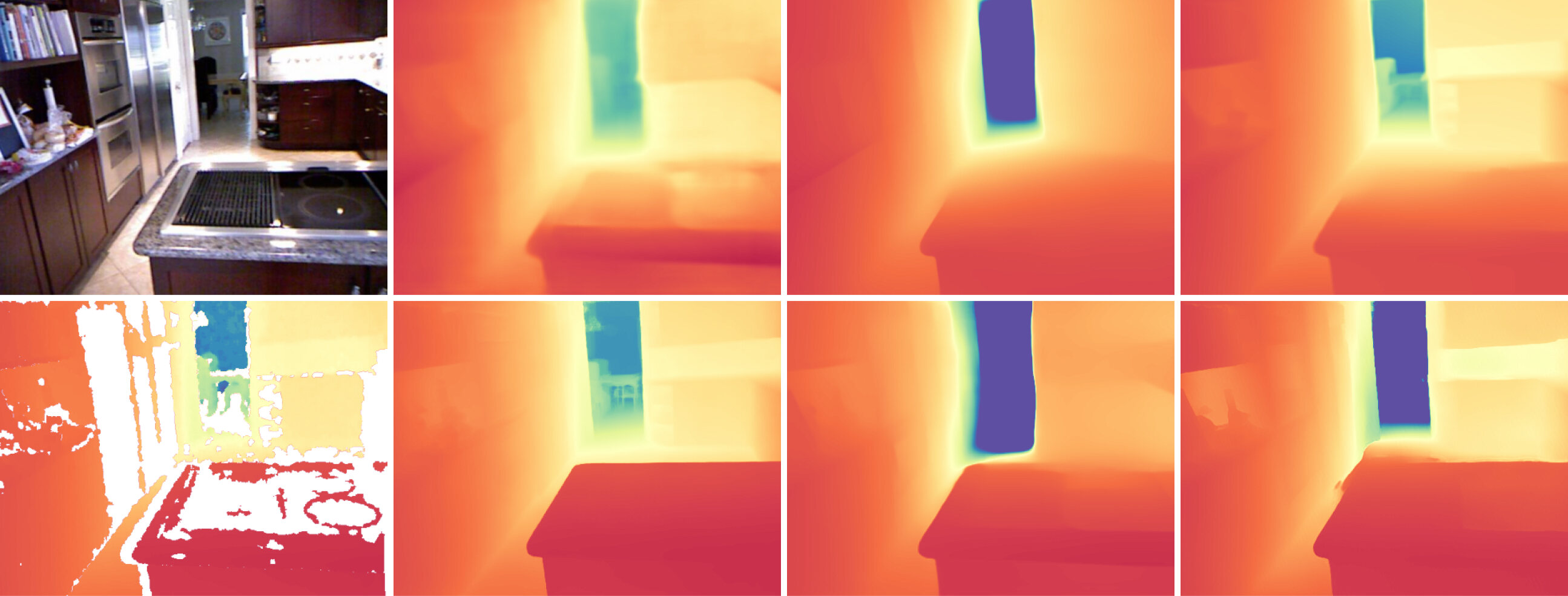}}\\
            & \lowerhead \\
            & { } \\
            \hline
            & { } \\
            \multirow{16.6}{*}{\rotatebox[origin=c]{90}{\kitti}}
            & \upperhead \\
            & \multicolumn{4}{c}{\includegraphics[width=\textwidth]{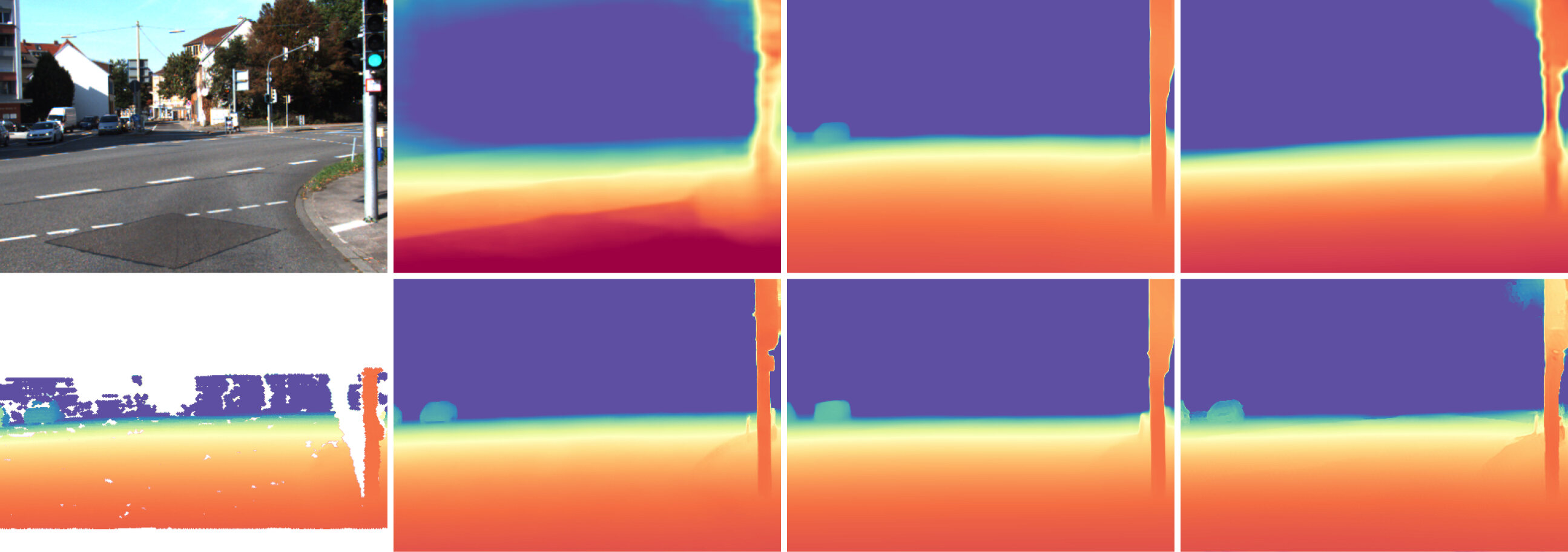}}\\
            & \lowerhead \\
        \end{tabular}
    }
\end{figure*}

\clearpage

\begin{figure*}[p]
    \centering
    \resizebox{\textwidth}{!}{
        \begin{tabular}[ht]{c  p{\viscolumn} p{\viscolumn} p{\viscolumn} p{\viscolumn} }
    
            \multirow{52.5}{*}{\rotatebox[origin=c]{90}{\kitti}}
            & \upperhead \\
            & \multicolumn{4}{c}{\includegraphics[width=\textwidth]{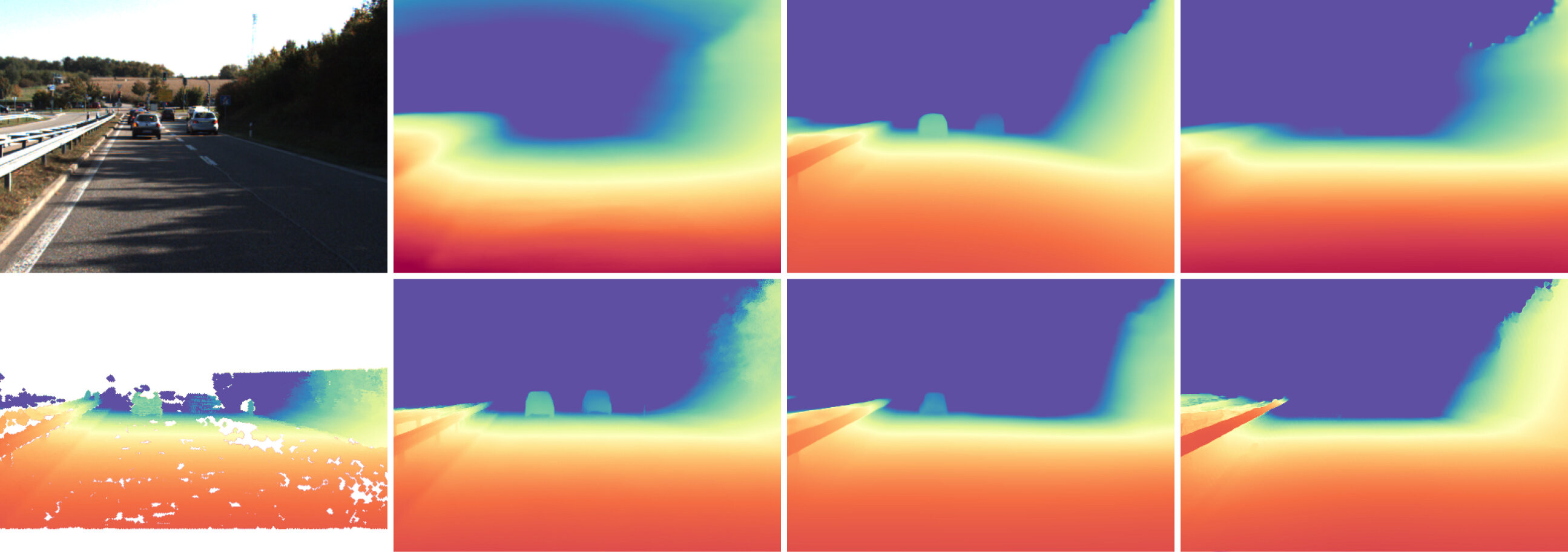}}\\
            & \lowerhead \\
            & { } \\
            & \upperhead \\
            & \multicolumn{4}{c}{\includegraphics[width=\textwidth]{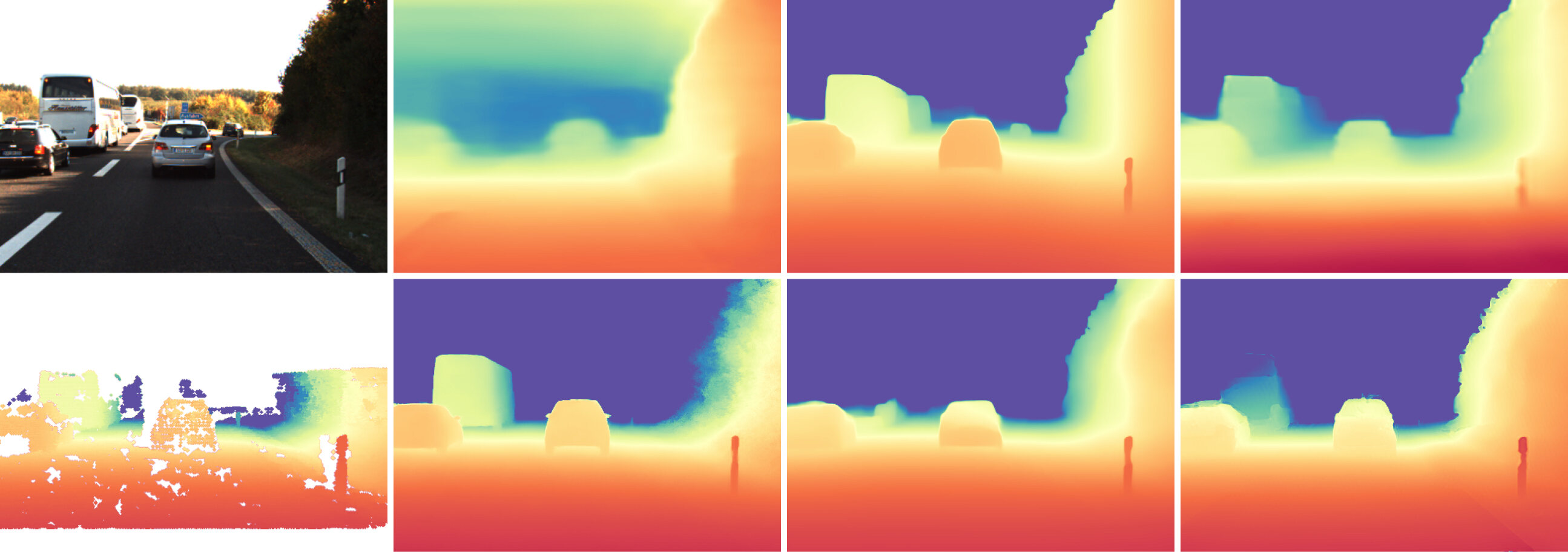}}\\
            & \lowerhead \\
            & { } \\
            & \upperhead \\
            & \multicolumn{4}{c}{\includegraphics[width=\textwidth]{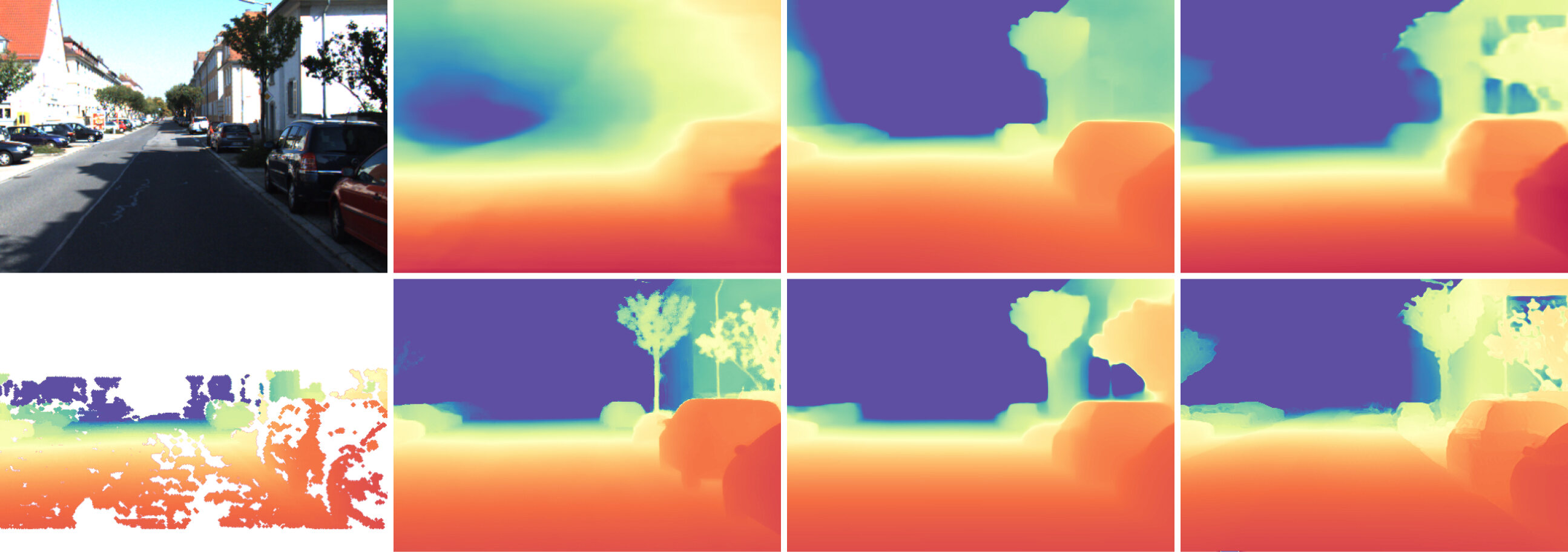}}\\
            & \lowerhead \\
        \end{tabular}
    }
\end{figure*}

\clearpage

\begin{figure*}[p]
    \centering
    \resizebox{\textwidth}{!}{
        \begin{tabular}[ht]{c  p{\viscolumn} p{\viscolumn} p{\viscolumn} p{\viscolumn} }
    
            \multirow{16.6}{*}{\rotatebox[origin=c]{90}{\kitti}}
            & \upperhead \\
            & \multicolumn{4}{c}{\includegraphics[width=\textwidth]{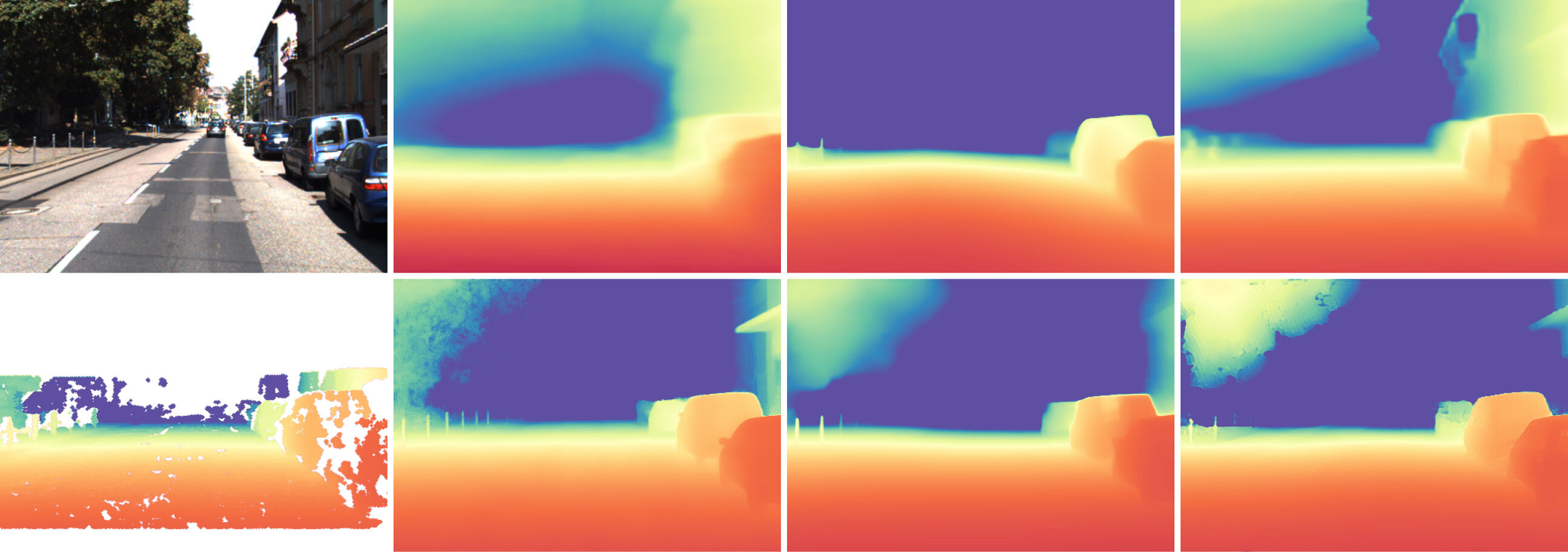}}\\
            & \lowerhead \\
            & { } \\
            \hline
            & { } \\
    
            \multirow{32.8}{*}{\rotatebox[origin=c]{90}{\eththreed}}
            & \upperhead \\
            & \multicolumn{4}{c}{\includegraphics[width=\textwidth]{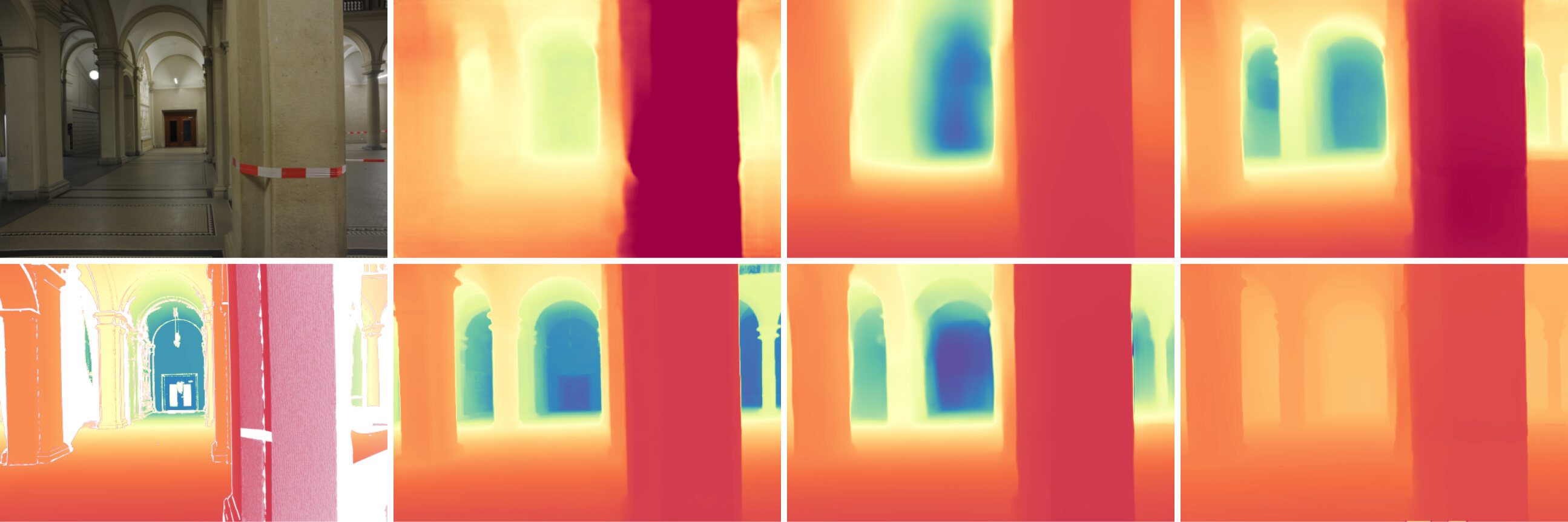}}\\
            & \lowerhead \\
            & { } \\
            & \upperhead \\
            & \multicolumn{4}{c}{\includegraphics[width=\textwidth]{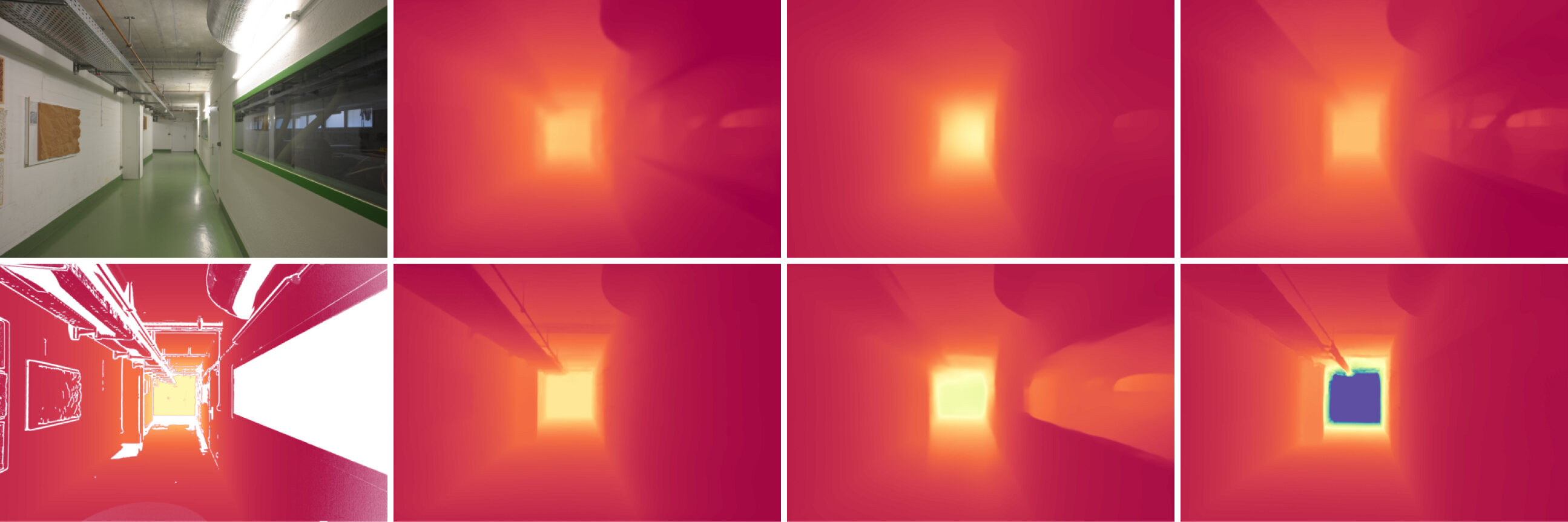}}\\
            & \lowerhead \\
        \end{tabular}
    }
\end{figure*}

\clearpage

\begin{figure*}[p]
    \centering
    \resizebox{\textwidth}{!}{
        \begin{tabular}[ht]{c  p{\viscolumn} p{\viscolumn} p{\viscolumn} p{\viscolumn} }
    
            \multirow{50}{*}{\rotatebox[origin=c]{90}{\eththreed}}
            & \upperhead \\
            & \multicolumn{4}{c}{\includegraphics[width=\textwidth]{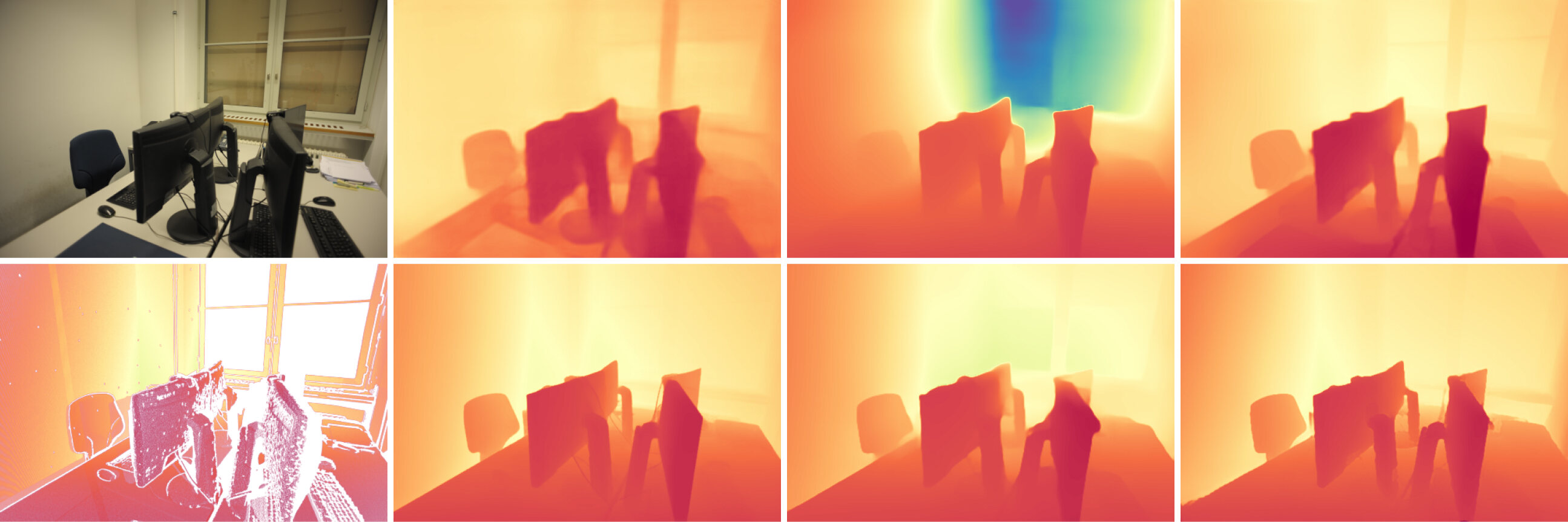}}\\
            & \lowerhead \\
            & { } \\
            & \upperhead \\
            & \multicolumn{4}{c}{\includegraphics[width=\textwidth]{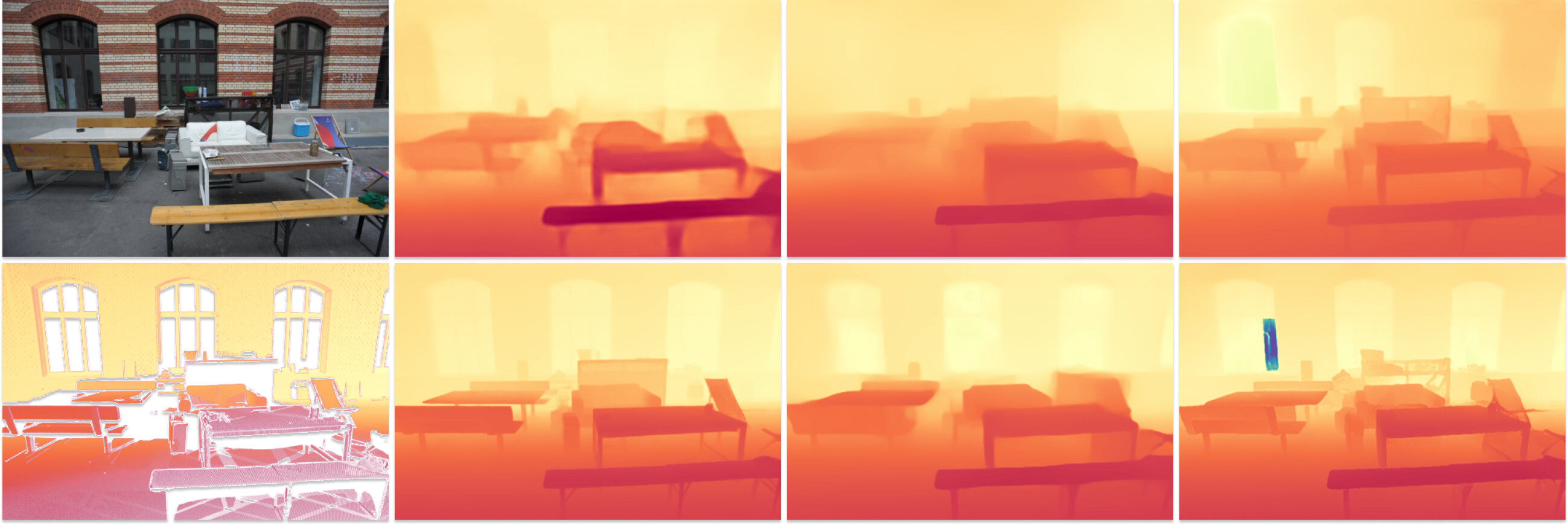}}\\
            & \lowerhead \\
            & { } \\
            & \upperhead \\
            & \multicolumn{4}{c}{\includegraphics[width=\textwidth]{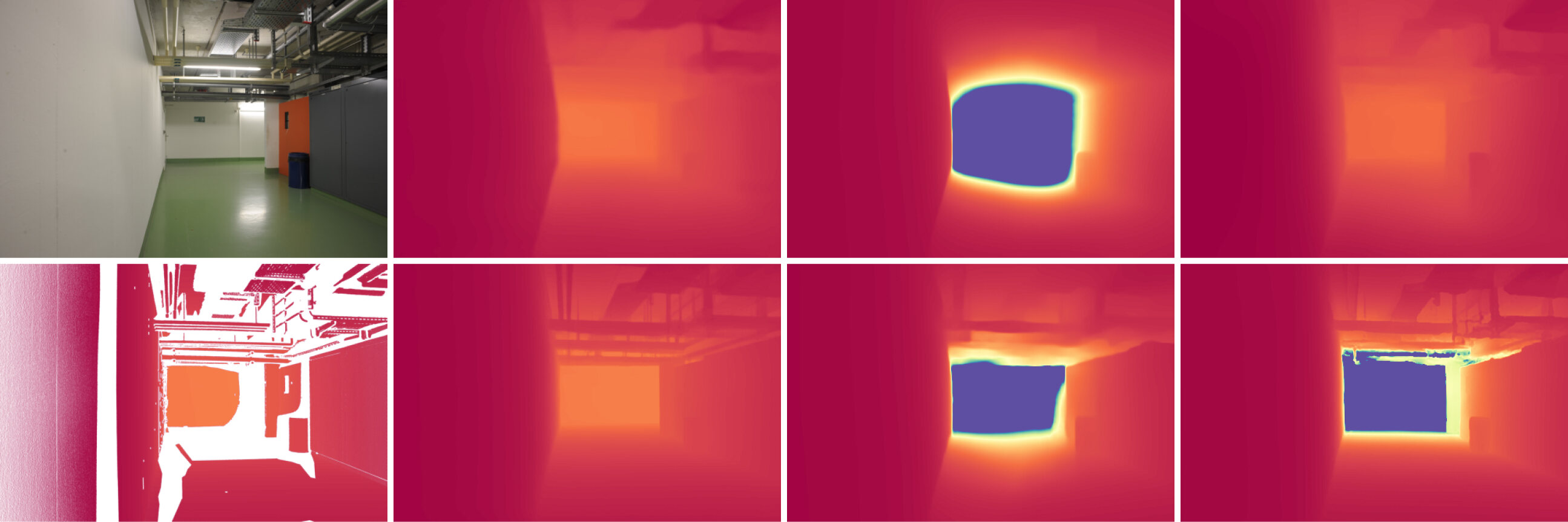}}\\
            & \lowerhead \\
        \end{tabular}
    }
\end{figure*}

\clearpage

\begin{figure*}[p]
    \centering
    \resizebox{\textwidth}{!}{
        \begin{tabular}[ht]{c  p{\viscolumn} p{\viscolumn} p{\viscolumn} p{\viscolumn} }
    
            \multirow{54.5}{*}{\rotatebox[origin=c]{90}{\scannet}}
            & \upperhead \\
            & \multicolumn{4}{c}{\includegraphics[width=\textwidth]{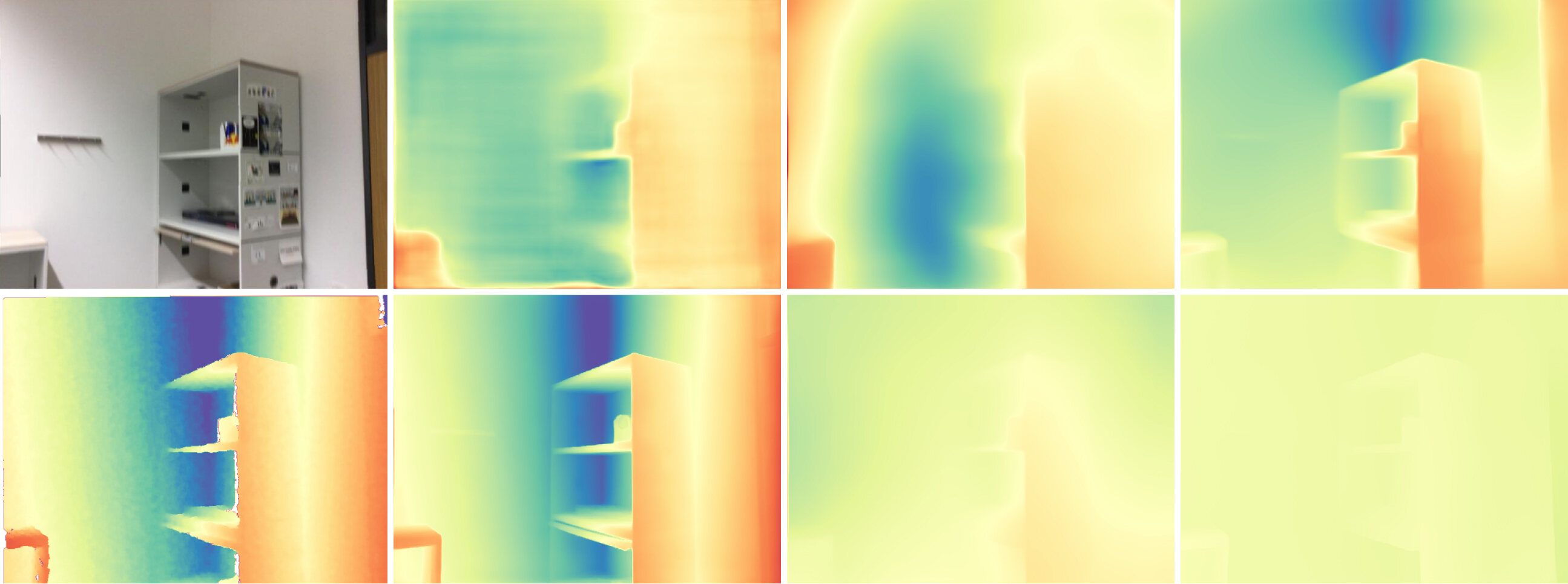}}\\
            & \lowerhead \\
            & { } \\
            & \upperhead \\
            & \multicolumn{4}{c}{\includegraphics[width=\textwidth]{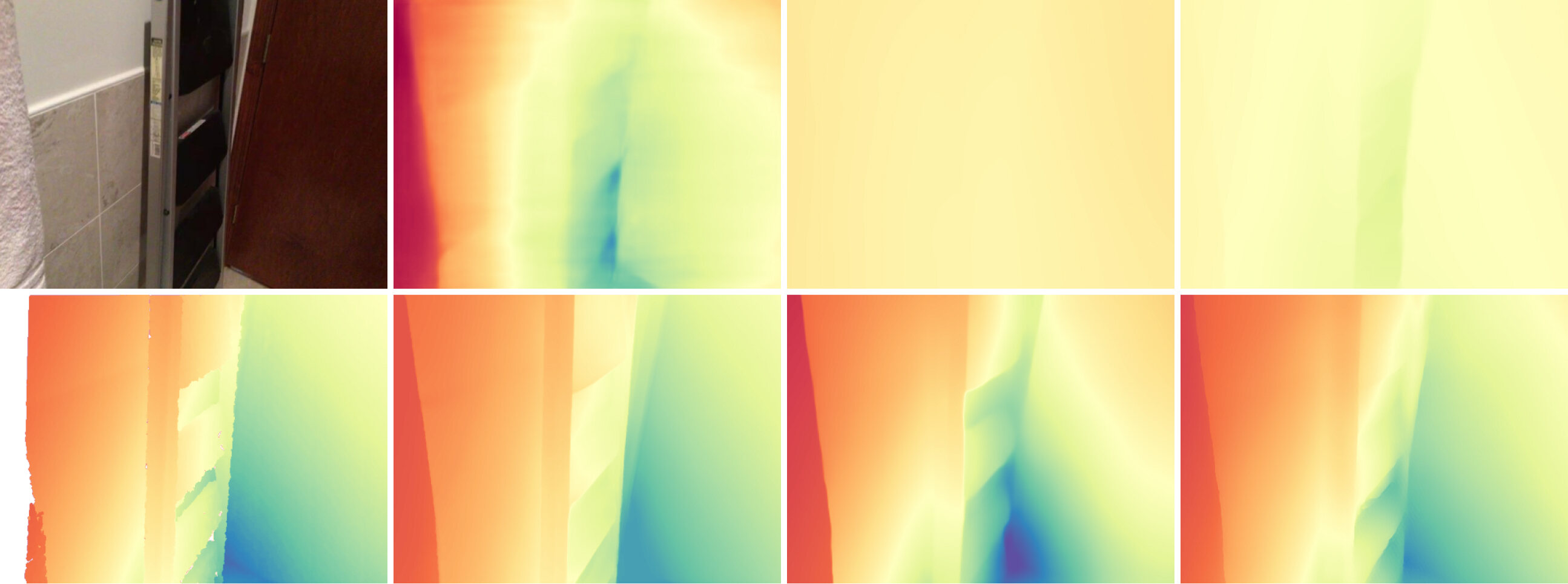}}\\
            & \lowerhead \\
            & { } \\
            & \upperhead \\
            & \multicolumn{4}{c}{\includegraphics[width=\textwidth]{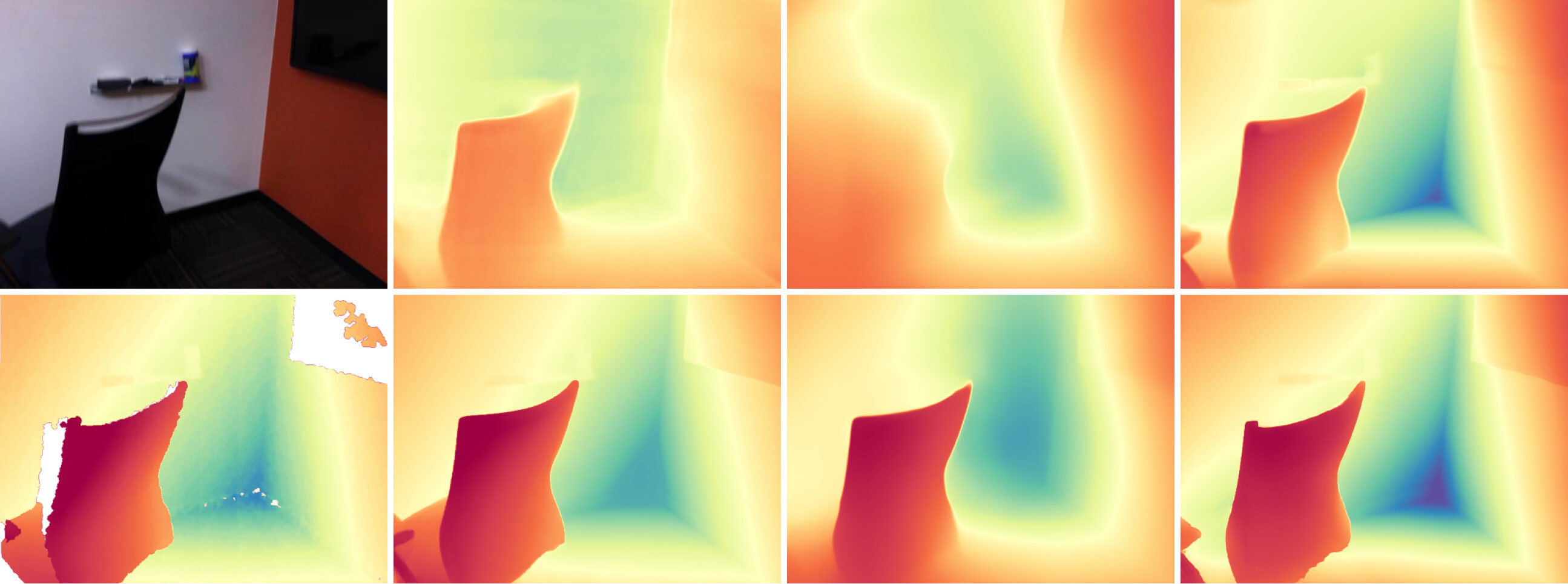}}\\
            & \lowerhead \\
        \end{tabular}
    }
\end{figure*}

\clearpage

\begin{figure*}[p]
    \centering
    \resizebox{\textwidth}{!}{
        \begin{tabular}[ht]{c  p{\viscolumn} p{\viscolumn} p{\viscolumn} p{\viscolumn} }
    
            \multirow{36.4}{*}{\rotatebox[origin=c]{90}{\scannet}}
            & \upperhead \\
            & \multicolumn{4}{c}{\includegraphics[width=\textwidth]{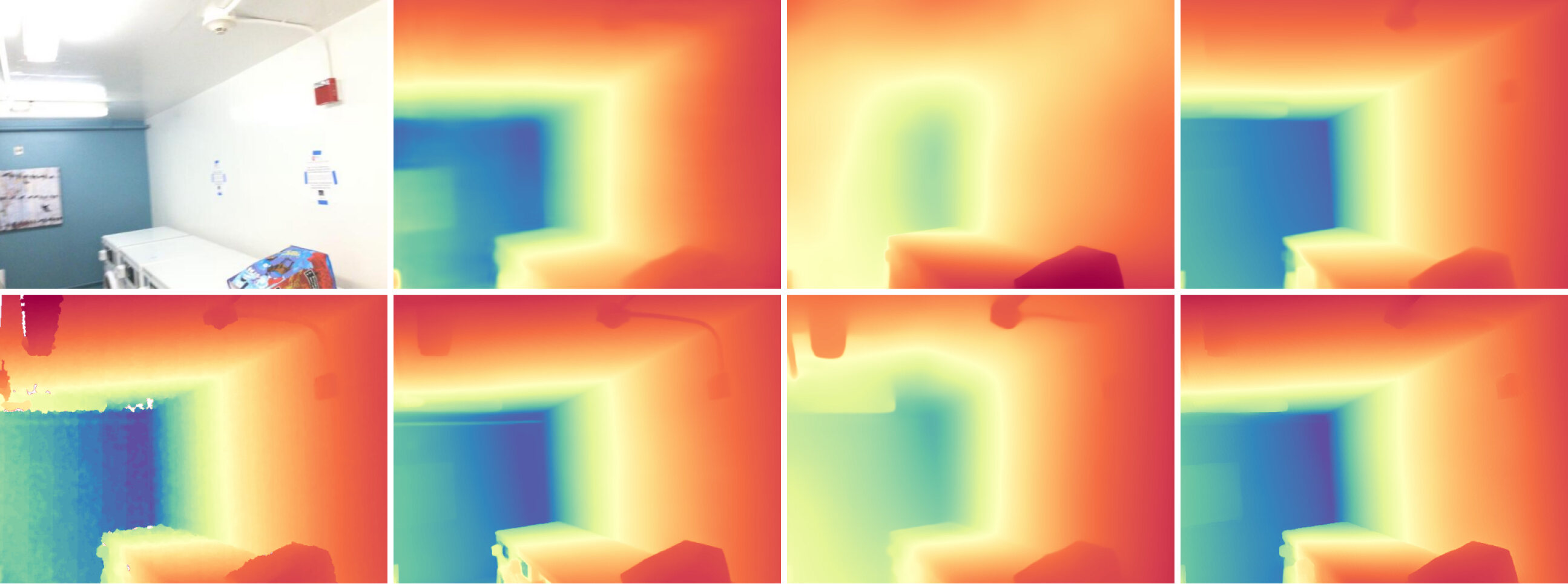}}\\
            & \lowerhead \\
            & { } \\
            & \upperhead \\
            & \multicolumn{4}{c}{\includegraphics[width=\textwidth]{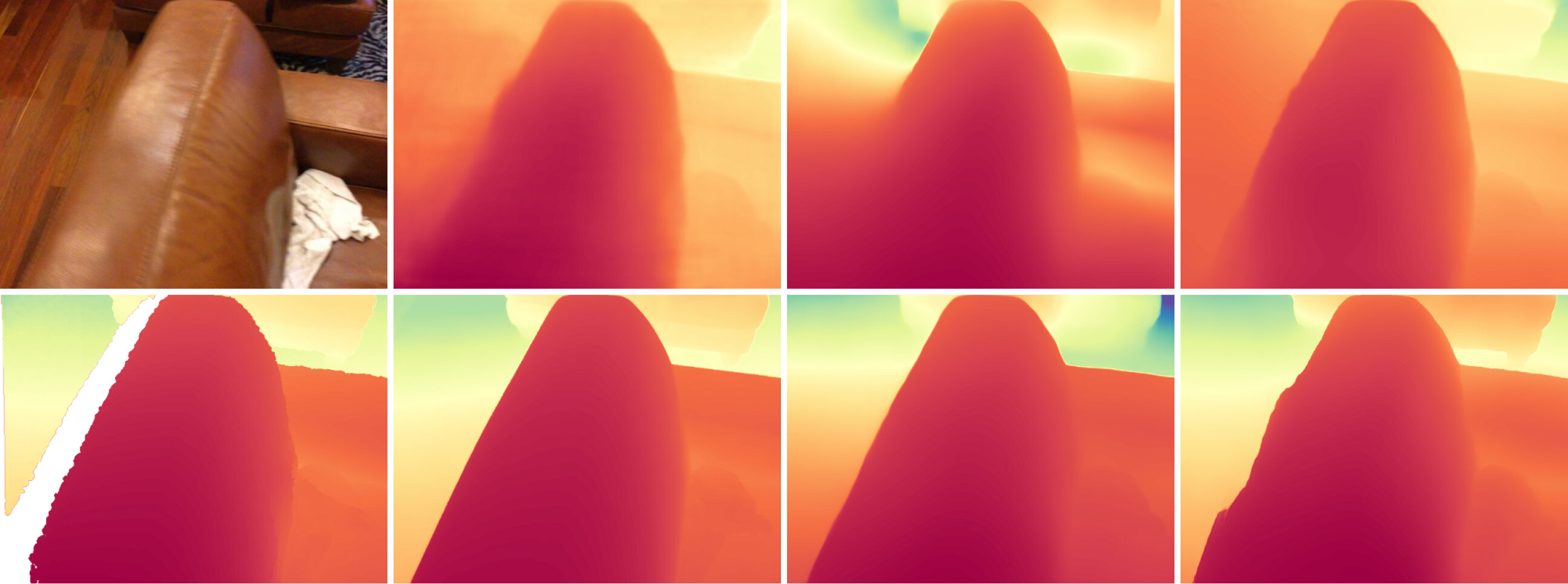}}\\
            & \lowerhead \\
            & { } \\
            \hline
            & { } \\
            \multirow{17}{*}{\rotatebox[origin=c]{90}{\diode}}
            & \upperhead \\
            & \multicolumn{4}{c}{\includegraphics[width=\textwidth]{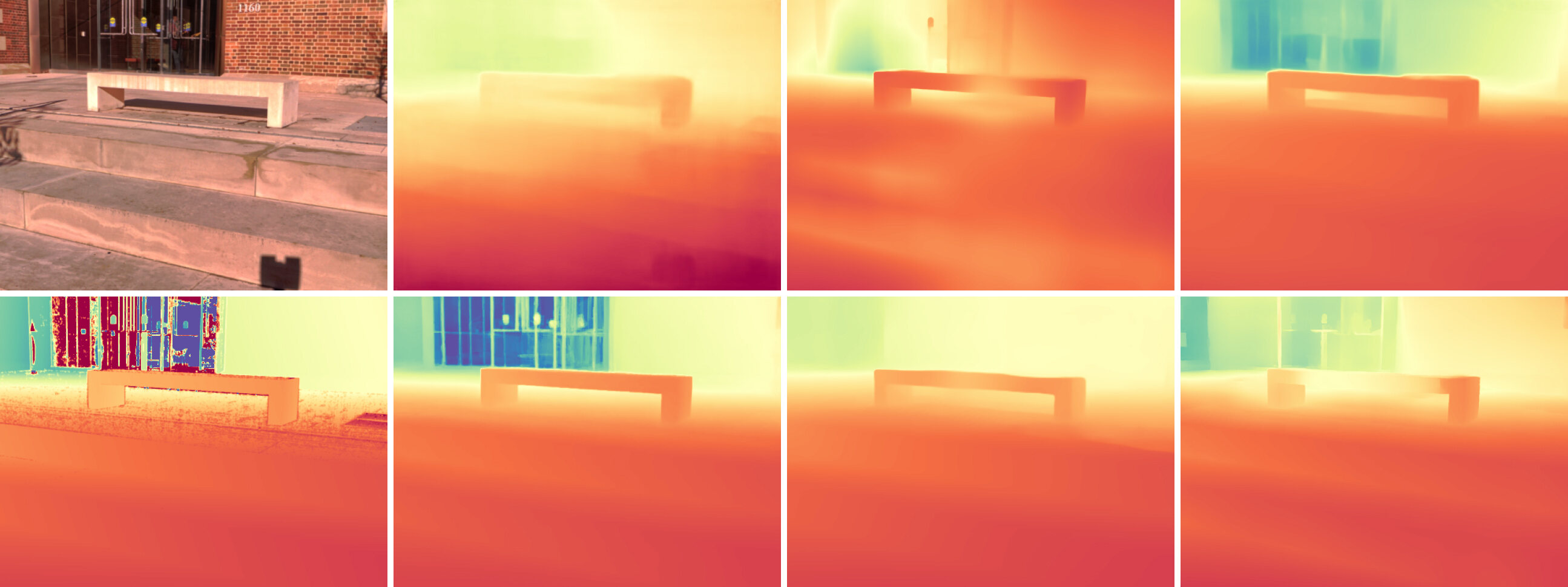}}\\
            & \lowerhead \\
        \end{tabular}
    }
    \end{figure*}

\clearpage

\begin{figure*}[p]
    \centering
        \resizebox{\textwidth}{!}{
        \begin{tabular}[ht]{c  p{\viscolumn} p{\viscolumn} p{\viscolumn} p{\viscolumn} }
    
            \multirow{54.5}{*}{\rotatebox[origin=c]{90}{\diode}}
            & \upperhead \\
            & \multicolumn{4}{c}{\includegraphics[width=\textwidth]{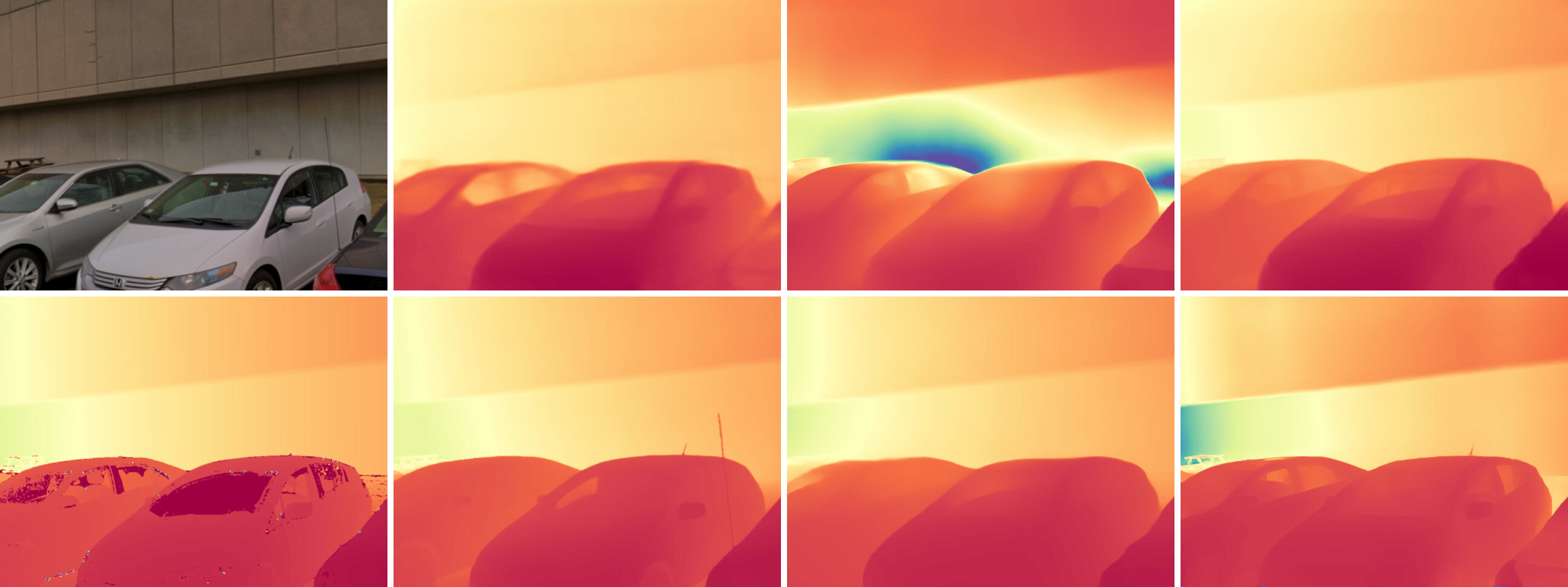}}\\
            & \lowerhead \\
            & { } \\
            & \upperhead \\
            & \multicolumn{4}{c}{\includegraphics[width=\textwidth]{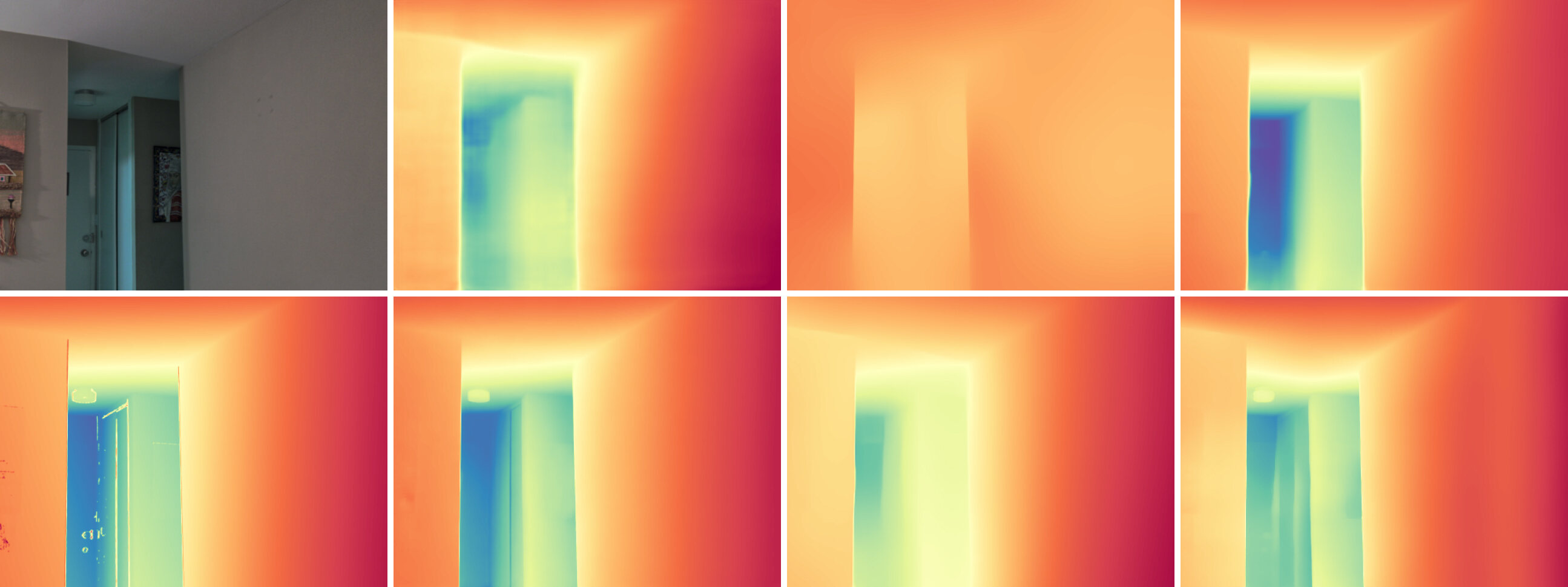}}\\
            & \lowerhead \\
            & { } \\
            & \upperhead \\
            & \multicolumn{4}{c}{\includegraphics[width=\textwidth]{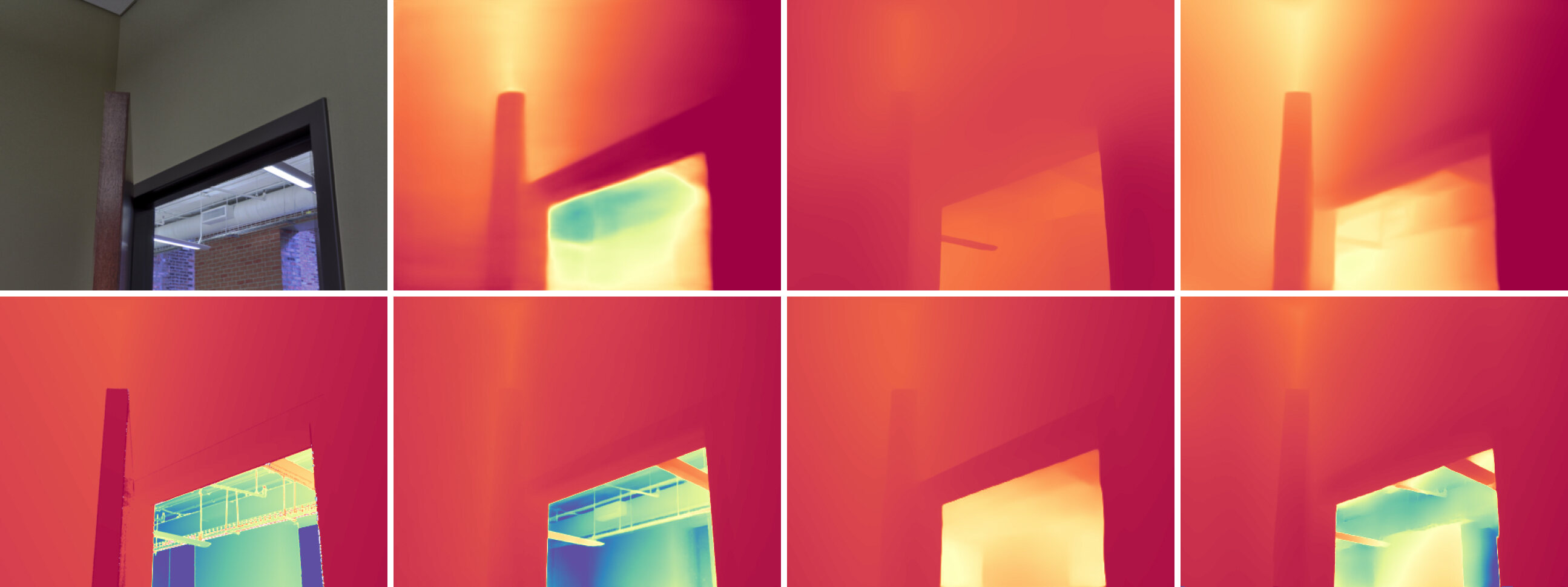}}\\
            & \lowerhead \\
        \end{tabular}
    }
 
    \vspace{-3mm}
    \caption{
        \textbf{Qualitative comparison (depth)} of monocular depth estimation methods across different datasets. Predictions are aligned to ground truth.
        For every sample, the color coding is consistent across all depth maps.
    }
    \label{fig:supp_qualitative_2d}
\end{figure*}

% ----------------- normal -----------------
% Define a new command for dataset labels if there is only one sample
% \newcommand{\sampleNameSingle}[1]{
%     \multirow{1}{*}[1.3cm]{\rotatebox{90}{#1}}
% }

\begin{figure*}[p]
    \centering

    \resizebox{\textwidth}{!}{
        \begin{tabular}[ht]{c  p{\viscolumn} p{\viscolumn} p{\viscolumn} p{\viscolumn} }
    
            \multirow{55.5}{*}{\rotatebox[origin=c]{90}{\nyu}}
            & \upperhead \\
            & \multicolumn{4}{c}{\includegraphics[width=\textwidth]{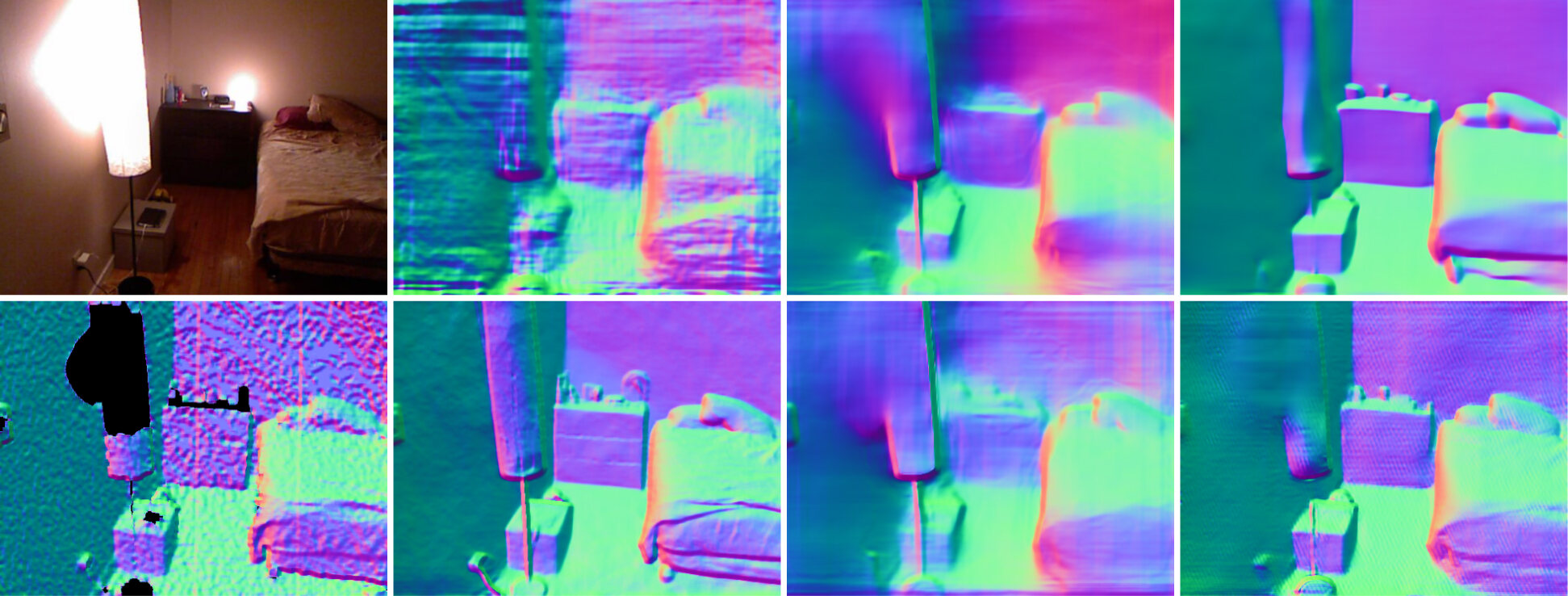}}\\
            & \lowerheadnormal \\
            & { } \\
            & \upperhead \\
            & \multicolumn{4}{c}{\includegraphics[width=\textwidth]{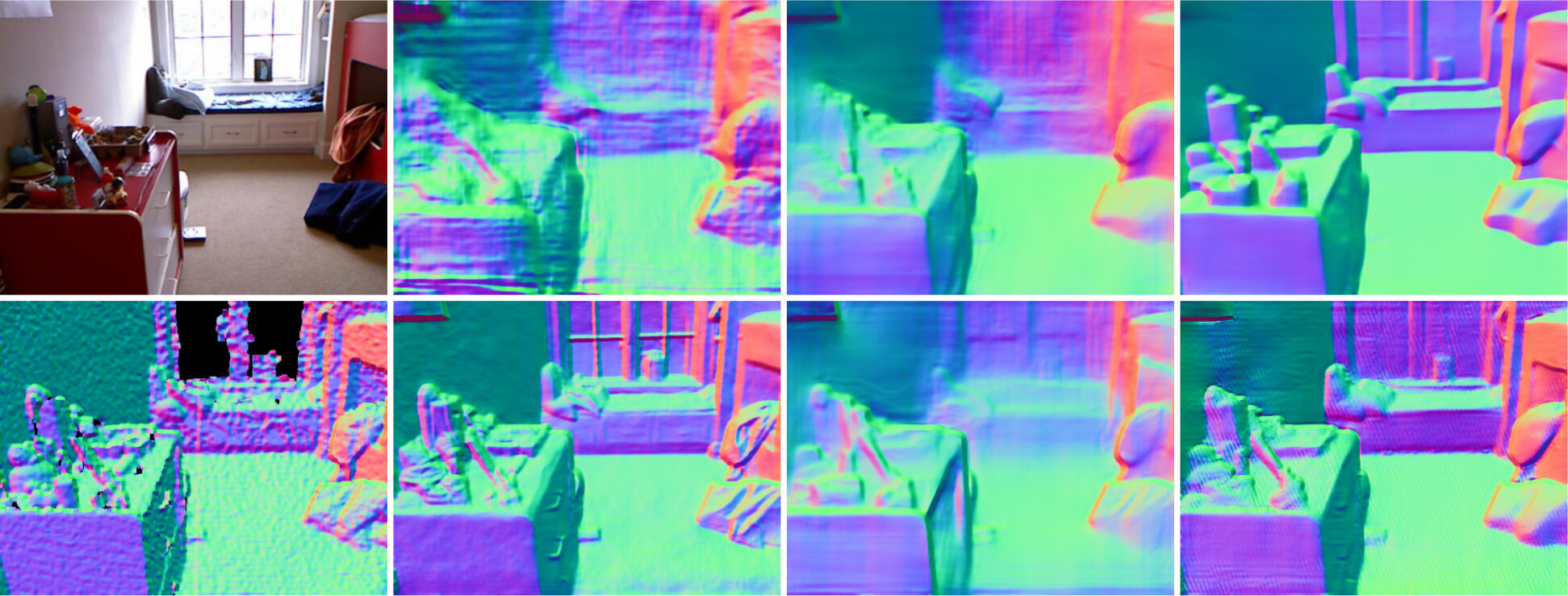}}\\
            & \lowerheadnormal \\
            & { } \\
            & \upperhead \\
            & \multicolumn{4}{c}{\includegraphics[width=\textwidth]{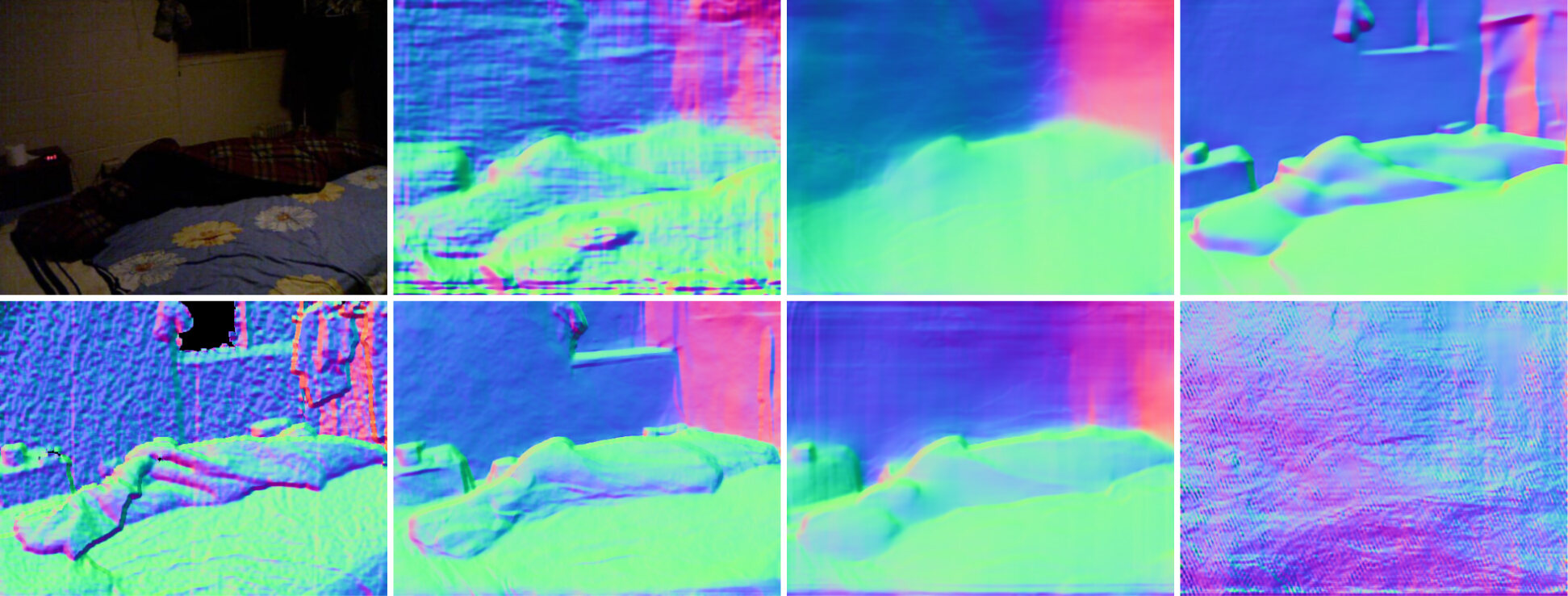}}\\
            & \lowerheadnormal \\
        \end{tabular}
    }
    
\end{figure*}

\clearpage

\begin{figure*}[p]
    \centering
    \resizebox{\textwidth}{!}{
        \begin{tabular}[ht]{c  p{\viscolumn} p{\viscolumn} p{\viscolumn} p{\viscolumn} }
    
            \multirow{50}{*}{\rotatebox[origin=c]{90}{\eththreed}}
            & \upperhead \\
            & \multicolumn{4}{c}{\includegraphics[width=\textwidth]{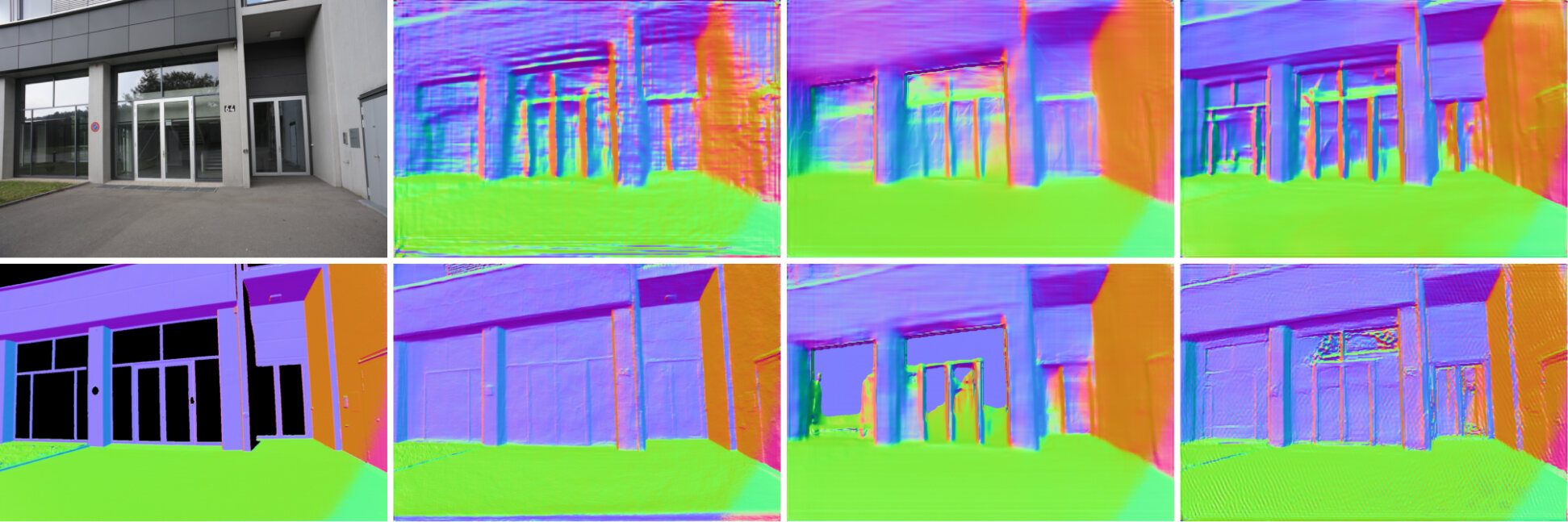}}\\
            & \lowerheadnormal \\
            & { } \\
            & \upperhead \\
            & \multicolumn{4}{c}{\includegraphics[width=\textwidth]{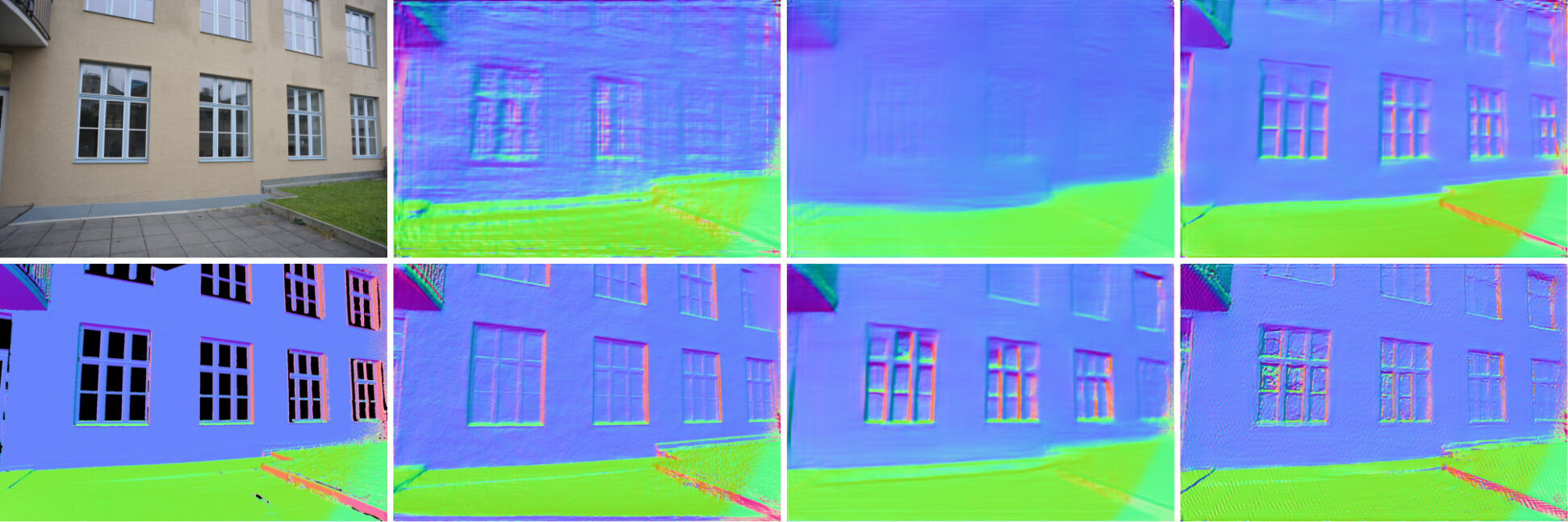}}\\
            & \lowerheadnormal \\
            & { } \\
            & \upperhead \\
            & \multicolumn{4}{c}{\includegraphics[width=\textwidth]{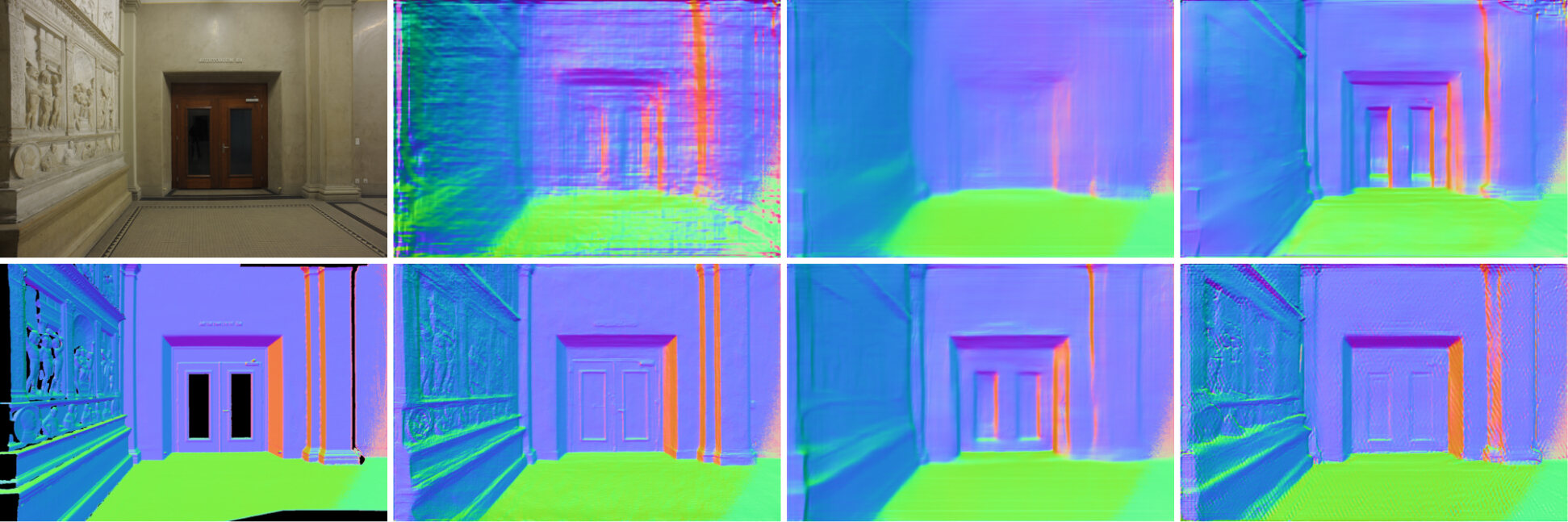}}\\
            & \lowerheadnormal \\
        \end{tabular}
    }
\end{figure*}

\clearpage

\begin{figure*}[p]
    \centering
    % Define a new command for dataset labels if there is only one sample
% \newcommand{\sampleNameSingle}[1]{
%     \multirow{1}{*}[1.3cm]{\rotatebox{90}{#1}}
% }
\renewcommand{\arraystretch}{0.9}
    \resizebox{\textwidth}{!}{
        \begin{tabular}[ht]{c  p{\viscolumn} p{\viscolumn} p{\viscolumn} p{\viscolumn} }
    
            \multirow{39}{*}{\rotatebox[origin=c]{90}{\scannet}}
            & \upperhead \\
            & \multicolumn{4}{c}{\includegraphics[width=\textwidth]{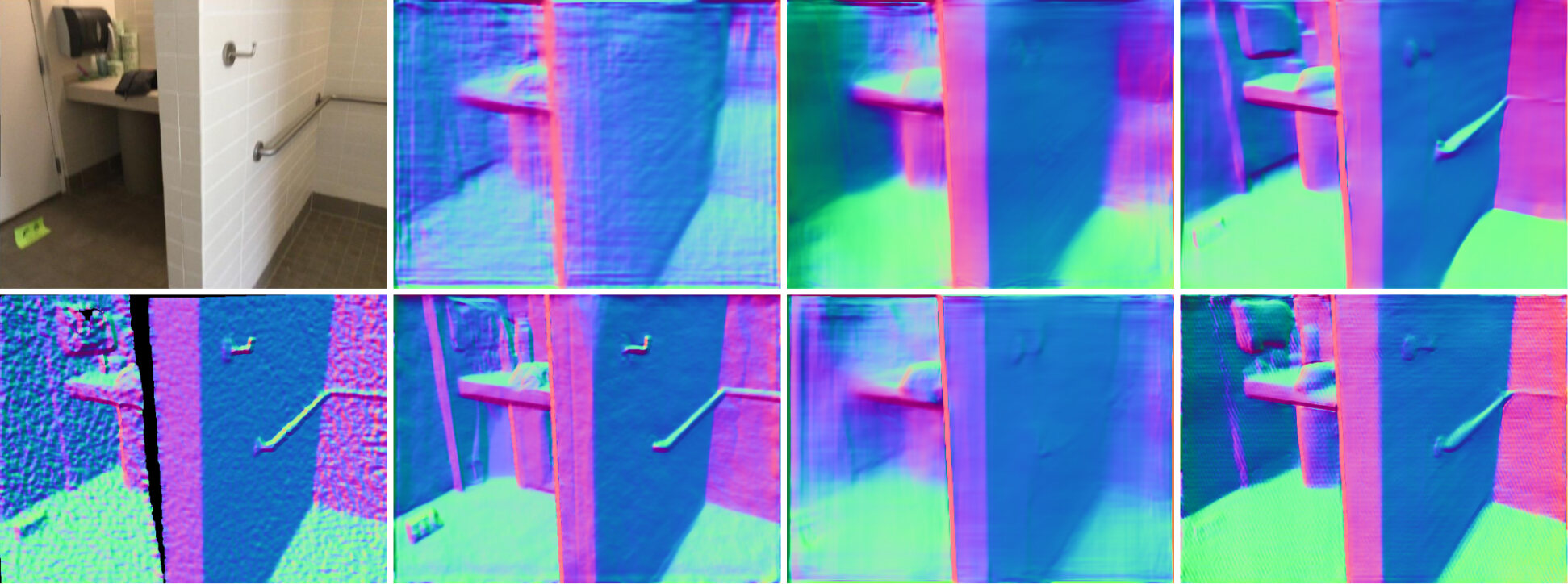}}\\
            & \lowerheadnormal \\
            & { } \\
            & \upperhead \\
            & \multicolumn{4}{c}{\includegraphics[width=\textwidth]{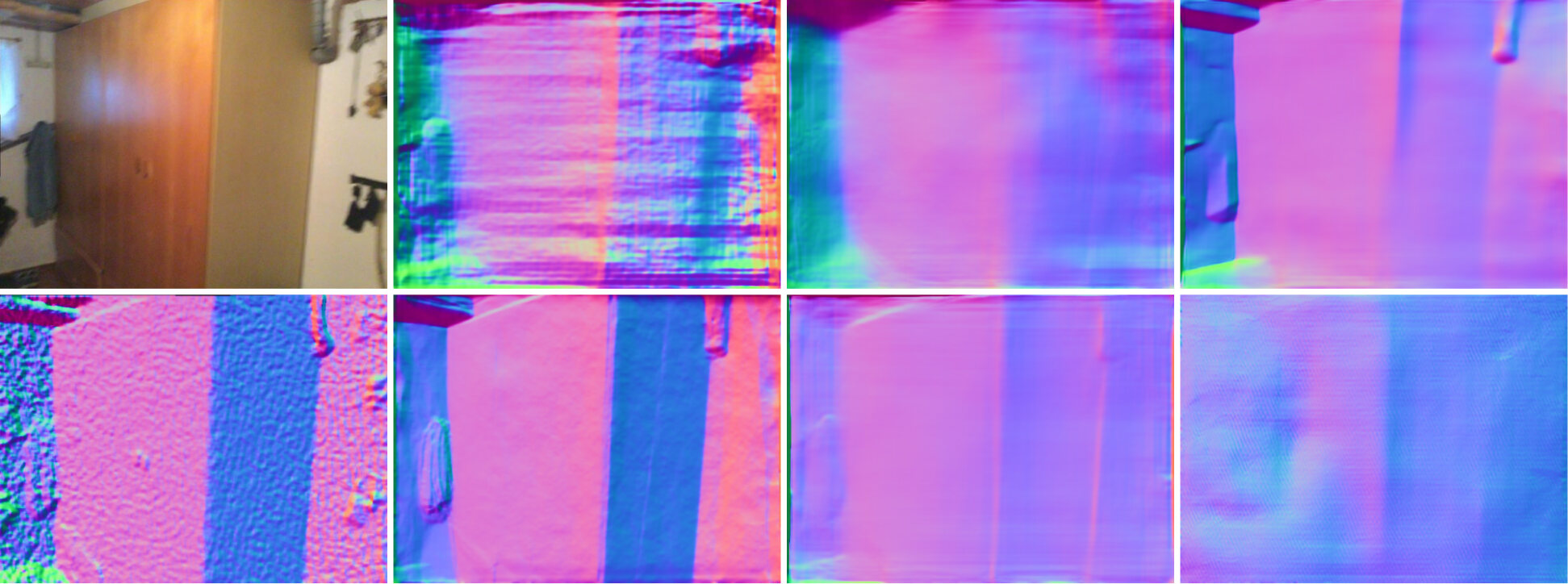}}\\
            & \lowerheadnormal \\
            & { } \\
            \hline
            & { } \\
            \multirow{17.5}{*}{\rotatebox[origin=c]{90}{\diode}}
            & \upperhead \\
            & \multicolumn{4}{c}{\includegraphics[width=\textwidth]{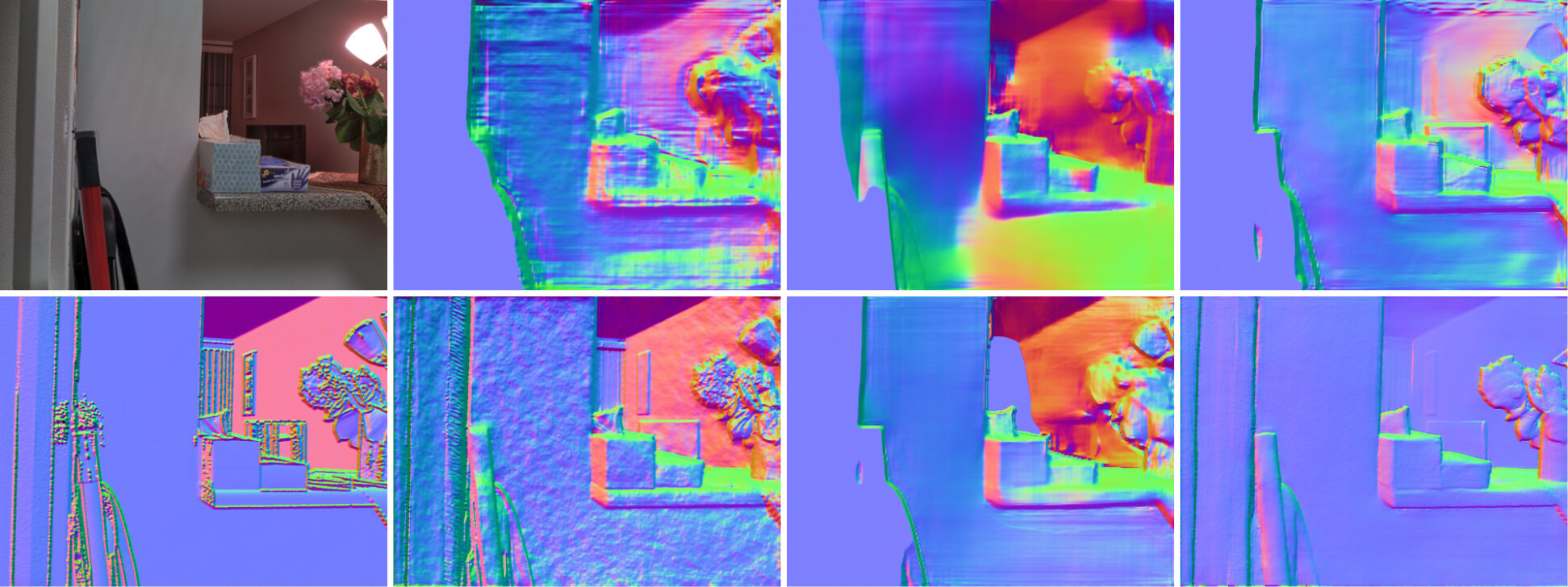}}\\
            & \lowerheadnormal \\
        \end{tabular}
    }

    \vspace{-4mm}
    \caption{
        \textbf{Qualitative comparison (unprojected, colored as normals)} of monocular depth estimation methods across different datasets. Ground truth normals are derived from the ground truth depth maps.
    }
    \label{fig:supp_qualitative_normal}
\end{figure*}

\end{document}